\documentclass[lettersize,journal]{IEEEtran}

\IEEEoverridecommandlockouts                              % This command is only needed if 
\usepackage{graphicx} % for pdf, bitmapped graphics files
\makeatletter
\let\MYcaption\@makecaption
\makeatother

\usepackage[font=scriptsize]{subcaption}

\makeatletter
\let\@makecaption\MYcaption
\makeatother

\usepackage{amsmath} % assumes amsmath package installed
\usepackage{amssymb}  % assumes amsmath package installed
\usepackage{algorithm, algcompatible} % for writing algorithms
\usepackage{bm}
\usepackage{xcolor}
\usepackage{dblfloatfix}
\usepackage{multirow}
\usepackage{cite} % For compressed ranges (works with IEEETran)

\usepackage{amsthm}

\newtheorem{proposition}{Proposition}

\theoremstyle{definition}
\newtheorem{definition}{Definition}[section]

\usepackage[shortlabels]{enumitem}

\newcommand{\scriptO}{{O}}
\newcommand{\scripto}{{o}}
\DeclareMathOperator{\atantwo}{atan2}

\DeclareMathOperator*{\argmin}{arg\,min}

\newcommand{\Obs}{\mathcal{O}}
\newcommand{\Vis}{\mathcal{V}}
\newcommand{\VisProb}{\mathcal{P}_{\textrm{vis}}}
\newcommand{\func}[1]{\texttt{#1}}
\newcommand{\FALSE}{\textbf{false}}
\newcommand{\TRUE}{\textbf{true}}

\newcommand{\obs}{
  \mathchoice
    {{\scriptstyle\mathcal{O}}}% \displaystyle
    {{\scriptstyle\mathcal{O}}}% \textstyle
    {{\scriptscriptstyle\mathcal{O}}}% \scriptstyle
    {\scalebox{.6}{$\scriptscriptstyle\mathcal{O}$}}%\scriptscriptstyle
  }

\algnewcommand{\LINECOMMENT}[1]{\STATE \(//\) #1}
\algtext*{ENDIF}
\algtext*{ENDFOR}
\algtext*{ENDWHILE}
\algtext*{ENDLOOP}

\usepackage[nolist]{acronym}

\renewcommand{\baselinestretch}{0.98}

\usepackage[colorlinks=true]{hyperref}  %hyperref still needs to be put at the end!

\title{\LARGE \bf
\acf{STALC} 
}

\author{
Cora A. Duggan$^{1,2}$, 
Adam Goertz$^{1}$,
Adam Polevoy$^{1,2}$,
Mark Gonzales$^{2}$,\\
Kevin C. Wolfe$^{1}$,
Bradley Woosley$^{3}$,
John G. Rogers III$^{3}$,
and Joseph Moore$^{1,2}$% <-this % stops a space
\thanks{$^{1}$Johns Hopkins Applied Physics Laboratory, Laurel, MD
	20723, USA. Email: {\tt\small Cora.Duggan@jhuapl.edu, jlmoore@jhu.edu}}%
\thanks{$^{2}$Department of Mechanical Engineering, Johns Hopkins University, Baltimore, MD 21218, USA.}%
\thanks{$^{3}$DEVCOM Army Research Laboratory, Adelphi, MD 20783, USA.}%
\thanks{Distribution Statement A: Approved for public release. Distribution is unlimited.}
}

\begin{document}

\begin{acronym}
    \acro{MIP}[MIP]{Mixed-Integer Programming}
    \acro{MILP}[MILP]{Mixed-Integer Linear Programming}
    \acro{STALC}[STALC]{Stratified Topological Autonomy for Long-Range Coordination}
    \acro{GNN}[GNN]{Graph Neural Network}
    \acro{GAT}[GAT]{Graph Attention Network}
    \acro{ML}[ML]{Machine Learning}
    \acro{RL}[RL]{Reinforcement Learning}
    \acro{PPO}[PPO]{Proximal Policy Optimization}
    \acro{GPU}[GPU]{Graphics Processing Unit}
    \acro{MPPI}[MPPI]{Model Path Predictive Integral Control}
    \acro{DEM}[DEM]{Digital Elevation Model}
    \acro{MMDP}[MMDP]{Multi-Robot Markov Decision Process}
    \acro{MAPF}[MAPF]{Multi-Agent Path Finding}
    \acro{DTG}[DTG]{Dynamic Topological Graph}
    \acro{MCMP}[MCMP]{Multi-Robot Coordinated Maneuver Problem}
    \acro{MCRP}[MCRP]{Multi-Robot Coordinated Reconnaissance Problem}
    \acro{SOTA}[SOTA]{State-of-the-Art}
    \acro{NMPC}[NMPC]{Nonlinear Model Predictive Control}
\end{acronym}

\maketitle
\thispagestyle{empty}
\pagestyle{empty}

%%%%%%%%%%%%%%%%%%%%%%%%%%%%%%%%%%%%%%%%%%%%%%%%%%%%%%%%%%%%%%%%%%%%%%%%%%%%%%%%
\begin{abstract}
In this paper, we present \acf{STALC}, a hierarchical planning approach for multi-robot coordination in real-world environments with significant inter-robot spatial and temporal dependencies. At its core, \ac{STALC} consists of a multi-robot graph-based planner which combines a topological graph with a novel, computationally efficient mixed-integer programming formulation to generate highly-coupled multi-robot plans in seconds. To enable autonomous planning across different spatial and temporal scales, we construct our graphs so that they capture connectivity between free-space regions and other problem-specific features, such as traversability or risk. We then use receding-horizon planners to achieve local collision avoidance and formation control. To evaluate our approach, we consider a multi-robot reconnaissance scenario where robots must autonomously coordinate to navigate through an environment while minimizing the risk of detection by observers. Through simulation-based experiments, we show that our approach is able to scale to address complex multi-robot planning scenarios. Through hardware experiments, we demonstrate our ability to generate graphs from real-world data and successfully plan across the entire hierarchy to achieve shared objectives.
\end{abstract}


%%%%%%%%%%%%%%%%%%%%%%%%%%%%%%%%%%%%%%%%%%%%%%%%%%%%%%%%%%%%%%%%%%%%%%%%%%%%%%%%
\section{Introduction}

Planning for highly collaborative multi-robot teams remains a fundamental research challenge, especially when significant interaction between team members is required \cite{korsah2013taxonomy}. Even when restricted to discrete actions and states, 
the planning problem
can quickly become computationally intractable \cite{Torreno2017}. This situation is further exacerbated when robot teams must autonomously reason about continuous-time dynamics, dynamic obstacles, and complex environments. In this paper, we present \acf{STALC}, a unified approach for coordinated motion planning with multi-robot teams. \ac{STALC} addresses challenges associated with highly-coupled multi-robot policies and planning across different temporal and spatial resolutions.

Our primary contribution is an optimal graph-based planning approach which can efficiently generate high-level multi-robot policies when there is a high-degree of interdependence between robot actions. Our approach relies on the notion of a \emph{\acf{DTG}}, where graph edges vary dynamically based on the state of the robot team. By ensuring that our objective function is convex when integrality constraints are relaxed, we formulate an efficient mixed-integer program that can solve for coordinated multi-robot plans in seconds. Our graph-based representation of the environment affords us the opportunity to capture important features of the lower-level planning problems. To do so, we construct graphs that embed a type of problem-dependent roadmap, where edges between nodes correspond to paths through free-space, and edge weights embed costs associated with those paths. 
We solve the graph problem for a time-extended horizon to enable \textit{long-range} coordination among robots, which we define as planning for robot interactions across several future steps and large spatial distances (e.g., kilometer scale).
When executing a long-range plan, we use the underlying collision-free paths associated with our graph edges to formulate objectives for receding-horizon planning and enable more local behaviors like robot collision avoidance and formation control. 

\begin{figure}[t!]
   \centering
   \includegraphics[width=0.9\linewidth]{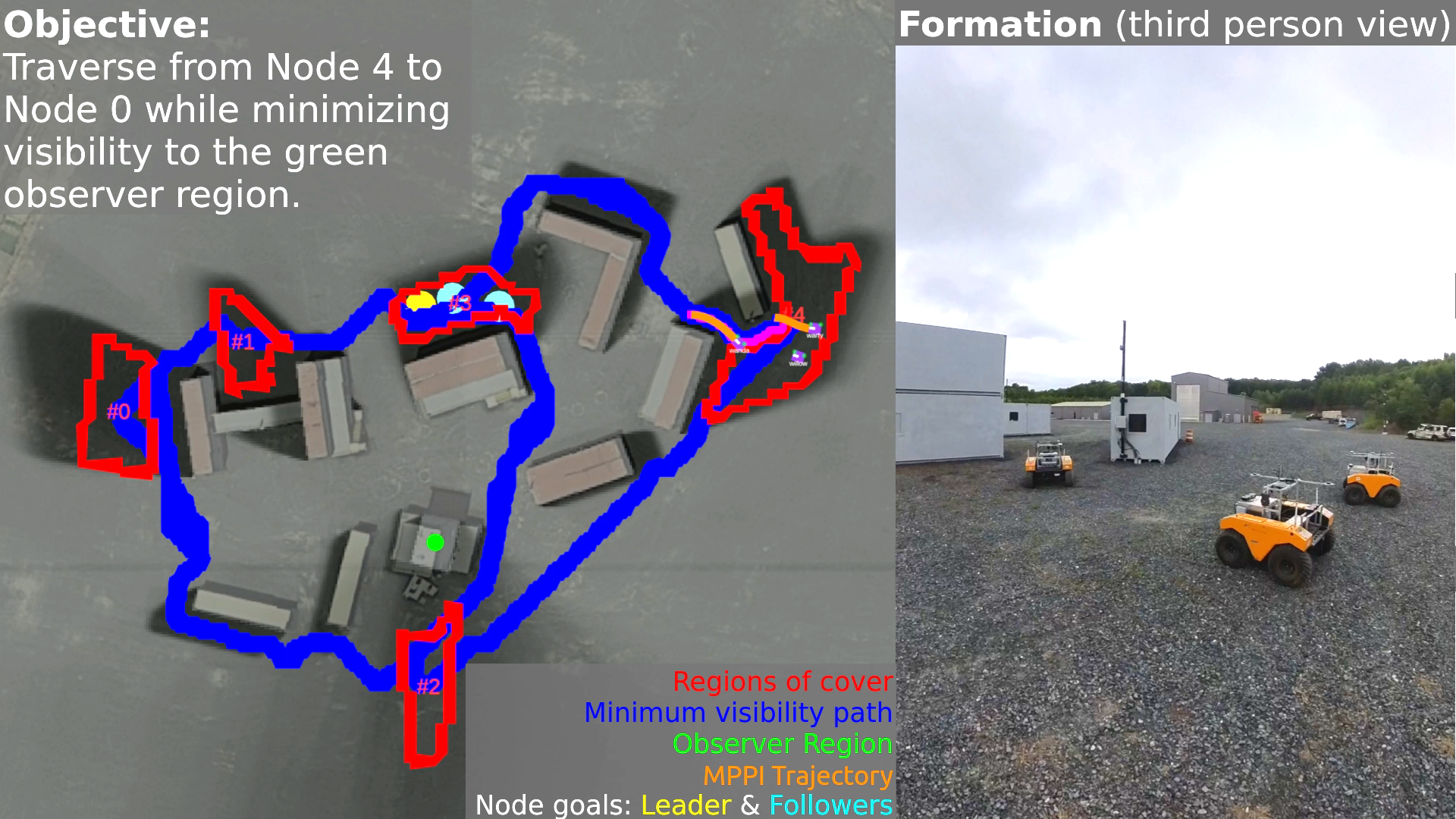}
   \vspace*{-2mm}
   \caption{Experimental operational scenario to minimize visibility while traversing an urban environment. A team of three robots autonomously navigate the environment while balancing the risk and rewards of moving in formations and providing overwatch for a reconnaissance mission.}
   \label{fig:cover_graph1_formation}
   \vspace*{-4mm}
\end{figure}

More specifically, we focus on solving a highly-coupled \acf{MCRP}. 
In this problem, the robot team must autonomously navigate through an environment while minimizing the probability of detection by observers at expected locations. 
The risk of detection can be mitigated through tactical \textit{maneuvers}, which we define as planned single or multi-robot action sequences that consider not only obstacle-free motion, but also the overarching mission objectives. 
For example, robots may coordinate to provide support to navigating team members through a line-of-sight ``overwatch'' maneuver.
These types of maneuvers can require both spatial and temporal coordination between robots.

To address this scenario, we formulate a visibility metric for segmenting the environment and constructing our topological graph. We evaluate our approach to solving the \ac{MCRP} through simulation-based experiments and 
demonstrate that our method improves computational efficiency when scaling with both graph size and team size by several orders of magnitude compared to existing state-of-the-art approaches.
In hardware experiments (e.g., Fig.~\ref{fig:cover_graph1_formation}), we demonstrate our method's ability to generate graphs from real-world data, produce feasible plans across the hierarchy, and realize mission objectives in forested and urban environments. 

This manuscript substantially extends the work in \cite{dimmig2023mip} which introduced a method for multi-robot planning on a \ac{DTG}. In the work presented herein, we report algorithms for generating the \ac{DTG} and utilizing the optimized plans, resulting in a full end-to-end system for coordinated multi-robot planning.
Our key contributions are as follows:
\begin{itemize}
\item A novel, computationally-efficient mixed-integer optimization formulation capable of generating highly-coupled multi-robot plans on graphs.
\item An algorithm to construct \acp{DTG} from terrain data for the multi-robot reconnaissance problem.
\item A hierarchical planning architecture for end-to-end multi-robot planning of coordinated maneuvers.
\item An evaluation of our approach in simulation and hardware experiments and comparison to existing approaches.
\end{itemize}

%%%%%%%%%%%%%%%%%%%%%%%%%%%%%%%%%%%%%%%%%%%%%%%%%%%%%%%%%%%%%%%%%%%%%%%%%%%%%%%%
\section{Related Work}
\label{sec:related_work}

Coordination of multi-robot teams presents a major computational challenge, especially when robots must collaborate toward shared objectives across different temporal and spatial scales \cite{Yan2013,Verma2021}. While the coordination problem can be modeled generally as a \ac{MMDP}, these formulations scale exponentially with the number of robots \cite{campbell2013multiagent}. 
To improve computational tractability, researchers have employed various abstractions, such as discretizing the state and action spaces or formulating problems that reduce the degree of robot interaction. 

\subsection{Environment Segmentation and Graph Generation}
Topological graphs, where nodes embed critical locations and edges encode their spatial connectivity, have served as an explicit method of connecting task-based state discretizations to an underlying metric space topology. 
Such graph-based roadmaps have a long history for single robot planning \cite{latombe2012robot} and have been constructed via a wide variety of segmentation strategies \cite{bormann2016room}. These strategies include segmenting free-space in indoor environments \cite{hollinger2008proofs} and terrain traversability in outdoor environments \cite{tremblay2025topological}. In some cases, for heterogeneous robot teams, distinct graphs have been constructed to capture different robot mobility characteristics (e.g., \cite{stankiewicz2018motion}).

\subsection{Multi-Robot Task Allocation}
In settings where a high-level, task-based problem abstraction can be treated independently of low-level coordination,
\cite{korsah2013taxonomy} and \cite{gerkey2004formal} provide comprehensive overviews of multi-robot task allocation approaches and propose two different taxonomies. 
As discussed in \cite{korsah2013taxonomy}, even under discrete state and action approximations, task allocation problems with significant coupling between multi-robot policies present a major combinatorial challenge. The literature for multi-robot task allocation is vast (see, \cite{ka2024systematic}). In this paper, we focus primarily on approaches that reason about significant spatial interdependence among tasks and robots.

\subsection{Multi-Robot Coordination on Topological Graphs}
Several multi-robot coordination problems assume an underlying topological graph structure, including the \ac{MAPF} problem \cite{silver2005cooperative}, the vehicle routing problem \cite{bredstrom2008combined}, and the set covering problem \cite{balas1976set}.

While some market-based \cite{lagoudakis2004simple,mosteo2008multi} and game-theoretic approaches have leveraged underlying graphs for spatial coordination \cite{Shishika2018}, most methods for multi-robot coordination using graphs can be categorized as search-based, optimization-based, and learning-based approaches. 

\subsubsection{Search-based Approaches}
Search-based approaches to multi-robot coordination exploit problem structure or effective heuristics to search through the joint multi-robot state space. In particular, conflict-based search \cite{Sharon2015} has emerged as a powerful two-stage approach for the \ac{MAPF} problem. Earlier approaches (e.g., \cite{standley2010finding}) employed A$^*$ search to solve this class of problem. Heuristic search algorithms have also been employed to solve versions of the vehicle routing problem with temporal constraints (e.g., \cite{braysy2002tabu}). In \cite{Limbu2023}, researchers present a search-based approach for solving the problem of team coordination on graphs, where robot team members can provide support to one another \cite{limbu2024team}. By proposing the notion of a joint state graph, they solve this multi-robot coordination problem via A$^*$ search. In many cases, these general algorithmic approaches either rely on assumptions about problem structure (e.g., \cite{Sharon2015}), or scale poorly with the number of robots (e.g., \cite{standley2010finding, Limbu2023}).

\subsubsection{Optimization-based Approaches}
\ac{MIP} has proved to be another effective means of generating multi-robot coordination policies on graphs. For instance, in \cite{koes2005heterogeneous}, the authors present a \ac{MILP} approach for coordinating heterogeneous multi-robot teams under spatial and temporal constraints. In \cite{mitchell2015multi}, researchers use \ac{MILP} to formulate a multi-robot coverage problem. \ac{MIP} formulations have also proved effective for solving variants of the multi-vehicle robot routing problem with time constraints (e.g., \cite{Kamra2015}). In \cite{Flushing2017}, the authors explore the problem of joint task allocation and scheduling on graphs under communication constraints. The authors of \cite{Asfora2020} and \cite{Yu2016} use \ac{MILP} formulations to solve a multi-robot non-adversarial search and \ac{MAPF} problem, respectively. In \cite{dimmig2023mip}, researchers formulate a \ac{MILP} approach to solve a multi-robot coordination problem on graphs with support, similar to the problem described in \cite{limbu2024team}. While a powerful general framework, \ac{MIP} is often not computationally efficient enough to be run online (e.g., \cite{Asfora2020}) and authors frequently resort to simplifying heuristics \cite{Yu2016}.

\subsubsection{Learning-based Approaches}
\ac{ML} has also been applied to address the challenges associated with multi-robot coordination on topological graphs. In particular, \acp{GNN} \cite{scarselli2008graph} have emerged as a powerful means of representing multi-robot policies. In \cite{li2020graph}, the authors use \acp{GNN} to generate policies for decentralized multi-robot path planning. In \cite{tolstaya2021multi}, researchers use \acp{GNN} to address the multi-robot coverage problem, where the underlying graph is constructed from waypoints, robot positions, paths through free space, and communications constraints. Researchers have also explored more advanced \ac{GNN} frameworks. For instance, in \cite{Wang2020,Wang2022}, the authors use a type of \ac{GNN} known as a \ac{GAT} to achieve multi-robot coordination with temporal and spatial constraints. 

In \cite{dai2024dynamic} and \cite{dai2025heterogeneous}, authors use \ac{RL} to achieve dynamic coalition forming and routing and task allocation and scheduling, respectively. In \cite{limbu2024scaling}, researchers use Q-learning and \ac{PPO} for multi-robot coordination on graphs, and demonstrate the ability to scale to larger numbers of robots and graph sizes when compared to the search-based method in \cite{Limbu2023}. 

While able to scale to larger environments and team sizes, learning-based approaches often suffer from poor generalization to environments that were not included in the training distribution and cannot provide optimality guarantees \cite{limbu2024scaling}.
\subsection{Multi-Robot Task and Motion Planning}
Topological graph representations of the environment can more tightly couple task allocation and motion planning. For example, \cite{stankiewicz2018motion} encodes free-space connectivity in graph edges and embeds metric path costs as edge weights, allowing task allocation to account for team travel costs. Similarly, precomputed metric plans between nodes can be used to guide local planners toward a global solution. \cite{kumar2012multi} employs a hierarchy of graphs constructed using probabilistic roadmaps to solve a multi-robot routing problem. \cite{salvado2021combining} presents an alternative roadmap approach to task allocation and motion planning.

Other multi-robot approaches attempt to achieve an even tighter coupling between high and low-level planning. For instance, \cite{turpin2014capt} presents an approach for joint trajectory optimization goal assignment. In \cite{arul2019lswarm}, the authors describe an approach for achieving both collision avoidance and coverage constraints. In \cite{motes2020multi}, the authors leverage conflict-based search to achieve multi-robot task allocation and motion planning. Recently, \ac{RL} approaches \cite{lai2025roboballet,lin2024decentralized} have demonstrated solving the joint multi-robot task and motion planning problem. 
\vspace{1mm}

In this paper, we solve a joint multi-robot coordination and motion planning problem for a reconnaissance navigation task using an underlying topological roadmap. We first construct a topological graph by segmenting the environment based on visibility and traversability. We then build on the approach in \cite{dimmig2023mip} to efficiently solve a highly-coupled multi-robot graph-based coordination problem using \ac{MILP}. During plan execution, we leverage the metric paths embedded in our graph to guide local, low-level motion planning and formation control. We show that our approach, unlike other \ac{MIP}-based approaches, is computationally tractable enough to afford real-time updates, and that our roadmap-like approach can effectively guide low-level motion plans in real-world environments.

%%%%%%%%%%%%%%%%%%%%%%%%%%%%%%%%%%%%%%%%%%%%%%%%%%%%%%%%%%%%%%%%%%%%%%%%%%%%%%%%
\section{Problem Formulation}
\label{sec:problem_formulation}

We consider a general \acf{MCMP}, where a team of $n_A$ homogeneous robots must coordinate over time to achieve a shared objective. 
For $i\in[1,n_A]$, the dynamics of the $i^\textrm{th}$ robot are given as $x_{t+1}^i = f(x_t^i, u_t^i)$, where $x_t^i\in\mathbb{R}^{n_x}$ are robot states and $u_t^i\in\mathbb{R}^{n_u}$ are robot actions at time index $t\in[1,n_T]$. Let the robot team state at time $t$ be $X_t = \{x_t^1, \dots, x_t^{n_A} \}$. Given a joint state objective function $g_x$, a cost on individual robot actions $g_u$, a set of joint state constraints $c_x$, and input constraints $c_u$, we write the \ac{MCMP} as follows: 
\begin{equation}
\begin{alignedat}{3}
\min_{u^i_t \enspace \forall i,t} \quad & \sum_{t=1}^{n_T} \bigg( g_x(t, X_t, X_D) + \sum_i g_u(u^i_t)\bigg) \\
\text{s.t.} \quad & x^i_{t+1} = f(x^i_t, u_t^i), \quad X_1= X_B, &\forall i,t~ \\
&c_x(X_t)\le 0, \quad c_u^i(u_t^i) \leq 0, &\forall i,t.
% &X_0= X_B.
\end{alignedat}
\label{eq:basic_opt_problem}
\end{equation}
Here, $X_D$ is a global goal state and $X_B$ is the initial state of the multi-robot team. 
Given that $g_x$ is a substantially complex, environment-dependent objective function that introduces tight, long-horizon inter-robot spatial-temporal dependencies, the optimization problem in (\ref{eq:basic_opt_problem}) presents a fundamental combinatorial challenge and scales exponentially with $n_A$ \cite{solovey2016finding}. As discussed in Sec.~\ref{sec:related_work}, to improve computational tractability, the problem in (\ref{eq:basic_opt_problem}) can be decomposed into a discrete high-level, multi-robot coordination problem and a continuous lower-level multi-robot motion planning problem. 

\subsection{High-Level Multi-Robot Coordination}
To formulate the high-level multi-robot coordination problem, we discretize the state and action spaces by introducing the notion of a \acf{DTG}. 
Consider a weighted graph of the form $G=(V, E, W_E)$ with a set of nodes $v\in V$, edges $e\in E$, and edge weights $w_e\in W_E$. Let the set of graph nodes, $V$, be given by a mapping $M_v: \mathcal{R} \mapsto V$, where $v_j = M_v(r_j)$. $\mathcal{R}$ is a set of non-overlapping regions $r\in\mathcal{R}$, such that $r\subseteq \mathbb{R}^2$ and $r_j \cap r_k = \varnothing,~\forall j \neq k$. 

Now let $\mathcal{T}_{\textrm{feas}}$ be the set of all feasible paths between regions, and let $\mathcal{T}_{v_j v_k}$ be a path between regions $r_j$ and $r_k$. We then define the set of directed graph edges as ordered pairs of nodes connected via free-space, ${E=\{(v_j,v_k)\in V\times V: \exists\mathcal{T}_{v_j v_k}\in \mathcal{T}_{\textrm{feas}}\}}$.

Finally, let the state of the robot team on the graph at time $t$ be $S_t=\{ s_t^{1}, \dots, s_t^{n_A} \}$, where $s_t^{i} \in E$ is the edge occupied by the $i^\textrm{th}$ robot at time $t$. 
We can now define a \ac{DTG} as follows:
\begin{definition}[\acf{DTG}]
Given a weighted graph $G=(V, E, W_E)$, a \ac{DTG} is a weighted graph where the edge weights are defined as a function of the state $S_t$, such that $w_{e,t} \in W_E : S_t \rightarrow \mathbb{R}_{>0}$. 
\label{def:dtg}
\end{definition}
To describe the evolution of a robot state on the graph, we define the Next Edge Action Set, $A$, as follows:  
\begin{definition}[Next Edge Action Set]
For a directed edge $e =(v_j,v_k) \in E$, the Next Edge Action Set $A$ is defined as $A(e)=\{(v_k,v_l) \in E : v_l\in V\}$.
\label{def:action_set}
\end{definition}
Given Def. \ref{def:dtg} and \ref{def:action_set}, we can formulate the graph-based multi-robot coordination problem as follows:
\begin{equation}
\begin{alignedat}{3}
\min_{s^i_t\enspace \forall i,t} \enspace &\sum_{t=1}^{n_T} \sum_{i=1}^{n_A} w_{s_t^i,t}(S_t) \\
\text{s.t.} \enspace &s^i_{t+1} \in A(s^i_t), \quad \forall i,~t \in [1,n_T-1] \\ 
&S_1 = S_B,\quad S_{n_T} = S_D.
\end{alignedat}
\label{eq:dtg_optimal}
\end{equation}
Here, $S_B$ and $S_D$ are the initial and final robot graph states.
\subsection{Local Multi-Robot Motion Planning}
\label{sec:local_planning}
While this graph-based problem formulation, if solved, results in high-level coordination, we still ultimately desire to generate policies over continuous state and action spaces. Using the resulting routes from (\ref{eq:dtg_optimal}), we can formulate a series of reduced local optimization problems of the same form as (\ref{eq:basic_opt_problem}). Each local optimization problem corresponds to a coalition of robots $\mathcal{C}_{e}$ sharing edge $e\in E$, where $\mathcal{C}_{e} =\{i\in \mathbb{Z}^+ : s^i = e \}$ with joint state space $X^{\mathcal{C}_e}$. In this case $g_{x}$, $g_{u}$, $c_{x}$, and $c_{u}$ may be replaced by edge-specific functions $g_{x,e}$, $g_{u,e}$, $c_{x,e}$, and $c_{u,e}$. The goal becomes $X^{\mathcal{C}_e}_G$, the initial state becomes $X_B^{\mathcal{C}_{e}}$, and the time steps become $t_e \in [1,n_{T_e}]$, which correspond to the time traversing edge $e$. 

\subsection{\acl{MCRP}}
In this paper, we focus on a particular instance of the \ac{MCMP}, which we refer to as the \acf{MCRP}. In this scenario, a team of robots must traverse an environment to a goal location in minimum time while minimizing the risk of detection by observers at expected locations. To mitigate risk, the robot team can employ the following strategies:
\begin{enumerate}[label=\roman*), ref=\roman*]
    \item \textbf{Minimize Visibility}: Individual robots can reduce risk by minimizing visibility with respect to observers. \label{action:min_vis}
    \item \textbf{Provide Overwatch}: Robots can oversee portions of team members' movement through high-visibility areas. \label{action:overwatch}
    \item \textbf{Move in Formation}: Robots can work together and move in formations through high-visibility areas. \label{action:formation}
\end{enumerate}
These actions are based on foundations of tactical mission planning \cite{grindle2004ipb, thomas2024ipb, rogers2023tactical}. 
While many methods for representing risk have been explored, we employ a simple strategy of capturing risk with path costs \cite{bertsekas2011dynamic}.

As proposed, the \ac{MCRP} is a useful exemplar of the more general \ac{MCMP}, as it introduces tight spatial coupling through strategy~(\ref{action:overwatch}) and encourages coalition forming in strategy~(\ref{action:formation}). We can achieve these strategies via a specific formulation of the weighted edge cost $w_{e,t}$ in a \ac{DTG}.
We note that the \ac{MCRP} reduces to Team Coordination on Graphs with Risk Edges (TCGRE) from \cite{limbu2024team} when only considering strategy~(\ref{action:overwatch}). 

The \ac{MCRP} is composed of a task allocation problem and a local motion planning problem.
Following the iTax taxonomy in \cite{korsah2013taxonomy}, the task allocation portion of the \ac{MCRP} falls in the cross-schedule dependency class (XD), where robots must coordinate for multi-task (MT), multi-robot tasks (MR), over a time-extended allocation (TA). In \cite{korsah2013taxonomy}, this subclass is defined as XD [MT-MR-TA]. At the time of this taxonomy, the authors were not aware of any solutions to this problem. A more recent analysis of this problem in \cite{miloradovic2025formal} additionally confirms this class of problems has not been explored further.

%%%%%%%%%%%%%%%%%%%%%%%%%%%%%%%%%%%%%%%%%%%%%%%%%%%%%%%%%%%%%%%%%%%%%%%%%%%%%%%%
\section{\acs{STALC} Technical Approach}
\label{sec:technical_approach}

\begin{figure*}[tbh]
   \centering
   \includegraphics[width=0.9\linewidth]{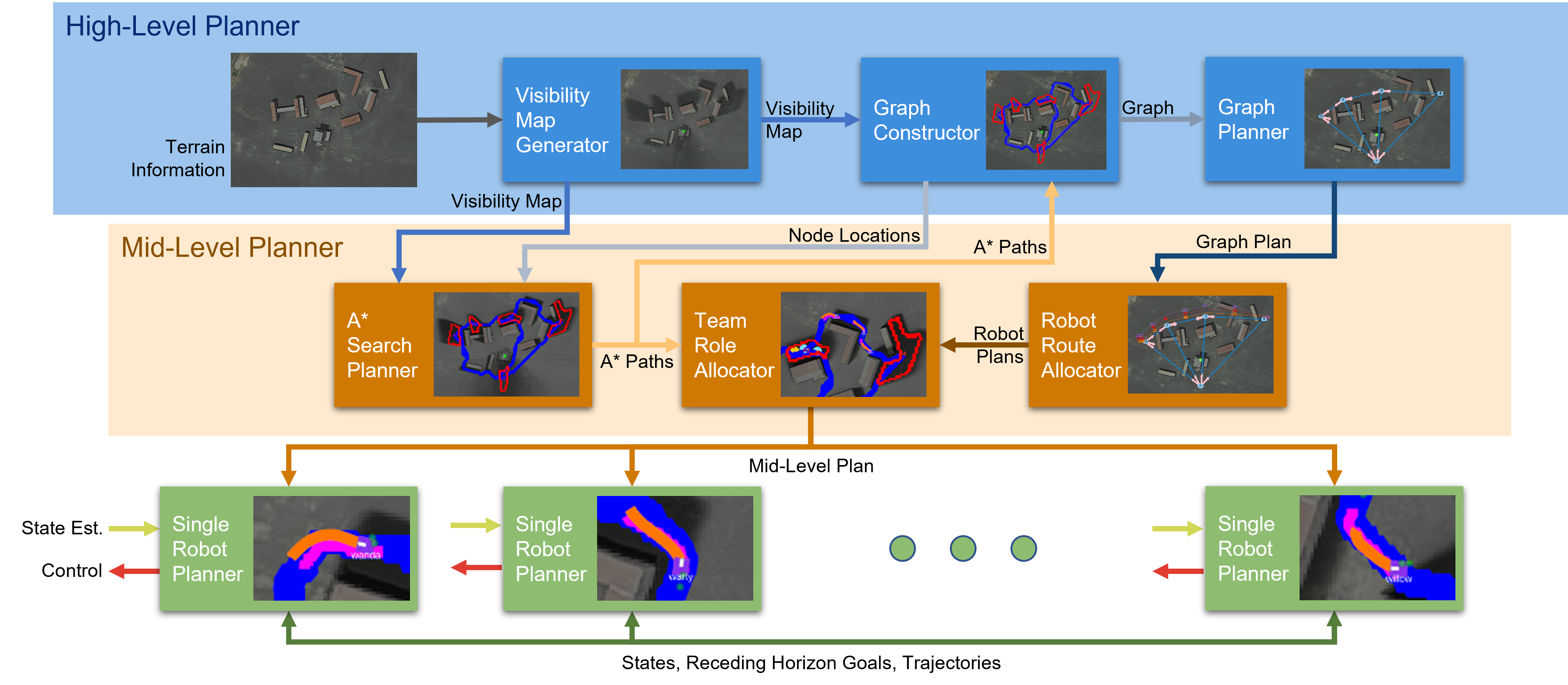}
   \vspace*{-4mm}
   \caption{\acs{STALC}: Hierarchical planning architecture for a multi-robot team to autonomously coordinate while traversing an environment. The high-level planner segments the environment into a topological graph structure and generates a plan on the graph for the robot team through the environment. The graph construction is informed by the mid-level planner which generates A$^*$ paths through the environment. The mid-level planner then takes the output of the graph planner and assigns routes to each robot through the environment. Each group of robots moving together are then assigned roles in their coalition (i.e., leader or follower). This information is sent to each robot's low-level planner to generate and execute dynamically feasible, collision-free paths.}
   \vspace*{-4mm}
   \label{fig:flow_diagram}
\end{figure*}

Our technical approach is designed to address 
the high-level coordination problem presented in (\ref{eq:dtg_optimal}) and the local optimization problem described in Sec.~\ref{sec:local_planning}. To this end, we present a three-level planning hierarchy, where a top-level graph-based planner coordinates multi-robot teams, a mid-level A$^*$ motion planner generates feasible paths between nodes, and low-level \ac{MPPI} \cite{Williams2016} enables receding-horizon multi-robot motion planning. All three levels share underlying metric cost maps to maintain consistent planning across the hierarchy.
Both A$^*$ search and \ac{MPPI} are surrogate approaches, selected for their widespread adoption, and could be substituted for other methods for path planning \cite{karur2021survey} and stochastic \ac{NMPC} \cite{kazim2024recent}, respectively. This paper does not claim developments in these areas. Our key contributions are in the overall planning architecture and graph planning components.

While we believe our approach could apply more generally, in this paper, we focus on solutions to (\ref{eq:dtg_optimal}) and the local problem in Sec.~\ref{sec:local_planning} that solve the \ac{MCRP}.
We assume prior knowledge of the terrain and probability distributions for expected observer positions, as this information would have motivated the reconnaissance mission in this area. This information may be a coarse representation that evolves as we execute a plan or be provided with limited time before a scenario. In both cases, we require an approach for efficient planning to enable rapid response times to the given scenario and/or evolving information.

Fig.~\ref{fig:flow_diagram} shows a flow diagram of our hierarchical planning architecture applied to the MCRP. First, we use prior information to compute visibility scores, relative to the expected observer distributions, across an area of interest. Using the resulting visibility map, we segment the environment into regions of cover (or low visibility), which we use as nodes in a graph structure. We then calculate minimum visibility paths, using A$^*$ search, between regions to define graph edges.
Between each node and edge pair, we compute overwatch scores based on the visibility to the edge from within the node. 
Together, this information forms a topological graph of the environment. We encode this graph in our \ac{MILP} graph planner to compute an optimal plan for the multi-robot team. To form our mid-level plan, we assign each robot a route, which corresponds to an A$^*$ path, and a role in the coalitions of robots that form (i.e., leader or follower). Each robot then controls to its path using \ac{MPPI} in our low-level planner, performs overwatch, and/or moves in formation with their coalition. 
The following sections outline the algorithms that comprise each block of our overall hierarchical planner. 
First, we introduce the construction of graphs for the \ac{MCRP} in Sec.~\ref{sec:graph_generation}.
In Sec.~\ref{sec:mip_approach}, we present our core contribution, the graph planning algorithm. 
Finally, we present the technical detail of our mid-level and low-level planners in Sections \ref{sec:mid_level} and \ref{sec:low_level}, respectively.

%%%%%%%%%%%%%%%%%%%%%%%%%%%%%%%%%%%%%%%%%%%%%%%%%%%%%%%%%%%%%%%%%%%%%%%%%%%%%%%%
\section{Graph Generation}
\label{sec:graph_generation}

Constructing a graph that reflects the topology and path costs of the underlying metric space is essential for unifying the graph-based coordination problem in (\ref{eq:dtg_optimal}) and the local multi-robot motion planning problem from Sec.~\ref{sec:local_planning}. Because the graph is inherently connected to the scenario domain, we focus specifically on constructing a graph for the MCRP. 

\subsection{Visibility Map}
Since the MCRP aims to minimize risk due to visibility, the graph is generated by first constructing a visibility cost map. We accomplish this by adapting the notion of a viewshed\cite{Wang2000GeneratingVW} to the context of a distribution over observer positions. The viewshed of an area is a binary mask marking the regions of a \acf{DEM} that are visible from a given observer position. Our goal is to estimate the probability that each point in the environment can be seen by the observer.

Line-of-sight analyses, such as viewsheds, are commonly used in terrain analysis for tactical decision making (e.g., in \cite{grindle2004ipb, thomas2024ipb} for determining cover and concealment). 
Uncertainty in terrain measurements and observer locations can significantly impact the accuracy of viewshed calculations. While we do not address terrain uncertainty in this work, we formulate a probabilistic notion of a viewshed that incorporates uncertainty in the observer position.

We assume prior knowledge of a \ac{DEM} of the environment and a probability distribution, $\Obs$, over $\mathbb{R}^3$ representing the expected observer position. We do not make assumptions about the form of $\Obs$ except that we are able to sample from it. In the experimental evaluation that follows, we utilize Gaussian distributions (i.e., $\Obs \sim \mathcal{N}(\mu, \Sigma)$). 
Taking $N$ samples $\obs \sim \Obs$ from the observer distribution, we compute a viewshed $\Vis_{\obs}$ for each sample. Discretizing the environment into a 2D grid, for each point $(x, y)$, the visibility probability is the expectation over the viewsheds from the sampled observer positions. 
\begin{align}
    \mathop{\mathbb{E}_{\obs \in \Obs}}[\Vis_{\obs}(x,y)] \approx \frac{\sum_{i=1}^{N}{\Vis_{\obs_i}(x,y)}}{N} .\label{eq:visibility_prob}
\end{align}

\begin{figure}[t]
    \centering
    \includegraphics[width=0.7\columnwidth]{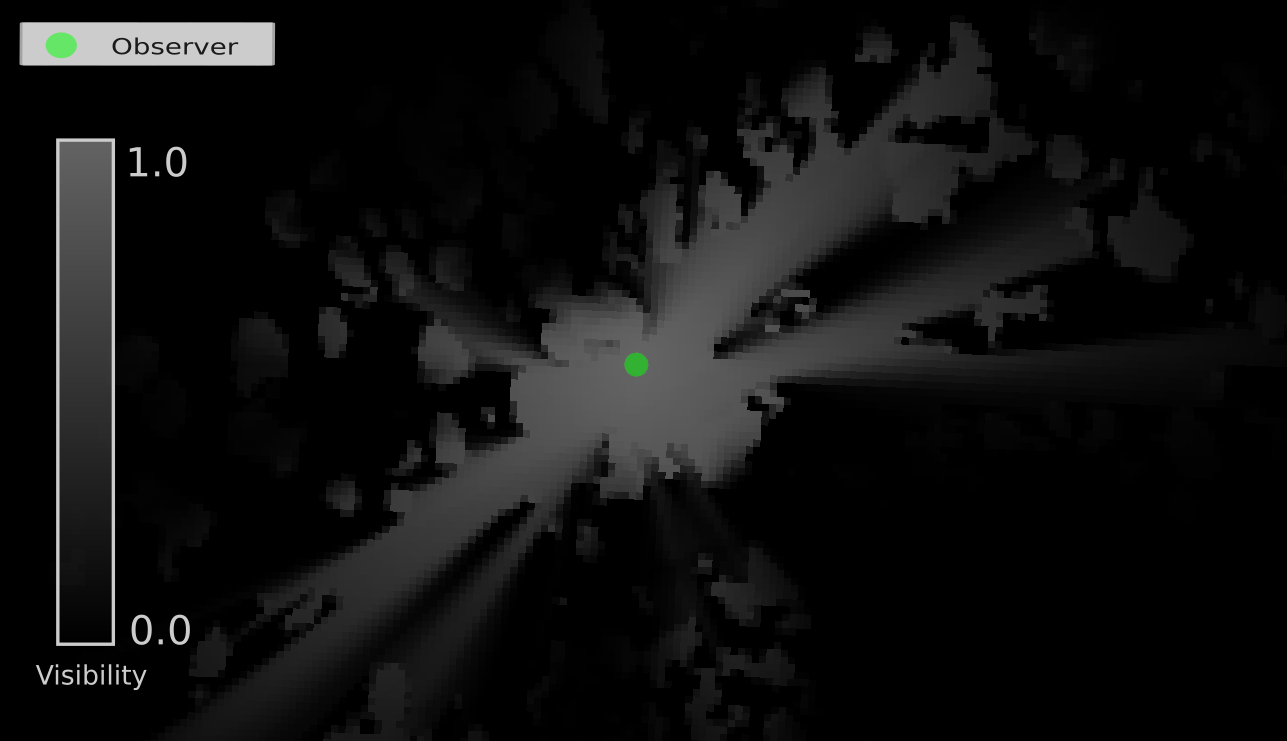}
    \captionsetup{font=scriptsize}
    \vspace*{-2mm}
    \caption{A visibility map for an observer positioned in a clearing. The gradient in each cell represents the visibility, with lighter being higher visibility.}
    \label{fig:only_vis_map}
    \vspace*{-4mm}
\end{figure}

We refer to the map containing the probability of being seen at each point
as the \textit{visibility map}. An example of a visibility map is depicted in Fig.~\ref{fig:only_vis_map} with a single observer. The lighter regions indicate higher visibility. 

Equation \ref{eq:visibility_prob} does not consider the distance from the observer when computing the probability of detection. In practice, however, distant objects appear less visible. We model the visibility of an object as proportional to its apparent angular size, $\theta$, which scales inversely with distance, $d$, from the observer (i.e., $\theta \sim \frac{1}{d}$). 
The distance is determined using the distance transform to compute the Euclidean distance from each map cell to the nearest observer distribution. 
For distributions with infinite support (e.g., Gaussian), we must choose a ``boundary'' for the distribution in order to compute the distance transform. In this work, we use the 2-$\sigma$ ellipse. 
Detection probability is then scaled by the ratio of the nearest enemy distance to the maximum effective distance $d_{\textrm{max}}$. $d_{\textrm{max}}$ is chosen based on the observer characteristics (e.g., when the robot's angular size falls below a threshold or spans fewer than a set number of pixels in a camera image). 
We define the distance-weighted visibility map, $\VisProb(x, y)$ for a position $(x,y)$, where distance to the nearest observer is $d(x,y)$.
\begin{align}
    \VisProb(x, y) = \mathop{\mathbb{E}_{\obs \in \Obs}}[\Vis_{\obs}(x,y)]\max\Big(1 - \frac{d(x,y)}{d_{\textrm{max}}}, 0\Big) .\label{eq:dist_weighted_visibility}
\end{align}

%%%%%%%%%%%%%%%%%%%%%%%%%%%%%%%%%%%%%%%%%%%%%%%%%%%%%%%%%%%%%%%%%%%%%%%%%%%%%%%%

\begin{algorithm}[thb]
\caption{Construct Topological Graph}
\label{alg:graph_construction}
\footnotesize
\begin{algorithmic}[1]
	\renewcommand{\algorithmicrequire}{\textbf{Input:}}
	\renewcommand{\algorithmicensure}{\textbf{Output:}}
	\renewcommand{\COMMENT}[2][.5\linewidth]{%
	  \leavevmode\hfill\makebox[#1][l]{//~#2}}
		\REQUIRE~~
		\begin{tabbing}
		\hspace*{3em} \= \kill % set the tabbings
            $\mathcal{D}$ \> Digital Elevation Model\\
			$C_m$ \> Obstacle map\\ 
            $\Obs$ \> Observer distribution\\
            $N$ \> Number of samples from observer distribution\\
            $\nu$ \> Visibility threshold\\
            $\Xi_{\textrm{min}}$ \> Minimum cover region size\\
            $\Xi_{\textrm{max}}$ \> Maximum cover region size\\
            $\lambda_p$ \> Visibility cost weight
		\end{tabbing}
		\ENSURE~~
		\begin{tabbing}
		\hspace*{3em} \= \kill % set the tabbings
            $V$ \> Graph nodes (cover region centroids)\\
            $E$ \> Graph edges\\
			$W_{\textrm{vis}}$ \> Visibility weight adjacency matrix\\
            $\mathcal{W}_{\textrm{ow}}$ \> Set of overwatch weight matrices, $W_{\textrm{ow}}(v)$\\
		\end{tabbing}
        \STATE $\VisProb \gets $ \func{ComputeVisMap}($\mathcal{D}$, $\Obs$, $N$)
        \STATE $\mathcal{M}_{c} \gets $ \func{GetCoverMask}($\VisProb$, $\nu$, $C_m$)
        \STATE $\mathcal{R} \gets $ \func{GetCoverRegions}($\mathcal{M}_{c} $, $\Xi_{\textrm{min}}$)
        \STATE $\mathcal{R} \gets$ \func{SplitRegions}($\mathcal{R}$, $\Xi_{\textrm{max}}$)
        \STATE $V \gets $ \func{PlaceNodes}($C_m$, $\mathcal{R}$)
        \STATE $\mathcal{T}_{\textrm{feas}} \gets $ \nameref{alg:path-computation}($\VisProb$, $C_m$, $\mathcal{R}$, $V$, $\lambda_p$)
        \STATE $n_V \gets \vert \mathcal{R} \vert $
        \STATE $W_{\textrm{vis}} \gets \textbf{0}_{n_V \times n_V}$
        \STATE $E \gets \emptyset$
        \FORALL {$\{(v_j, v_k) :  \mathcal{T}\} \in \mathcal{T}_{\textrm{feas}}$}
            \STATE $W_{\textrm{vis}}(v_j, v_k) \gets $ \func{PathCost}($\VisProb$, $\mathcal{T}$)
            \STATE $E \gets E \cup (v_j, v_k)$
        \ENDFOR
        \LINECOMMENT{Overwatch Opportunities}
        \FOR {$v \in V$} 
            \STATE $\Obs_{\textrm{ow}}(v) \gets $ \func{UniformDistribution}($r_v$)
            \STATE $\mathcal{P}_{\textrm{ow}}(v) \gets $ \func{ComputeVisMap}($\mathcal{D}$, $\Obs_{\textrm{ow}}(v)$, $N$)
            \STATE $W_{\textrm{ow}}(v) \gets \textbf{0}_{n_V \times n_V}$
            \FORALL {$\{(v_j, v_k) :  \mathcal{T}\} \in \mathcal{T}_{\textrm{feas}}$}
                \STATE $W_{\textrm{ow}}(v)(v_j, v_k) \gets $ \func{PathCost}($\mathcal{P}_{\textrm{ow}}(v)$, $\mathcal{T}$)
            \ENDFOR
        \ENDFOR
    \STATE \textbf{return} $V, E, W_{\textrm{vis}}, \mathcal{W}_{\textrm{ow}}$
\end{algorithmic}
\end{algorithm}

\begin{algorithm}[thb]
\caption{\func{ComputePaths}}
\label{alg:path-computation}
\footnotesize
\begin{algorithmic}[1]
	\renewcommand{\algorithmicrequire}{\textbf{Input:}}
	\renewcommand{\algorithmicensure}{\textbf{Output:}}
	\renewcommand{\COMMENT}[2][.5\linewidth]{%
	  \leavevmode\hfill\makebox[#1][l]{//~#2}}
		\REQUIRE~~
		\begin{tabbing}
		\hspace*{4.5em} \= \kill % set the tabbings
			$\VisProb$ \> % Distance normalized 
            Visibility map\\ 
			$C_m$ \> % Obstacle map
            Obstacle map \\ 
            $\mathcal{R}$ \> Cover regions\\
            $V$ \> Graph nodes\\
            $\lambda_p$ \> Visibility cost weight
		\end{tabbing}
		\ENSURE~~
		\begin{tabbing}
		\hspace*{4.5em} \= \kill % set the tabbings
			$\mathcal{T}_{\textrm{feas}}$ \> Map: $(v_j, v_k) \mapsto \mathcal{T}$ \\
		\end{tabbing}
        \STATE $\mathcal{T}_{\textrm{feas}} \leftarrow $ \func{EmptyMap}()
        \STATE $is\_start \gets \{ v : \FALSE \; \forall v \in V \}$
        \STATE $is\_goal \gets \{ v : \FALSE \; \forall v \in V \}$
        \STATE $path\_to \gets \{ v : \textbf{null} \; \forall v \in V \} $
        \STATE $path\_from \gets \{ v : \textbf{null} \; \forall v \in V \}$
    	\FORALL {$v_j \in V$}
        \FORALL {$v_k \in V$}
            \IF {$v_j = v_k$} 
                \STATE \textbf{continue}
            \ENDIF
            \STATE $\mathcal{T} \gets $ \func{ComputeOnePath}($C_m$, $\VisProb$, $v_j$, $v_k$, $\lambda_p$)
            \IF { \func{PathIsRedundant}($\mathcal{R}$, $\mathcal{T}$, $v_j$, $v_k$) }
                \STATE $path\_to[v_k] \gets $ \func{MinCost}($\mathcal{T}$, $path\_to[v_k]$) \label{lst:line:min-cost-to}
                \STATE $path\_from[v_j] \gets $ \func{MinCost}($\mathcal{T}$, $path\_from[v_j]$)\label{lst:line:min-cost-from}
            \ELSE
                \STATE $\mathcal{T}_{\textrm{feas}}[(v_j, v_k)] \gets \mathcal{T}$
                \STATE $is\_start[v_j] \gets $ \TRUE
                \STATE $is\_goal[v_k] \gets $ \TRUE
            \ENDIF
        \ENDFOR
        \ENDFOR
        \STATE $\mathcal{T}_{\textrm{feas}} \gets $ \func{ReconnectNodes}($is\_start$, $is\_goal$, \\
        \quad\quad\quad\quad\quad\quad\quad\quad\quad\quad\quad\quad$path\_from$, $path\_to$)
    \STATE \textbf{return} $\mathcal{T}_{\textrm{feas}}$
\end{algorithmic}
\end{algorithm}

\subsection{Constructing a Topological Graph}
\label{sec:graph_gen:topo}

Using Def. \ref{def:dtg} for a \ac{DTG}, the graph $G = (V, E, W_E)$ is composed of nodes $V$, edges $E$, and weights $W_E$ that are a function of the state of the robots. For the \ac{MCRP} problem, we decompose the edge weights $W_E$ into two components: visibility edge weights $W_{\textrm{vis}}$ and overwatch edge weights $\mathcal{W}_{\textrm{ow}}$. 
We define $W_{\textrm{vis}}$ to be a weighted adjacency matrix and $\mathcal{W}_{\textrm{ow}}$ is a set of weighted adjacency matrices for each node, $\mathcal{W}_{\textrm{ow}} = \{W_{\textrm{ow}}(v) : v \in V\}$, where element $W_{\textrm{ow}}(v)(j,k)$ denotes the overwatch that a team of robots positioned at node $v$ can provide for a team traversing edge $(j,k)$.
Algorithm~\ref{alg:graph_construction} presents our method for constructing the components of the topological graph. This algorithm references the function \func{ComputePaths} defined in Algorithm~\ref{alg:path-computation}. The following subsections describe the key steps (represented in both algorithms as functions) in more detail.

\subsubsection{\func{ComputeVisMap}}
The visibility map computation uses an efficient viewshed algorithm \cite{Wang2000GeneratingVW} to generate $N$ viewshed maps sampled from the observer distribution $\Obs$. Equations (\ref{eq:visibility_prob}) and (\ref{eq:dist_weighted_visibility}) are used to compute the probability of detection for each cell in the map. These calculations form our visibility map $\VisProb$. 

\subsubsection{\func{GetCoverMask}}

The cover calculation creates a mask of all cells which have low visibility and are traversable.
\begin{equation}
    \mathcal{M}_{c} (x,y) = \begin{cases}
        1 & \text{if } \mathcal{P}_{\textrm{vis}}(x,y) < \nu ~ \text{and} ~ C_m(x,y) = 0 \\
        0 & \text{otherwise} . \\
    \end{cases}
\end{equation}

\subsubsection{\func{GetCoverRegions}}
We compute cover regions, $\mathcal{R} = \{ r \in \func{ConnComp}(\mathcal{M}_{c} ) : \func{area}(r) > \Xi_{\textrm{min}} \}$, by finding all of the discrete, connected components of the cover mask whose areas exceed a threshold value, $\Xi_{\textrm{min}}$. For this, we use functions labeled \func{ConnComp} and \func{area}, respectively.

\begin{figure}[tbh]
    % \vspace*{-3mm}
    \centering
	\begin{subfigure}[t]{0.48\columnwidth}
		\centering
		\includegraphics[width=\textwidth]{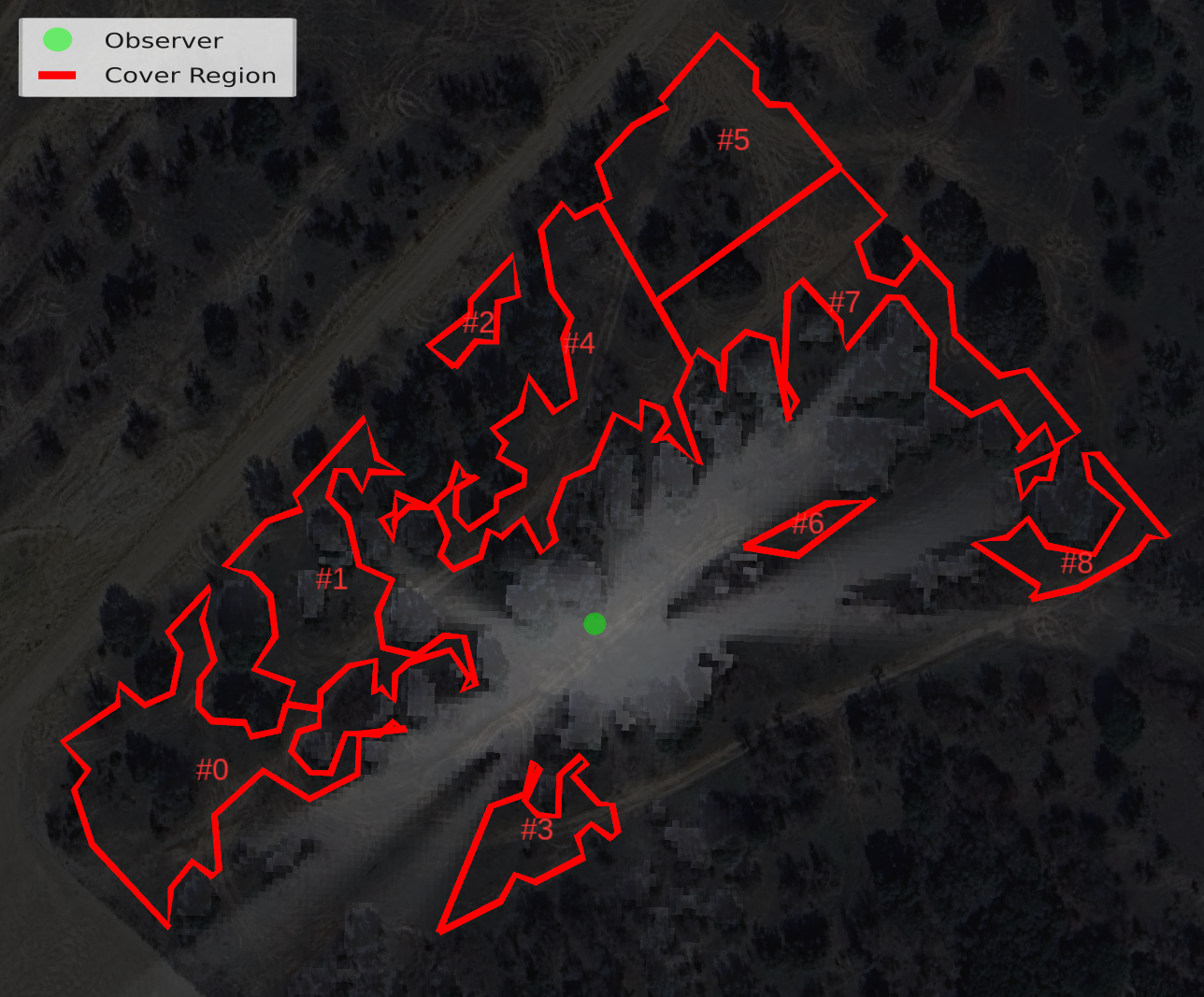}
		\captionsetup{font=scriptsize}
        \caption{Cover regions on visibility map}
        \label{fig:vis_map}
	\end{subfigure}
	\begin{subfigure}[t]{0.48\columnwidth}
		\centering
		\includegraphics[width=\textwidth]{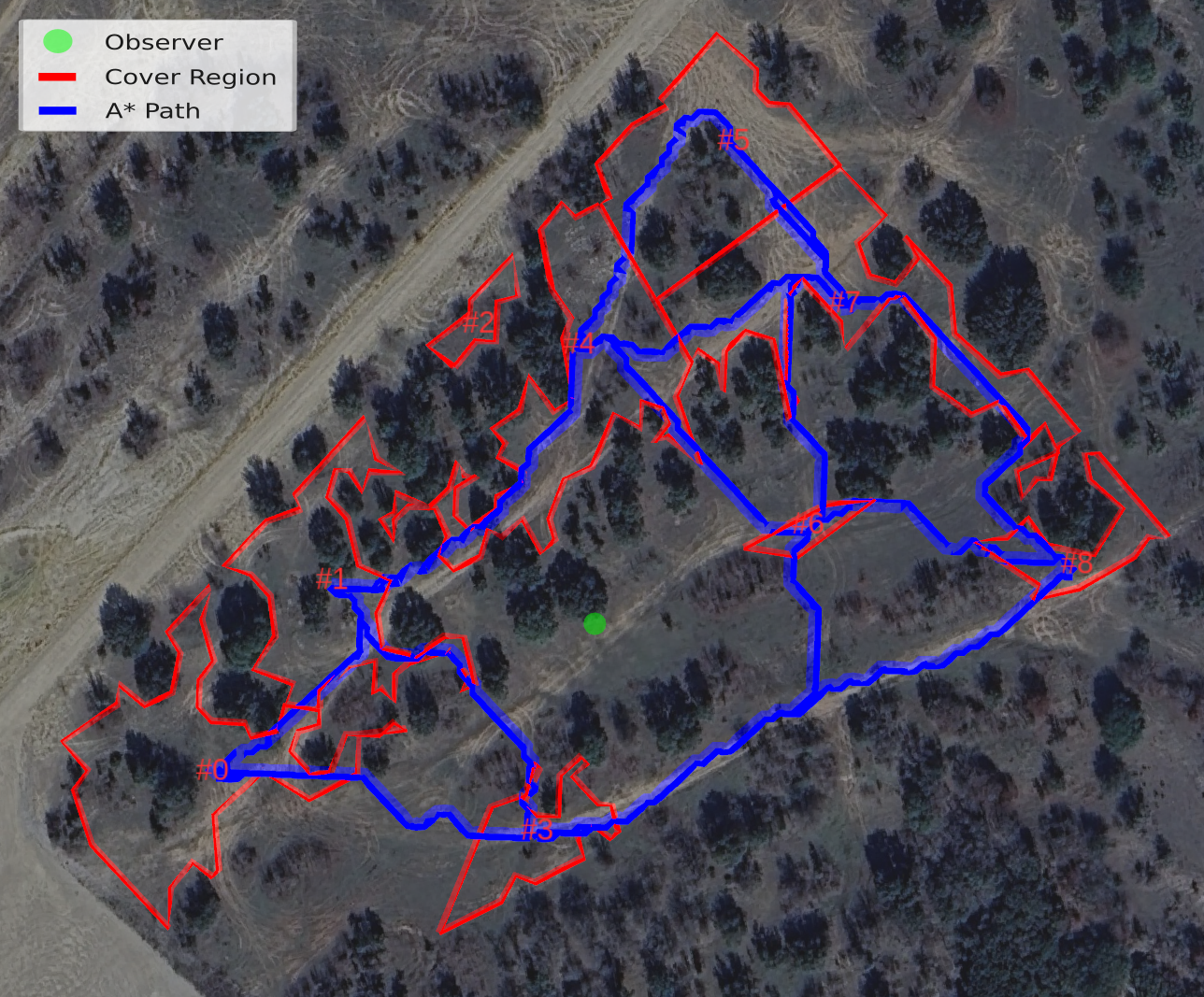}
		\captionsetup{font=scriptsize}
        \caption{Pruned A$^*$ paths between cover regions}
        \label{fig:astar_paths}
	\end{subfigure}
	\caption{Regions of cover and A$^*$ paths overlaid on satellite imagery of the environment: cover regions are outlined in red and node numbers are located within the regions near the centroid of each region. }
    \vspace*{-4mm}
\end{figure} % regions and a*

\subsubsection{\func{SplitRegions}}
For some maps and observer positions, the cover regions can be large and nonconvex. Consequently, we enforce a maximum size, $\Xi_{\textrm{max}}$, using the algorithm from \cite{sumit2014polygonsplitting} to find the minimum-length cut that divides one region into two smaller ones. The result of this cut is 2 regions, where one region is of size $\Xi_{\textrm{max}}$ and the other is the remainder.  We perform these cuts repeatedly until no region is larger than $\Xi_{\textrm{max}}$. Other strategies for splitting regions could be used; empirically, we find that iterative minimum length cuts tend to produce reasonably compact regions.

\subsubsection{\func{PlaceNodes}}

For each region, we place a node at the closest traversable point in the region to its centroid. Since the regions can be nonconvex, the centroid is not guaranteed to be contained in the region. The following defines the mapping $M_v$ between regions and nodes: $V = \{\,\argmin_{p \in r}(\Vert p - \func{centroid}(r)\Vert_2) : r \in \mathcal{R}\,\} .$ Here, the function \func{centroid} finds the centroid of the region.
In Fig.~\ref{fig:vis_map}, we overlay outlines of the cover regions on an example environment. % with the visibility map on satellite imagery.  

\subsubsection{\func{ComputePaths}}
We present Algorithm~\ref{alg:path-computation} for computing the feasible paths, $\mathcal{T}_{\textrm{feas}}$. $\mathcal{T}_{\textrm{feas}}$ maps from nodes $(v_j, v_k)$ to a path $\mathcal{T}$ between those nodes. Empty maps are initialized with the function \func{EmptyMap}. The resulting paths are composed of spatial locations: $\mathcal{T}  = \{\tau_1, \dots, \tau_n\}$, $\tau_i = [x_i, y_i]^T$. 

\subsubsection{\func{ComputeOnePath}} \label{section:graph_pruning}
We use an optimal path planning algorithm to compute paths that minimize visibility between each pair of nodes in the graph. In this work, we use a standard A$^*$ search. 
We choose A$^*$ for its optimality and simplicity, but in principle any optimal path planner capable of planning over costmaps could be used. 
We define the heuristic cost, $h(\tau)$, as the Euclidean distance from the current position to the goal. The function $g(\tau)$ denotes the cost to follow the optimal path from the start position to $\tau$. The successor cost, $c(\tau, \tau')$, is the cost to go from a cell in the map $\tau$ to adjacent cells $\tau'$. Finally, $n(\tau)$ is the visibility cost of traversing a single cell, which we will define in (\ref{eq:single-cell-log-odds}).
\begin{align}
    & c(\tau, \tau') = ||\tau - \tau'|| (1 + \lambda_p n(\tau')) \label{eq:successor_cost} \\
    & g(\tau') = g(\tau) + c(\tau, \tau').
\end{align}
In (\ref{eq:successor_cost}), $\lambda_p$ is a scaling factor to weight the contribution of the visibility cost. Since the true cost-to-go is equal to the heuristic cost plus a non-negative visibility term, the heuristic $h(\tau)$ is trivially admissible. 

\subsubsection{\func{PathIsRedundant}}
When computing the set of graph edges, 
we initially consider the topological graph to be fully-connected. However, many of these edges will overlap due to minimizing visibility or the routing of the environment, so we take an additional step to eliminate redundant edges. 
We define an edge $e = (v_j, v_k)$ as redundant if its associated path $\mathcal{T}_{v_j v_k}$ passes through a region other than the two regions it is intended to connect (i.e., $\mathcal{T}_{v_j v_k} \cap \big( \bigcup_{r \in \mathcal{R} \setminus \{r_j, r_k\}} r \big) \neq \varnothing$).

Whenever a redundant edge is removed, we update the minimum cost (with function \func{MinCost}) of the pruned paths originating at $r_s$ and ending at $r_g$ for later use in \func{ReconnectNodes}. 

\subsubsection{\func{ReconnectNodes}}
In rare circumstances, the pruning step may result in a node which is a sink (i.e., not the start of any path), a source (i.e., not the goal of any path), or fully disconnected. 
In these cases, we restore the lowest cost edges needed to maintain a directed graph with no sources or sinks. If there is not a viable route to a particular node (e.g., due to obstacles in the environment), that node will remain disconnected and will not be used in the graph planning process. An example of the pruned paths is shown in Fig.~\ref{fig:astar_paths}. 

\subsubsection{\func{PathCost}}
When traversing a path $\mathcal{T}$ through the environment, we consider the total probability of non-detection, $p_{nd}$, as the product of the probabilities of non-detection at each cell (assuming independence). We use a general $\mathcal{P}$ to represent a visibility map $\VisProb$ or overwatch map $\mathcal{P}_{\textrm{ow}}(v)$ (which we will define in Sec.~\ref{sec:overwatch}).
\begin{align}
    p_{nd} = \prod_{\tau \in \mathcal{T}}(1 - \mathcal{P}(\tau)).
\end{align}

Taking the negative $\log$ of this quantity, we can compute a negative $\log$ non-detection probability by summing the contributions of each cell along the path.
\begin{align} 
   -\log(p_{nd}) = \sum_{\tau \in \mathcal{T}}-\log(1 - \mathcal{P}(\tau)).
   \label{eq:log_odds}
\end{align}

As the probability of detection approaches one, the cost to traverse a cell approaches infinity. In order to avoid infinite values in the optimization problem, we bound the maximum probability of detection at each cell to $1 - \epsilon$ (for a small positive~$\epsilon$). So the cost to traverse a single cell is the negative-log of the probability of non-detection.
\begin{align} 
   n(\tau) = -\log(\max(1 - \mathcal{P}(\tau), \epsilon)).
   \label{eq:single-cell-log-odds}
\end{align}

In the final step of the path cost calculation, we compute the weight for each edge $e = (v_j, v_k)$ that remained after the pruning step by considering the optimal path $\mathcal{T}_{v_j v_k}$. 
\begin{align}
    W(v_j, v_k) = \sum_{\tau \in \mathcal{T}_{v_j v_k}}n(\tau) .\label{eq:visibility_score}
\end{align} 

\subsubsection{Overwatch Opportunities} \label{sec:overwatch}
\begin{figure}[tbh]
    % \vspace*{-3mm}
    \centering
	\begin{subfigure}[t]{0.48\columnwidth}
		\centering
		\includegraphics[width=\textwidth]{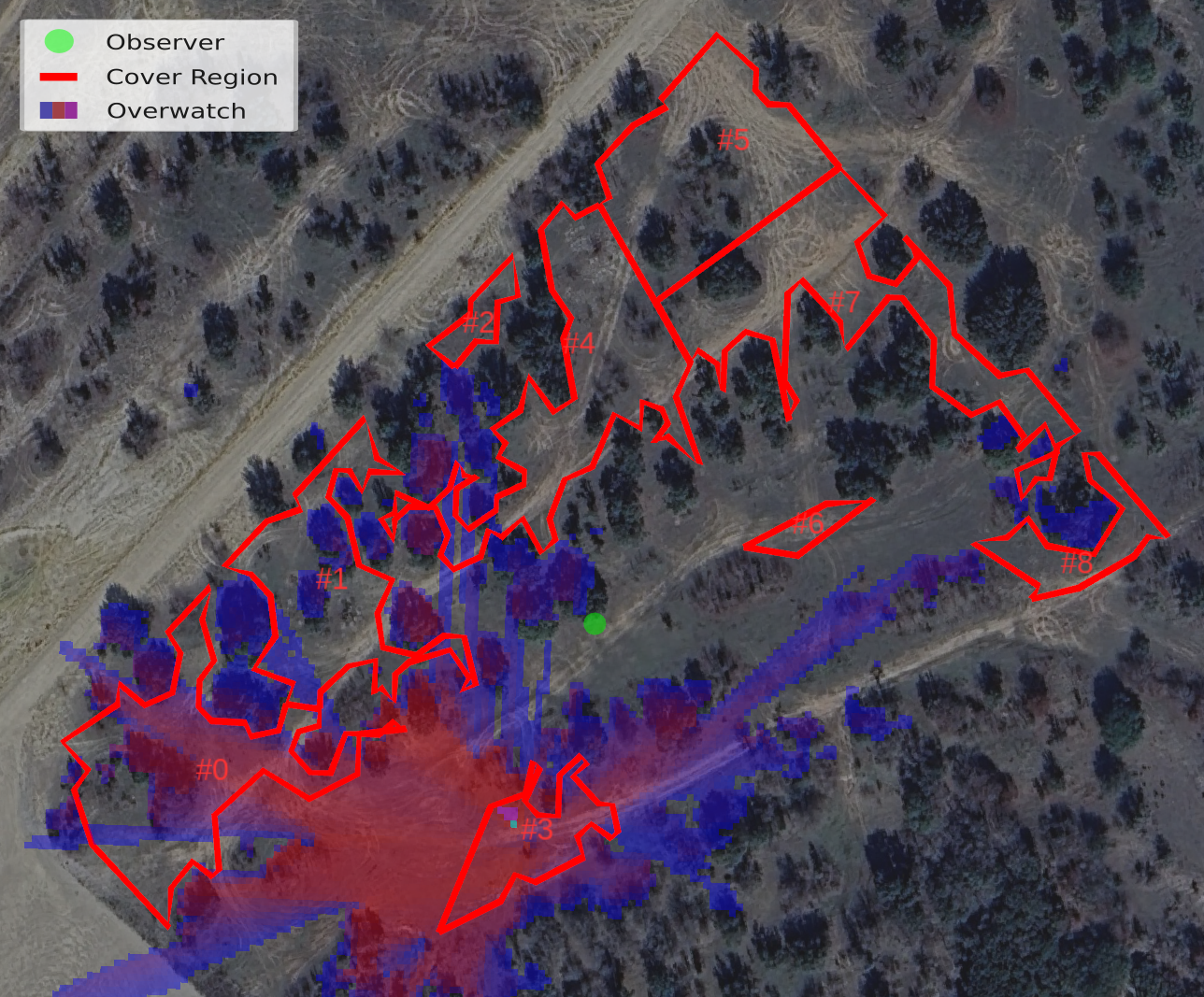}
		\captionsetup{font=scriptsize}
        \caption{Node 3 Overwatch Map}
        \label{fig:overwatch}
	\end{subfigure}
	\begin{subfigure}[t]{0.48\columnwidth}
		\centering
		\includegraphics[width=\textwidth]{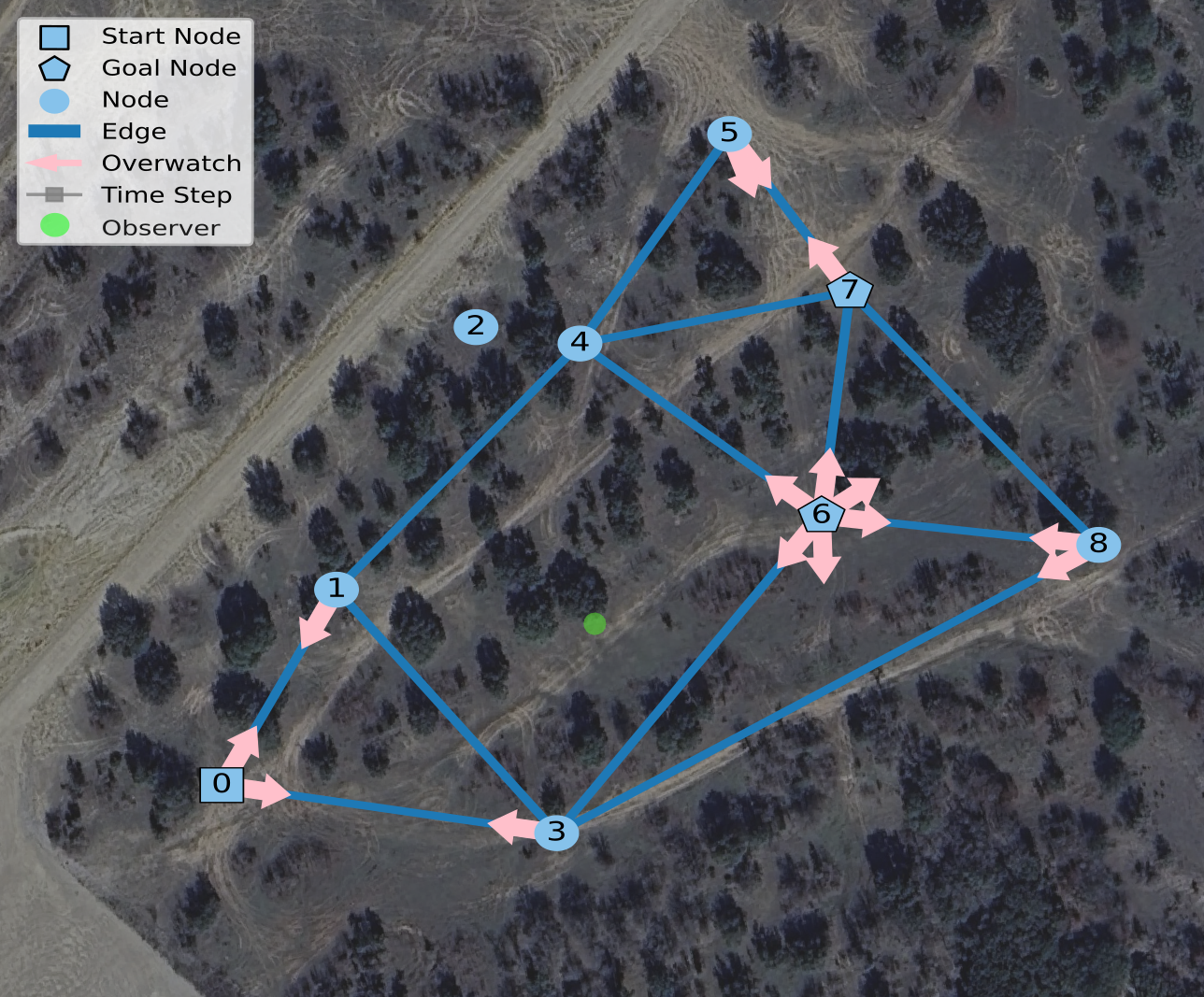}
		\captionsetup{font=scriptsize}
        \caption{Schematic Topological Graph}
        \label{fig:graph}
	\end{subfigure}
	\caption{Example overwatch map for node 3 in (a) and resulting schematic topological graph with overwatch opportunities indicated by arrows pointing from the overwatch node to edges that can be observed in (b). }
    \vspace*{-4mm}
\end{figure}

The overwatch weighted adjacency matrix, $W_{\textrm{ow}}(v)$, represents the benefit of overwatch from node $v$. Computing this matrix is analogous to the visibility map calculation. Instead of sampling from the observer distribution, we sample uniformly from the cover region associated with node $v$  (i.e., $r_v$). We define the uniform distribution (with function \func{UniformDistribution}) associated with $r_v$ as $\Obs_{\textrm{ow}}(v)$. We apply equations (\ref{eq:visibility_prob}) and (\ref{eq:dist_weighted_visibility}) to produce a visibility map for an observer located at node $v$, which we call an \textit{overwatch map}, $\mathcal{P}_{\textrm{ow}}(v)$. The overwatch edge weights, $W_{\textrm{ow}}(v)$, are computed using the \func{PathCost} function to evaluate the visibility from the node to each edge in the graph. 
The overwatch edge weight scales with how visible a particular edge is from the overwatch node and the distance along the edge that is visible. Thus, higher edge weights correspond to more advantageous overwatch opportunities.
An example overwatch map is shown in Fig.~\ref{fig:overwatch}.

 % regions and a*

\subsection{Topological Graph Refinement}
\label{sec:env_seg_constr}

We graphically depict the information embedded in our topological graph through the schematic representation shown in Fig.~\ref{fig:graph}. The graph includes nodes for each cover region, edges representing the pruned A$^*$ paths, and overwatch opportunities from the overwatch maps, denoted by arrows from the overwatch node to the edge that can be monitored. 
Although our formulation considers a directed graph, we construct graphs with both edge directions; thus, our schematics are shown with undirected edges.

When transforming the environment segmentation into this schematic representation, we use heuristics to refine the graph based on the robots' capabilities and the scenario. In particular, we set a maximum distance that can be traversed and remove edges from $E$ that do not meet this threshold. 

We add overwatch opportunities from $\mathcal{W}_{\textrm{ow}}$ to our graph after applying scenario-based heuristics. 
First, we impose a maximum overwatch distance (e.g., from sensor limits) by checking distances to edge endpoints. We then add a scale factor to the overwatch weights based on the priorities of the scenario (e.g., if overwatch is more or less valuable). 
Finally, we only incorporate overwatch opportunities when the cost reduction exceeds a prescribed fraction of the edge weight, and we bound larger reductions to ensure traversal always incurs nonzero cost 
(e.g., overwatch could provide between a 40\% and 90\% cost reduction). 
These thresholds and scaling factors are scenario-dependent and we demonstrate different values for these parameters in our analysis.

%%%%%%%%%%%%%%%%%%%%%%%%%%%%%%%%%%%%%%%%%%%%%%%%%%%%%%%%%%%%%%%%%%%%%%%%%%%%%%%% 
\vspace*{4mm}
\section{Graph Planner}
\label{sec:mip_approach}

To generate an optimal plan for a team of robots coordinating on a graph, such as the graph defined in Sec.~\ref{sec:graph_generation}, we developed a \ac{MIP} approach to solve an instance of the problem presented in (\ref{eq:dtg_optimal}). In this section, we first define our key formulation innovations in Sec.~\ref{sec:form_innovations}. We then detail how we utilize each innovation to formulate the specific cost function and constraints for the \ac{MCRP} in Sections~\ref{sec:cost_function} and \ref{sec:constraints}, respectively.
Ultimately, our derivation results in a statement of our optimization problem in Sec.~\ref{sec:optimization_problem} and a discussion of our formulation considerations in Sec.~\ref{sec:mip_considerations}.

\subsection{Key \ac{MILP} Formulation Innovations}
\label{sec:form_innovations}

A key contribution of this work is our computationally-efficient \ac{MILP} optimization problem. 
We apply three core innovations to preserve linearity in our \ac{MILP} formulation and significantly reduce the computation time to find an optimal solution to the problem: our integer decision variable formulation, utilization of indicator variables, and utilization of methods for formulating convex cost functions.

\subsubsection{Integer Decision Variable Formulation}
\label{sec:int_decision_vars}

Common formulations for multi-agent problems consider all possible states of robots at locations in the graph \cite{Sharon2015, Limbu2023}. For $n_L$ locations, this is $n_L^{n_A}$ possible states. In a \ac{MIP} formulation, enumerating these states as binary variables would be computationally intractable. Another common formulation is to encode each robot's location at each time step \cite{Yu2016, surynek2022problem}. 
This enumerates as $n_T n_L n_A$ total variables which is also highly computationally intensive for large numbers of agents and locations. 

We drastically reduce the size of the decision space by instead considering the number of robots at each location at each time step with $n_T n_L$ integer variables. 
We then present methods for analyzing interactions between robots (e.g., sharing edges or providing overwatch) to maintain the same utility of the larger decision space, while eliminating the exponential growth. 
This formulation obfuscates the paths of each individual robot in the \ac{MILP} problem, though we detail in Sec.~\ref{sec:allocator} how they can be assigned afterwards. Additionally, in the formulation of the optimization problem, we ensure that the paths follow the topology of the graph through sequential flow constraints considering the movement of all robots.

With this formulation, we embed the multi-robot interactions directly in the edge weights of our \ac{DTG}. 
In particular, we consider benefits from moving in formations and as a team to directly scale with the number of robots on an edge; thus our choice of decision variable greatly simplifies these calculations. For interactions through overwatch opportunities, we define costs and constraints for each overwatch opportunity that depend on the number of robots at the overwatch nodes and related edges to reduce the cost of traversing those edges. If we instead enumerated variables for each robot, fully defining each of these possible interactions would significantly increase the problem's complexity. Our choice of decision variables is pivotal to keeping the problem computationally tractable in a complex decision space. 

\subsubsection{Indicator Variables}
\label{sec:binary_variables}

In some cases, adding a variable for whether a condition is met aids in simplifying an expression or ensuring linearity. For this purpose, we use indicator variables \cite{williams2013model}, which are linked to another variable in the problem. The following proposition defines our form for indicator variables:%
\begin{proposition}    
For $h \in [0, h^{\ulcorner}]$, an indicator variable $\alpha \in \{0,1\}$ can be expressed, with any constant $a \geq 0$, as follows: 
\begin{align}
    \alpha = \begin{cases}
        1,~~ h > 0 \\
        0,~~ h = 0
    \end{cases}
    ~~~
    \Leftrightarrow
    ~~~~~
    \min a \alpha~\text{s.t.}~\alpha \geq \frac{h}{h^{\ulcorner}}.
    \label{eq:indicator_variable}
\end{align}
\label{math:prop:indicators}
\end{proposition}
We can then use the newly created binary variable $\alpha$ in our cost functions and constraints. As long as $\alpha$ is used in contexts that will be minimized, the value of $\alpha$ will be strict to the piecewise expression in (\ref{eq:indicator_variable}). This is due to the formulation; $\alpha$ can only be $0$ or $1$, and the constraint follows that $\frac{h}{h^{\ulcorner}} > 0$ for $h>0$ and $\frac{h}{h^{\ulcorner}} = 0$ for $h=0$. For $\alpha$ to be minimized means that the associated cost function has the lowest value when $\alpha=0$ (compared to $\alpha=1$). Indicator variables of this form can aid in expressing complex equations more efficiently in our optimization problem.  

\subsubsection{Convex Cost Functions}
\label{sec:convex_cost_func}

We aim to express our costs and constraints convexly, since convexity can significantly reduce the computation time of an optimization program. 
Since we are using \ac{MIP}, we consider convexity when our integrality constraints are relaxed. For brevity, references to convexity of a function in this paper will assume this relaxation. 

To produce a convex and positively homogeneous function, we can form a \textit{perspective function}, as originally presented in \cite{tyrrell1970convex}. In particular, we utilize the following result from \cite{dacorogna2008role}: 
\begin{proposition}    
Let $f: \mathbb{R}^n \rightarrow [-\infty, \infty].$ Then, the perspective function $\tilde{f}$ 
is convex if and only if $f$ is convex.
\begin{align}
    \tilde{f}(x,y) = yf\Big(\frac{x}{y}\Big), \quad x \in \mathbb{R}^n,~y \in \mathbb{R}_{>0}.
\end{align}
\label{math:prop:perspective}
\end{proposition}

In this work, we formulate piecewise cost functions based on the different regimes of operating alone or with a team. Using perspective functions, we can derive \textit{convex} piecewise cost functions, which we will show in Sec.~\ref{sec:cf_traversing}. To incorporate our piecewise functions linearly in our problem formulation, we can express these functions through a linear cost term with linear constraints using Proposition~\ref{math:prop:convex}, which is commonly referred to as the epigraph form of the problem \cite{boyd2004convex}.
\begin{proposition}
If $f(x)$ is a convex piecewise linear function with constraints $c_1(x), \dots, c_n(x)$, then the following are equivalent for some auxiliary variable $\breve{f}$: 
\begin{align}
    \min f(x)
    ~
    \Leftrightarrow
    ~
    \min \breve{f}~\text{s.t.}~\breve{f} \geq c_1(x), \dots, \breve{f} \geq c_n(x).
\end{align}
\label{math:prop:convex}
\end{proposition}

\subsection{\acs{MIP} Cost Function}
\label{sec:cost_function}

We define the parameters that we use to formulate our \ac{MIP} problem in Table~\ref{tab:mip_parameters} and the decision variables in Table~\ref{tab:mip_decision_vars}. Furthermore, we specify the constraints for the lower bounds (LB) and upper bounds (UB) of our decision variables in Table~\ref{tab:mip_decision_vars}.
Using the information embedded in our \ac{DTG} and the innovations described above, we formulate a cost function specific to the \ac{MCRP} which minimizes the risk of traversing and total time to reach a set of target locations. The dynamic edge cost in our \ac{DTG}, $w_{e,t}$ in Def.~\ref{def:dtg}, is a function of two cost functions we will define in the following section: (i) a cost of traversing $C_{\bar{W}_{e,t}}$, which embeds a fixed cost to traverse an edge, $\bar{w}_e$, and teaming considerations, and (ii) a cost reduction from overwatch $C_{\Omega_{\scripto,t}}$ related to a benefit $\omega_\scripto$ from overwatch. In particular, $w_{e,t} = C_{\bar{W}_{e,t}} + C_{\Omega_{\scripto,t}}$.
The cost values $\bar{w}_e$ and $\omega_\scripto$ are elements of the graph adjacency matrices defined in Sec.~\ref{sec:graph_gen:topo} (i.e., $\bar{w}_e \in W_{\textrm{vis}}$ and $\omega_\scripto \in \mathcal{W}_{\textrm{ow}}$).

\begin{figure}[b]
    % \vspace*{-3mm}
	\centering
	\begin{subfigure}[t]{0.49\columnwidth}
		\centering
		\includegraphics[height=0.12\textheight]{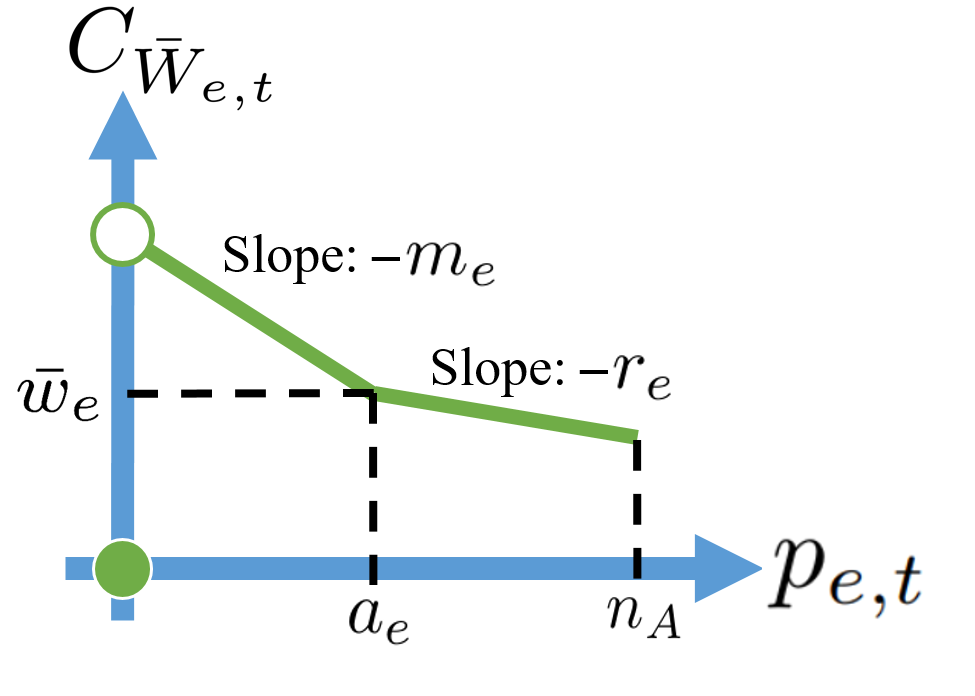}
        \captionsetup{font=scriptsize}
		\caption{Cost of traversing edge~$e$ at time~$t$ versus number of robots on the edge}
		\label{fig:cost_of_traversing}
	\end{subfigure}
	\hfill
	\begin{subfigure}[t]{0.49\columnwidth}
		\centering
		\includegraphics[height=0.12\textheight]{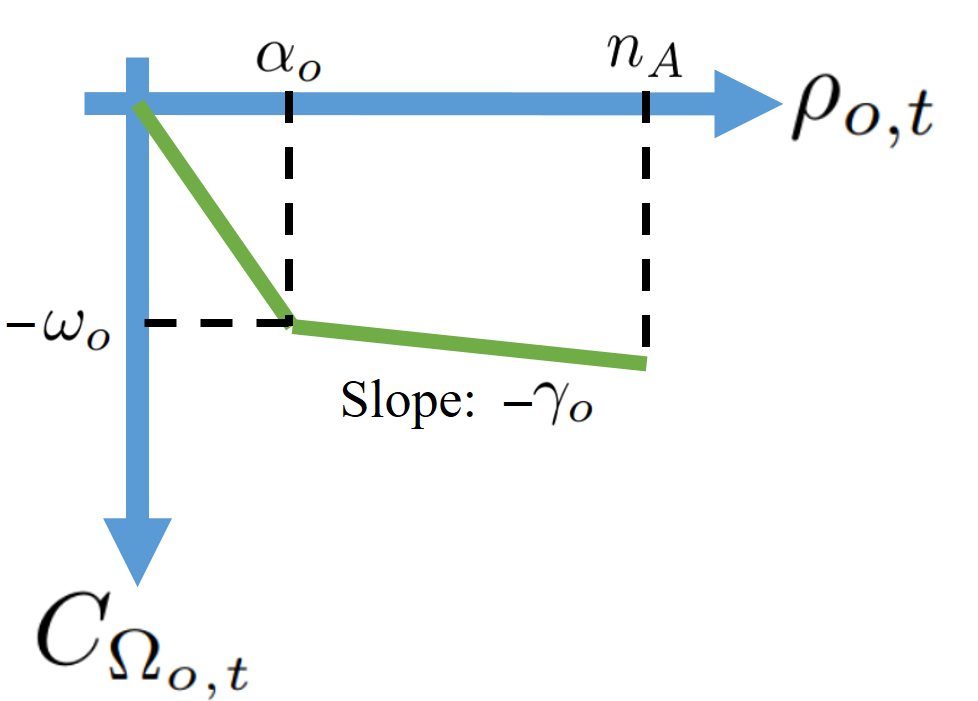}
        \captionsetup{font=scriptsize}
		\caption{Cost of overwatch opportunity~$\scripto$ at time~$t$ versus number of overwatch robots}
		\label{fig:cost_of_overwatch}
	\end{subfigure}%1.4cm 1.4cm 1cm 0.5
	\caption{Piecewise Linear Cost Functions}
	\label{fig:pwl_costs}
    % \vspace*{-4mm}
\end{figure}

\subsubsection{Cost of Traversing}
\label{sec:cf_traversing}

We formulate a cost of traversing with three main components: (a) edge cost, (b) vulnerability cost and (c) teaming rewards. 
Using the parameters from Table~\ref{tab:mip_parameters}, our base edge cost is the positive valued $\bar{w}_e$ for a team of robots to traverse edge $e$. 
For particularly vulnerable edges, it is desired for more agents to move together across the edge to enable moving in a formation for greater awareness. 
Thus, for each edge, to encode vulnerability, a minimum desired number of robots $a_e$ to traverse the edge is specified and an additional positive cost $m_e$ is incurred for each robot until the minimum is met.
Finally, we consider an incentive for further teaming through positive rewards $r_e$ for additional robots moving together over the minimum desired. 
These components comprise the piecewise linear cost of traversing an edge at a particular time, $C_{\bar{W}_{e,t}}$, which can be expressed as a function of the number of robots on the edge at that time, $p_{e, t}$. This cost function is depicted in Fig.~\ref{fig:cost_of_traversing}.
\begin{align}
C_{\bar{W}_{e,t}} = \begin{cases}
0, & p_{e, t} = 0 \\
\bar{w}_e + m_e (a_e - p_{e, t}), & 0 < p_{e, t} \leq a_e \\
\bar{w}_e - r_e(p_{e, t} - a_e), & a_e < p_{e, t} \leq n_A.
\end{cases}
\label{eq:pwl_traversing}
\end{align}

\begin{table}[t]
	\centering
    % \vspace*{2mm}
	\caption{\acs{MIP} Parameters}
	\vspace*{-2mm}
    \label{tab:mip_parameters}
	\begin{center}
		\renewcommand{\arraystretch}{1.3}
		\begin{tabular}{ c | c | p{4.9cm} }
			%	\hline
			%	\multicolumn{6}{|c|}{Agent} \\
			%	\hline
			%Variable Group Name & 
			\textbf{Category} & \textbf{Var} & \textbf{Description} \\
			\hline
			\hline
			\multirow{8}{4.5em}{Problem Size}
			& $n_A$ & Number of agents/robots \\
			& $n_T$ & Number of time steps in the time horizon \\ 
			& $n_\scriptO$ & Number of overwatch opportunities \\
			& $n_E$ & Number of directional edges \\
			& $n_V$ & Number of nodes/vertices \\
			& $n_L$ & Number of locations ($n_E + n_V$) \\
			& $n_B$ & Number of start/begin locations \\
			& $n_D$ & Number of goal/destination locations \\
   			\hline
			\multirow{8}{4.5em}{Scenario Variables} 
            & $E$ & Set of edges $e$ \\
            & $V$ & Set of nodes/vertices $v$ \\
            & $L$ & Set of locations $l$ consisting of edges and vertices, $E \cup V$ \\
            & $B$ & Set of start/begin locations $b$, $B \subseteq L$ \\
            & $D$ & Set of goal/destination locations $d$, $D \subseteq L$ \\
            & $\scriptO$ & Set of overwatch opportunities $\scripto$, $(v_i, e_j)$ where node $v_i$ can overwatch edge $e_j$ \\
			\hline
			\multirow{3}{4.5em}{Problem Parameters} 
			& $t$ & Time step from $1$ to $n_T$ \\
			& $n_{b}$ & Number of robots at start location $b \in B$ \\
			& $n_{d}$ & Number of robots at goal location $d \in D$ \\
			\hline
			\multirow{4}{4.5em}{Cost of Traversing} 
			& $\bar{w}_e$ & Fixed cost to traverse edge $e \in E$ \\
			& $a_e$ & Minimum desired number of robots on $e$ \\ 
			& $m_e$ & Additional cost for robots on $e$ before $a_e$ \\
			& $r_e$ & Cost reduction on $e$ for robots over $a_e$ \\
			\hline
			\multirow{3}{4.5em}{Cost of Overwatch} 
			& $\omega_\scripto$ & Benefit of full overwatch for $\scripto~\in~\scriptO$ \\
			& $\alpha_\scripto$ & Number of robots for full overwatch for $\scripto$ \\ 
			& $\gamma_\scripto$ & Reward for overwatch robots over $\alpha_\scripto$ for~$\scripto$ \\ 
		\end{tabular}
	\end{center}
    % \vspace*{-1mm}
    % \bigskip 
\end{table}
\begin{table}[t]
    \vspace*{-4mm}
	\centering
	\caption{\acs{MIP} Decision Variables (At Time $t$)}
    \vspace*{-2mm}
	\label{tab:mip_decision_vars}
	\begin{center}
		\renewcommand{\arraystretch}{1.3}
		\begin{tabular}{ c | c | c | c | p{3.65cm} }
			\textbf{Var} & \textbf{Type} & \textbf{LB} & \textbf{UB} & \textbf{Description} \\
			\hline
			\hline
			%Location Occupancy & 
			$p_{l,t}$ & Integer & 0 & $n_A$ & Number of robots at location $l$ \\ % at time $t$ \\ 
			$\phi_{e,t}$ & Binary & 0 & 1 & Whether robots are on edge $e$ \\ % at time $t$ \\
			%Time Used & 
			$\psi_{t}$ & Binary & 0 & 1 & Whether robots have moved \\ % at time $t$ \\
			%Cost of Traversing & 
			$\breve{C}_{\bar{W}_{e,t}}$ & Cont. & 0 & $\infty$ & Cost of traversing edge $e$ \\ % at time $t$ \\
			%Cost of Overwatch & 
			$\breve{C}_{\Omega_{\scripto,t}}$ & Cont. & $-\infty$ & 0 & Cost of overwatch opportunity $\scripto$ \\ % at time $t$
			%	\hline
		\end{tabular}
	\end{center}
    \vspace*{-4mm}
\end{table}

To reduce computation time, we want a convex formulation of this cost function. 
The cost in (\ref{eq:pwl_traversing}) is not convex due to the zero point.
To restate this cost, we first add an indicator variable for whether an edge is used, $\phi_{e,t}$, using Proposition~\ref{math:prop:indicators}.
We then consider the function $f(p_{e,t}) = \bar{w}_e + m_e (a_e - p_{e, t})$, which is convex when we select $m_e \geq r_e$.
Using Proposition~\ref{math:prop:perspective}, we can state the perspective function of $f(p_{e,t})$.
\begin{align}
    \tilde{f}(p_{e,t}, \phi_{e,t}) = -m_e p_{e, t} + (\bar{w}_e + m_e a_e) \phi_{e,t}.
\end{align}
By Proposition~\ref{math:prop:perspective}, the perspective function $\tilde{f}(p_{e,t}, \phi_{e,t})$ is a convex function in $p_{e,t}$ and $\phi_{e,t}$.
Additionally, when $p_{e, t} = 0$ then $\phi_{e,t} = 0$ by definition, and $\tilde{f}(p_{e,t}, \phi_{e,t}) = 0$. Otherwise, for any $p_{e, t} > 0$ then $\phi_{e,t} = 1$ and $\tilde{f}(p_{e,t})$ is identical to the second case in (\ref{eq:pwl_traversing}). Thus, this new function combines the first two cases in (\ref{eq:pwl_traversing}) and results in a convex piecewise linear function as follows: 
\begin{align}
C_{\bar{W}_{e,t}} = \begin{cases}
-m_e p_{e, t} + (\bar{w}_e + m_e a_e) \phi_{e,t}, & 0 \leq p_{e, t} \leq a_e \\
\bar{w}_e - r_e(p_{e, t} - a_e), & a_e < p_{e, t} \leq n_A.
\end{cases}
\label{eq:pwl_traversing_convex}
\end{align}

Due to the convexity of this function,
we can equivalently express this cost in our optimization problem with a linear term in the cost function %(\ref{eq:cost_function}) 
that scales with our decision variable $\breve{C}_{\bar{W}_{e,t}}$ and two linear constraints, as described in Proposition~\ref{math:prop:convex}. 
\begin{align}
\breve{C}_{\bar{W}_{e,t}} &\geq - m_e p_{e, t} + (\bar{w}_e + m_e a_e) \phi_{e,t} \label{eq:cost:trav1} \\
\breve{C}_{\bar{W}_{e,t}} &\geq - r_e p_{e, t} + (\bar{w}_e + r_e a_e) \phi_{e,t}.\label{eq:cost:trav2} 
\end{align}

These constraints are for the two line segments seen in Fig.~\ref{fig:cost_of_traversing}. We utilize $\phi_{e,t}$ in (\ref{eq:cost:trav2}) as well, so that 
$\breve{C}_{\bar{W}_{e,t}}$ will go to zero when vehicles are not traversing edge $e$ at time $t$.
Since we are optimizing for minimum cost, $\breve{C}_{\bar{W}_{e,t}}$ will be tight to the piecewise linear cost function (\ref{eq:pwl_traversing_convex}).

\subsubsection{Cost of Overwatch}

When an edge can be monitored from a node, we consider this an overwatch opportunity. Having overwatch while traversing can offer a cost reduction proportional to the visibility of the edge from the overwatch node and the distance that can be monitored.
When robots are traversing a corresponding edge, any number of robots at an overwatch node results in overwatch, but more robots may provide a greater reward. 
Using the parameters from Table~\ref{tab:mip_parameters}, we specify a positive $\omega_\scripto$ as the benefit of overwatch opportunity~$\scripto$, $\alpha_\scripto$~is the number of robots needed for full overwatch (i.e., to receive the full reward), and a positive $\gamma_\scripto$ is a further reward for additional robots past $\alpha_\scripto$. 
We consider $\rho_{\scripto,t}$ to be the number of overwatch robots for opportunity $\scripto$ at time $t$. Thus, as depicted in Fig.~\ref{fig:cost_of_overwatch}, for a particular overwatch opportunity the ``cost'' is $C_{\Omega_{\scripto,t}}$. 
\begin{align}
C_{\Omega_{\scripto,t}} = \begin{cases}
-\frac{\omega_\scripto}{\alpha_\scripto} \rho_{\scripto,t}, & 0 \leq \rho_{\scripto,t} \leq \alpha_\scripto \\
-\omega_\scripto - \gamma_\scripto (\rho_{\scripto,t} - \alpha_\scripto), & \alpha_\scripto < \rho_{\scripto,t} \leq n_A.
\end{cases}
\label{eq:pwl_overwatch}
\end{align}

This ``cost'' will always be negative since it is rewarding overwatch. Similarly to the cost for traversing, since this piecewise linear cost is convex when integrality constraints are relaxed and $\omega_\scripto / \alpha_\scripto \geq \gamma_\scripto$, we can express the cost with a decision variable, $\breve{C}_{\Omega_{\scripto,t}}$, that enters linearly in our cost function 
with 
two linear constraints. 
\begin{align}
\breve{C}_{\Omega_{\scripto,t}} &\geq -\frac{\omega_\scripto}{\alpha_\scripto} \rho_{\scripto,t} \\
\breve{C}_{\Omega_{\scripto,t}} &\geq -\omega_\scripto - \gamma_\scripto (\rho_{\scripto,t} - \alpha_\scripto).
\label{eq:overwatch_constr_initial}
\end{align}

We utilize the variables $\rho_{\scripto,t}$ for ease of formulation, but
can restate these constraints without these additional variables to reduce computation time of the optimization problem. 
For a particular $\scripto \in \scriptO$, we consider the number of robots providing overwatch to be equal to the number of robots at the node, $v_i$, if there are robots on the edge, $e_j$.  
We set $\rho_{\scripto,t} = p_{v_i, t}$, resulting in the following constraints:
\begin{align}
\breve{C}_{\Omega_{\scripto,t}} &\geq - \frac{\omega_\scripto}{\alpha_\scripto} p_{v_i, t} \label{eq:cost:overwatch1} \\
\breve{C}_{\Omega_{\scripto,t}} &\geq - \omega_\scripto - \gamma_\scripto (p_{v_i, t} - \alpha_\scripto) .
\label{eq:cost:overwatch2}
\end{align}
To ensure there are robots traversing the corresponding edge~$e_j$, we add another constraint dependent on $p_{e_j, t}$. 
\begin{align}
\breve{C}_{\Omega_{\scripto,t}} \geq - \frac{\omega_\scripto}{\alpha_\scripto} n_A p_{e_j, t} .
\label{eq:cost:overwatch3}
\end{align}
This constraint 
has a steeper slope (due to scaling by~$n_A$) than
(\ref{eq:cost:overwatch1}) and (\ref{eq:cost:overwatch2})
when there are robots on the edge (i.e., when $p_{e_j, t}~>~0$).  
This ensures it is the least restrictive (which we can guarantee for convex (\ref{eq:pwl_overwatch})). 
When there are robots at the node and there are not robots on the edge (i.e., $p_{v_i, t}~>~0$ and $p_{e_j, t}~=~0$), 
(\ref{eq:cost:overwatch3})
is the most restrictive overwatch constraint. % when $p_{e_j, t}~=~0$, 
This forces the cost $\breve{C}_{\Omega_{\scripto,t}}$ to be zero, since there is not a benefit from overwatch when there are not robots traversing the edge. Similarly, when $p_{v_i} = 0$, $\breve{C}_{\Omega_{\scripto,t}} = 0$ since there are not robots providing overwatch. 

\subsubsection{Cost of Time}

The final cost we consider, $C_{T_t}$, is for minimizing the time to achieve the goal.
We formulate a cost that scales with time and is
multiplied by the binary decision variables $\psi_{t}$ that represent whether robots have moved at time $t$. 
Thus, this cost, defined for all time $t$, is an incentive to achieve the goal as quickly as possible. 
\begin{align}
C_{T_t} = t \psi_t.
\label{eq:cost:time}
\end{align}

\subsubsection{Overall Objective Function}

When we combine all of the cost terms, we express our overall objective function to minimize as follows: 
\begin{align}
C = \sum_{t = 1}^{n_T} \bigg( C_{T_t} + \sum_{e \in E} \breve{C}_{\bar{W}_{e,t}} + \sum_{\scripto \in \scriptO} \breve{C}_{\Omega_{\scripto,t}}\bigg). \label{eq:overall_cost}
\end{align}
Weights can be added to each term in (\ref{eq:overall_cost}) depending on the priority of a particular scenario (e.g., minimizing traversing cost versus minimizing time).

%%%%%%%%%%%%%%%%%%%%%%%%%%%%%%%%%%%%%%%%%%
\subsection{\acs{MIP} Constraints}
\label{sec:constraints}
%%%%%%%%%%%%%%%%%%%%%%%%%%%%%%%%%%%%%%%%%%

We add constraints to set support variables used in our cost functions and to restrict movement to the \ac{DTG}. 
These constraints assume that traversing an edge has cost and remaining at a node does not, such that remaining on an edge would not provide a benefit. 

\subsubsection{Edge Used Variables}

We utilize binary variables $\phi_{e,t}$, that track if an edge is used, in cases where cost is minimal when $\phi_{e,t}=0$. Following Proposition~\ref{math:prop:indicators}, we set  $\phi_{e,t}$ to $1$ if there are robots on edge $e$ at time $t$ and $0$ otherwise.
\begin{align}
\phi_{e,t} \geq \frac{1}{n_A} p_{e,t} .\label{eq:constr:edge_used}
\end{align}

\subsubsection{Time Tracking Variables}

We add binary variables, $\psi_t$, to track if there are robots on the edges of the graph at time $t$. 
We use the form in Proposition~\ref{math:prop:indicators} 
with the sum of the number of robots on edges at each time step.
This sum tracks whether robots are still moving since robots cannot go instantaneously between nodes.
Following Proposition~\ref{math:prop:indicators}, $\psi_t$ only contributes to increasing cost.
For each time step $t$, we bound $\psi_t$. 
\begin{align}
\psi_t \geq \frac{1}{n_A} \sum_{e \in E} p_{e,t} .\label{eq:constr:time_vars}
\end{align}

\subsubsection{Start Locations}

We add constraints for the start locations of each robot. For each start location $b \in B$, with $n_{b}$ robots at that location, we add the following constraint: 
\begin{align}
p_{b, 1} = n_{b} .\label{eq:constr:start}
\end{align}

\subsubsection{Goal Locations}

For each goal location, or destination, $d \in D$, we set the minimum number of robots required at that location, $n_{d}$ robots, through the following constraint:
\begin{align}
p_{d, n_T} \geq n_{d} .\label{eq:constr:goal}
\end{align}

\subsubsection{Maximum Robots}

We bound the total number of robots across all locations of the graph for each time $t$. 
\begin{align}
\sum_{l \in L} p_{l,t} = n_A .\label{eq:constr:max_robots}
\end{align}

\subsubsection{Sequential Flow}

We add a sequential flow constraint to ensure that movement across the graph is restricted to the structure of the graph.
In particular, for each node, the number of robots in the node and flowing into the node must be equal to the number of robots in the node and flowing out of the node in the next time step.
Thus, for each time $t \in [2, n_T]$ and node~$v_j$, we add the following constraint:
\begin{align}
\sum_{\substack{l_{ij} = (v_i, v_j) \in L}} p_{l_{ij},t-1} = \sum_{\substack{l_{ji} = (v_j, v_i) \in L}} p_{l_{ji},t}  .\label{eq:constr:sequential}
\end{align}
The first sum 
considers all locations of the form $l_{ij} = (v_i, v_j)$ and the second sum considers $l_{ji} = (v_j, v_i)$ for a fixed node $v_j$. 
Both sets of locations include $l_{jj}$ since all nodes have self-loops.  
This constraint allows robots to move from one edge to a node or to another edge without stopping at the node.

\subsubsection{Negative Edge Weights}

We add a constraint to ensure that edge weights do not become negative due to the reductions from overwatch opportunities. For each edge $e$ 
and time step $t$, we add the following constraint:
\begin{align}
     \breve{C}_{\bar{W}_{e,t}} + \sum_{\scripto=(v_i,e) \in \scriptO} \breve{C}_{\Omega_{\scripto,t}} \geq \phi_{e,t}.
     \label{eq:constr:neg_edge_weight}
\end{align}
When the edge is used ($\phi_{e,t} = 1$), this constraint ensures that the minimum cost associated with the edge is $1$. 
The cost is strict to this bound since we are minimizing costs. 

%%%%%%%%%%%%%%%%%%%%%%%%%%%%%%%%%%%%%%%%%%
\subsection{\acs{MIP} Optimization Problem}
\label{sec:optimization_problem}
%%%%%%%%%%%%%%%%%%%%%%%%%%%%%%%%%%%%%%%%%%

Combining our objective function and constraints from the previous sections, our overall \ac{MIP} optimization problem is expressed in Table~\ref{tab:optimization_problem}.
By employing our key formulation innovations from Sec.~\ref{sec:form_innovations},
we can use mixed-integer \textit{linear} programming (MILP) to solve our problem since we have a linear objective function with linear constraints. We use the Gurobi optimizer \cite{GurobiOptimization2023} for solving this problem to optimality.

\begin{table}[tbh]
\centering
    % \vspace*{-2mm}
	\caption{\acs{MIP} Optimization Problem}
    \vspace*{-7mm}
	\label{tab:optimization_problem}
	\begin{center}
		\renewcommand{\arraystretch}{2.0}
        \resizebox{\columnwidth}{!}{%
		\begin{tabular}{ p{0.01cm} p{5.54cm} p{1.31cm} | c }
			\multicolumn{3}{c|}{\textbf{Optimization Problem}} & \textbf{Eq.} \\
			\hline
			\hline
			\multicolumn{3}{l|}{$\min~\sum\limits_{t = 1}^{n_T} \bigg( C_{T_t} + \sum\limits_{e \in E} \breve{C}_{\bar{W}_{e,t}} + \sum\limits_{\scripto \in \scriptO} \breve{C}_{\Omega_{\scripto,t}}\bigg)$ subject to} & (\ref{eq:overall_cost})\\
            \hline
            \multirow{5}{*}{\rotatebox[origin=c]{90}{Cost Constraints~}} 
            & $\breve{C}_{\bar{W}_{e,t}} \geq - m_e p_{e, t} + (\bar{w}_e + m_e a_e) \phi_{e,t},$ &$\forall e, t$ & (\ref{eq:cost:trav1})\\
            & $\breve{C}_{\bar{W}_{e,t}} \geq - r_e p_{e, t} + (\bar{w}_e + r_e a_e) \phi_{e,t},$ &$\forall e, t$ & (\ref{eq:cost:trav2})\\
            & $\breve{C}_{\Omega_{\scripto,t}} \geq - \frac{\omega_\scripto}{\alpha_\scripto} p_{v_i, t},$ &$\forall \scripto, t $ & (\ref{eq:cost:overwatch1})\\
            & $\breve{C}_{\Omega_{\scripto,t}} \geq - \omega_\scripto - \gamma_\scripto (p_{v_i, t} - \alpha_\scripto),$ &$\forall \scripto, t $ & (\ref{eq:cost:overwatch2})\\
            & $\breve{C}_{\Omega_{\scripto,t}} \geq - \frac{\omega_\scripto}{\alpha_\scripto} n_A p_{e_j, t},$ &$\forall \scripto, t $ & (\ref{eq:cost:overwatch3})\\
            \hline
            \multirow{6}{*}{\rotatebox[origin=c]{90}{Constraints\quad\quad\quad}} 
            & $\phi_{e,t} \geq \frac{1}{n_A} p_{e,t},$ & $\forall e, t $ & (\ref{eq:constr:edge_used}) \\
            & $\psi_t \geq \frac{1}{n_A} \sum\limits_{e \in E} p_{e,t},$ &$\forall t $ & (\ref{eq:constr:time_vars}) \\
            & $p_{b, 1} = n_{b},$ &$\forall b $ & (\ref{eq:constr:start})\\
            & $p_{d, n_T} \geq n_{d},$ &$\forall d $ & (\ref{eq:constr:goal})\\
            & $\sum\limits_{l \in L} p_{l,t} = n_A,$ &$\forall t$ & (\ref{eq:constr:max_robots})\\
            & $\sum\limits_{l_{ij} = (v_i, v_j) \in L} p_{l_{ij},t-1} = \sum\limits_{l_{ji} = (v_j, v_i) \in L} p_{l_{ji},t},$ &$\forall v_j, \newline t \in [2, n_T]$ & (\ref{eq:constr:sequential}) \\
            & $\breve{C}_{\bar{W}_{e,t}} + \sum\limits_{\scripto=(v_i,e) \in \scriptO} \breve{C}_{\Omega_{\scripto,t}} \geq \phi_{e,t}$, &$\forall e, t$. & (\ref{eq:constr:neg_edge_weight}) \\
		\end{tabular}%
        }
	\end{center}
    \vspace*{-4mm}
\end{table}

%%%%%%%%%%%%%%%%%%%%%%%%%%%%%%%%%%%%%%%%%%
\subsection{\acs{MIP} Problem Formulation Considerations}
\label{sec:mip_considerations}
%%%%%%%%%%%%%%%%%%%%%%%%%%%%%%%%%%%%%%%%%%

We formulate our optimization problem assuming edges have cost and nodes do not, such that there is not a benefit to staying on an edge.
Additionally, we require that costs for not meeting a minimum number of robots on an edge are set to be greater than reductions for agents after that number (i.e., $m_e \geq r_e$), as noted in Sec.~\ref{sec:cf_traversing}. 
We assume a graph without negative cycles.
Overwatch for one edge can come from multiple nodes and robots at one node can overwatch robots traversing multiple edges, but the weight of the edge (resulting from the cost of traversing, benefit of overwatch, and cost reductions from vulnerability/teaming) cannot reduce to 0 or below. 
We enforce this property with the constraint in (\ref{eq:constr:neg_edge_weight}).
Intuitively, for an overwatch opportunity to be used, the cost reduction from overwatch needs to be more than the cost to get to the overwatch position. 

We solve problems with our \ac{MIP} formulation to optimality. However, while our \ac{MIP} solutions are always guaranteed to be optimal, the optimal solution is not guaranteed to be unique. Small problems with simple numbers can result in 
many equivalent cost solutions and cause computation time to increase. 
We demonstrate illustrative simulated scenarios with simple numbers where this is more of a consideration than in our hardware experiments where we utilize our graph generation approach from Sec.~\ref{sec:graph_generation}.

%%%%%%%%%%%%%%%%%%%%%%%%%%%%%%%%%%%%%%%%%%%%%%%%%%%%%%%%%%%%%%%%%%%%%%%%%%%%%%%%
\section{Mid-Level Planner}
\label{sec:mid_level}

Our mid-level planner serves to connect the graph planner solutions of (\ref{eq:dtg_optimal}) to the local multi-robot motion planning problem in Sec.~\ref{sec:local_planning} by assigning robots to edges and generating robot-specific mid-range plans.

%%%%%%%%%%%%%%%%%%%%%%%%%%%%%%%%%%%%%%%%%%
\subsection{Robot Route Allocator}
%%%%%%%%%%%%%%%%%%%%%%%%%%%%%%%%%%%%%%%%%%
\label{sec:allocator}
After generating an optimal plan from our optimization problem in Table~\ref{tab:optimization_problem}, we use the resulting decision variable $p_{l,t}$ in an assignment routine to allocate routes for each robot through the graph. 
We use each start location from our set $B$. At each time step $t$, each robot $i$ will move to states in the form $s_t^i = (v_j, v_k)$. Subsequent valid states are assigned from the Next Edge Action Set $A(s_t^i)$ defined in Def.~\ref{def:action_set}. 
All nodes have self loops (i.e., $v_j = v_k$ for nodes). 
We present Algorithm~\ref{alg:route_assignment} with the details of this assignment process. 
We define the following functions for our algorithm: \func{RemoveItem($B$)} to remove an item from set $B$ (and output that item) and \func{AddItem($m$)} to add item $m$ to an existing list.
Tracking the number of robots at each location rather than routes for each robot directly necessitates this post-processing step, however the paradigm shift to integer decision variables (as described in Sec.~\ref{sec:int_decision_vars}) enables a significant reduction in computational complexity by removing the dependence on the number of robots in the size of our decision space. 

\begin{algorithm}[thb]
\caption{Robot Route Allocator}
\label{alg:route_assignment}
\footnotesize
\begin{algorithmic}[1]
	\renewcommand{\algorithmicrequire}{\textbf{Input:}}
	\renewcommand{\algorithmicensure}{\textbf{Output:}}
	\renewcommand{\COMMENT}[2][.5\linewidth]{%
	  \leavevmode\hfill\makebox[#1][l]{//~#2}}
		\REQUIRE~~
		\begin{tabbing}
		\hspace*{1.5em} \= \kill % set the tabbings
			$p_{l,t}$ \> Number of robots at location $l$ at time $t,$\\ \> $~\forall~l\in L,~t \in [1, n_T]$\\ 
			$B$ \> Set of $n_A$ start locations \\
			$n_A$ \> Number of agents/robots \\
			$n_T$ \> Number of time steps 
		\end{tabbing}
		\ENSURE~~
		\begin{tabbing}
		\hspace*{1.5em} \= \kill % set the tabbings
			$M$ \> List of routes for each robot ($M^i ~\forall i \in [1, n_A]$) \\
		\end{tabbing}
	\FORALL {$i \in [1, n_A]$}
        \STATE %$M^i \leftarrow $ \func{NewList}(); \quad
        $b \leftarrow$ \func{RemoveItem}($B$) 
        \STATE $M^i \leftarrow$ \func{AddItem}($b$); \quad
        $p_{b, 1} \leftarrow p_{b, 1} - 1$; \quad
        $s^i_1 \leftarrow b$ 
        \FORALL {$t \in [2, n_T]$}
            \FORALL {$l \in L$}
                \IF {$p_{l,t} > 0$ \textbf{and} $l \in A(s^i_{t-1})$} 
                    % \IF {$l \in A(s^i_{t-1})$}
                        \STATE $M^i \leftarrow$ \func{AddItem}($l$); \quad
                        $p_{l, t} \leftarrow p_{l, t} - 1$; \quad
                        $s^i_t \leftarrow l$
                        \STATE \textbf{break}
                    % \ENDIF
                \ENDIF
            \ENDFOR
        \ENDFOR
    \ENDFOR
    \STATE \textbf{return} $M$
\end{algorithmic} 
\end{algorithm}

%%%%%%%%%%%%%%%%%%%%%%%%%%%%%%%%%%%%%%%%%%
\subsection{Mid-Range Plan Generation}
%%%%%%%%%%%%%%%%%%%%%%%%%%%%%%%%%%%%%%%%%%
\label{sec:mid_range_plans}
When a coalition of robots, $\mathcal{C}_e$ is assigned to traverse an edge in the graph, members are ordered alphabetically; the first robot serves as leader, while the others are sequentially assigned follower positions.
In this paper, we use a leader-follower line formation, where each robot follows its predecessor.
Each robot's route, from Algorithm \ref{alg:route_assignment}, maps to an edge-specific path $\mathcal{T}_{v_iv_j}$ that minimizes visibility to potential observers, which we computed with A$^*$ in Sec.~\ref{section:graph_pruning}. The mid-range plan for a robot, $\mathcal{T}^i$, 
begins at the closest point on $\mathcal{T}_{v_iv_j}$ to the robot's current position and truncates at a receding horizon goal. For the leader, this goal is based on its maximum speed and planning time horizon. For the follower, this goal is based on a set following distance from the planned goal of the robot preceding it in the formation.
Thus, the coalition's set of mid-range plans $\mathcal{T}^{\mathcal{C}_e}$ is a function of $\mathcal{T}_{v_iv_j}$ and $X_B^{\mathcal{C}_e}$.

%%%%%%%%%%%%%%%%%%%%%%%%%%%%%%%%%%%%%%%%%%%%%%%%%%%%%%%%%%%%%%%%%%%%%%%%%%%%%%%%
\section{Low-Level Planner}
\label{sec:low_level}
As described in Sec.~\ref{sec:problem_formulation}, we solve a local optimization problem of the form in (\ref{eq:basic_opt_problem}) to execute edge-specific multi-robot motion planning. Rather than solving the optimization problem over the joint state space, we leverage a distributed receding-horizon \ac{NMPC} control approach \cite{luis2019trajectory}, where coupled costs and constraints are approximated by robot team members sharing state and policy information from the prior planning interval. Given the set of mid-range plans $\mathcal{T}^{{C}_e}$ generated in  Sec.~\ref{sec:mid_range_plans}, we can also use a short, computationally tractable time-horizon and a receding-horizon goal $X^{\mathcal{C}_e}_D$ constructed from the set of terminal points in $\mathcal{T}^{{C}_e}$. Our approach uses only a state-cost $g_{x,e}$, and does not include an input cost $g_{u,e}$, or state or input constraints $c_{x_e}$ or $c_{u_e}$.

For the dynamics $f(x_t)$ in (\ref{eq:basic_opt_problem}), we use Dubins Car kinematics with bounded wheel velocity and acceleration:
\begin{align}
x_t &= \begin{bmatrix} p_t \ \theta_t \end{bmatrix}^T,\enspace u_t = \begin{bmatrix} v_t \ \alpha_t \end{bmatrix}^T,\enspace w_t = \begin{bmatrix} \frac{1}{r} & \frac{b}{2r} \\ \frac{1}{r} & -\frac{b}{2r} \end{bmatrix}u_t \nonumber\\ 
w^{\textrm{max}}_t &= \max(w_t, \frac{v_{\textrm{min}}}{r}, w_{t-1} - \frac{a_{\textrm{max}}}{r} \Delta t) \nonumber\\ 
w^{\textrm{clamp}}_t &= \min(w^{\textrm{max}}_t, \frac{v_{\textrm{max}}}{r}, w_{t-1} + \frac{a_{\textrm{max}}}{r} \Delta t) \\
x_{t+1} &= x_t + \begin{bmatrix} \cos(\theta_t) & 0  \\ \sin(\theta_t) & 0 \\ 0 & 1 \end{bmatrix} \begin{bmatrix} \frac{r}{2} & \frac{r}{2} \\ \frac{r}{b} & -\frac{r}{b} \end{bmatrix}w^{\textrm{clamp}}_t\Delta t. \nonumber
\end{align}
Here we define the state $x_t\in\mathbb{R}^3$, position $p_t\in\mathbb{R}^2$, orientation $\theta_t$, control $u_t\in\mathbb{R}^2$, linear velocity $v_t$, angular velocity $\alpha_t$, wheel radius $r$, wheel base $b$, wheel velocities $w_t\in\mathbb{R}^2$, minimum linear velocity $v_{\textrm{min}}$, maximum linear velocity $v_{\textrm{max}}$, maximum linear acceleration $a_{\textrm{max}}$, and time step $\Delta t$.

Since the cost functions from (\ref{eq:basic_opt_problem}) have been decoupled, for a given robot $i\in\mathcal{C}_e$, we can now write $g_{x}(X^{\mathcal{C}_e}_t)$ as $g_{x,e}(x^i_t)$. Dropping the $i$ for clarity, we design our cost function $g_{x,e}(x_t)$ to be a weighted sum of cost components to encourage navigating towards the goal (distance $g_\delta$, heading $g_h$, and pointing $g_p$), following the mid-range path (distance $g_{m\delta}$ and heading $g_{mh}$), avoiding untraversable terrain (terrain costmap $g_c$), and inter-robot collisions (trajectory collision $g_{tc}$).

To avoid obstacles, which may not exist in the global map, we leverage TerrainNet \cite{meng2023terrainnet} for local mapping. TerrainNet takes RGB-D images of the environment as input and produces semantic and elevation maps as output. 
We produce a local costmap, $C_{l}$, by assigning a cost, $[0, 255]$, to each semantic label, which correspond to the features of the environment (e.g., bush, rock, tree, dirt, etc.) that occupy the terrain.

This approach aids in traversing diverse terrain, but could be substituted for another local mapping approach.

The overall cost function is as follows:
\begin{align}
g_{x,e}(x_t) =&\ \gamma_{\delta} g_{\delta}(x_t) + \gamma_{h} g_{h}(x_t) + \gamma_{p} g_{p}(x_t) + \gamma_{m\delta} g_{m\delta}(x_t) \nonumber\\ &+ \gamma_{mh} g_{mh}(x_t) + \gamma_c g_{c}(x_t) + \gamma_{tc} g_{tc}(x_t).
\end{align}
In this cost function, $\gamma_{\delta}$, $\gamma_{h}$, $\gamma_{p}$, $\gamma_{m\delta}$, $\gamma_{mh}$, $\gamma_c$, and $\gamma_{tc}$ are heuristically selected cost weights.

We define the differences between the current state of the robot $x_t$ and the following: the goal state $x_D$, the mid-range path $x^m_t$, and the previous state $x_{t-1}$. Additionally, we define $\Delta x^i_{t^it}$, the difference between the current state of the robot and every state in the planned trajectory of each other robot in the same coalition, $x^i_{t^i}$. Here, $i\in\mathcal{C}^e$ is the robot index and $t^i$ is the time index along robot $i$'s trajectory.
\begin{align}
\begin{split}
    \Delta x^d_t =&\ x_D - x_t, \quad \Delta x^m_t = x^m_t - x_t \\
    \Delta x_t =&\ x_t - x_{t-1}, \quad \Delta x^i_{t^it} = x^i_{t^i} - x_t.
\end{split}
\end{align}

Finally, we can define the cost components. We use parameters for goal pointing radius $r_p$, 
the maximum mid-range path cost distance $\delta_m$,
the lethal costmap penalty $c$, the lethal costmap threshold $\ell$, the trajectory collision radius $r_t$, and the planning time horizon $T$. 
\begin{align}
    g_{\delta}(x_t) =&\ \|\Delta p^d_t\|  \nonumber\\
    g_{h}(x_t) =&\ \!\exp\!\left({\!\!\left(\!\!\frac{\Delta p^d_{t}}{2}\!\cdot\!\begin{bmatrix}\cos(\theta^d) \\ \sin(\theta^d)\end{bmatrix}\!\right)\!\!}^{2} \right)\! \cdot \min(\!-\!\cos(\Delta \theta^d_t), 0.9)  \nonumber\\
    g_{p}(x_t) =&\ \mathbb{I}_{\{\|\Delta p^d_t\| < r_p\}} \cdot \max(-\cos(\atantwo(\Delta p^d_t) - \theta^d), 0) \nonumber\\
    g_{m\delta}(x_t) =&\ \min\left(\frac{\|\Delta p^m_t\|}{\delta_m}, 1.0\right)^2 \\
    g_{mh}(x_t) =&\ \left(1 - \frac{\Delta p_t}{\|\Delta p_t\|} \cdot\begin{bmatrix}\cos(\theta^m) \\ \sin(\theta^m)\end{bmatrix}\right) \nonumber\\
    g_{c}(x_t) =&\ \begin{cases}
    1\mathrm{e}{10},& \text{if } C_{l}(p_t) \geq \ell  \text{ and } C_{l}(p_{t-1}) < \ell \\
    1\mathrm{e}{10},& \text{else if } C_{l}(p_t) \geq \ell  \text{ and } t = T \\
    c,              & \text{else if } C_{l}(p_t) \geq \ell  \text{ and } C_{l}(p_{t-1}) \geq \ell \\
    \frac{C_{l}(p_t)}{\ell} & \text{otherwise } \end{cases} \nonumber\\
    g_{tc}(x_t) =&\ \mathbb{I}_{\left\{\bigvee_{i, t^i} \left(\|\Delta p^i_{t^it}\|< r_t\right)\right\}} \cdot 1\mathrm{e}{10}. \nonumber
\end{align}
To optimize the objective function subject to the dynamics constraints, we use \ac{MPPI} \cite{Williams2016}, though this approach could be substituted for other methods for planning dynamically feasible paths on costmaps.

% For our experiments with Clearpath Warthogs, we used the following parameters: $r=0.279m$, $b=1.175m$, $v_{\textrm{min}}=-0.5\frac{m}{s}$, $v_{\textrm{max}}=2.0\frac{m}{s}$, $a_{\textrm{max}}=4.0\frac{m}{s^2}$, $r_p=1m$, $d_m=10m$, $c=30$, $\ell=200$, $\gamma_{d}=1$, $\gamma_{h}=0.1$, $\gamma_{p}=0.1$, $\gamma_{md}=5.0$, and $\gamma_{mh}=1.0$. For \ac{MPPI}, we used $\Delta t=\frac{1}{15}s$, $K=3000$ samples, time horizon $T=5s$, temperature $\lambda=0.01$ and sampling variance of $\Sigma=1.94$.

%%%%%%%%%%%%%%%%%%%%%%%%%%%%%%%%%%%%%%%%%%%%%%%%%%%%%%%%%%%%%%%%%%%%%%%%%%%%%%%%
\section{Results and Discussion}
We first evaluate our graph planning approach in simulated scenarios. We then demonstrate the full hierarchical planning architecture, \ac{STALC}, in autonomous hardware experiments. 
\subsection{Graph Planning Simulated Scenarios}
\subsubsection{Illustrative Example}

\begin{figure}
    % \vspace*{2mm}
	\centering
	\begin{subfigure}[t]{0.6\columnwidth}
		\centering
		%\fbox{}
		\includegraphics[width=\textwidth]{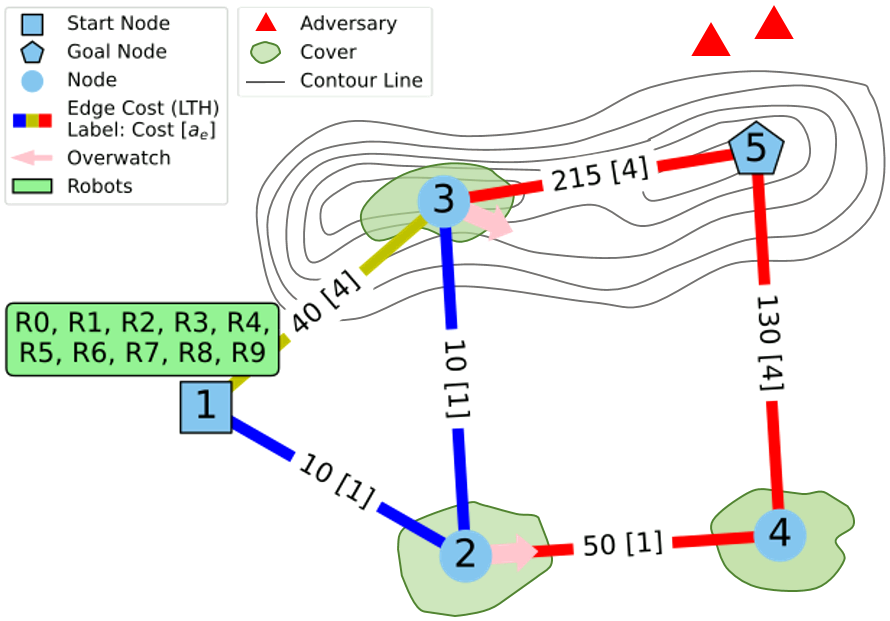}
		\caption{Time = 0, Total Cost = 0}
		\label{fig:toy_problem_solution_0}
	\end{subfigure}%
    
	%	\hfill
	\begin{subfigure}[t]{0.45\columnwidth}
		\centering
		\includegraphics[trim={1.7cm 1.9cm 0.9cm 0.3cm},clip, width=\textwidth]{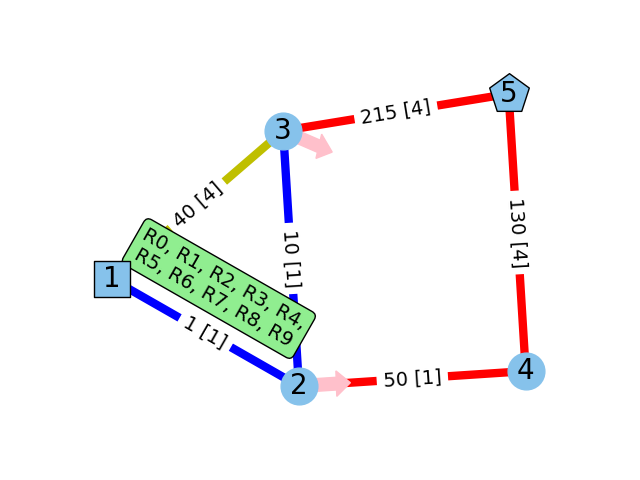}
		\caption{Time = 1, Total Cost = 11}
		\label{fig:toy_problem_solution_1}
	\end{subfigure}%
	%	\hfill
	\begin{subfigure}[t]{0.45\columnwidth}
		\centering
		\includegraphics[trim={1.7cm 1.9cm 0.9cm 0.3cm},clip, width=\textwidth]{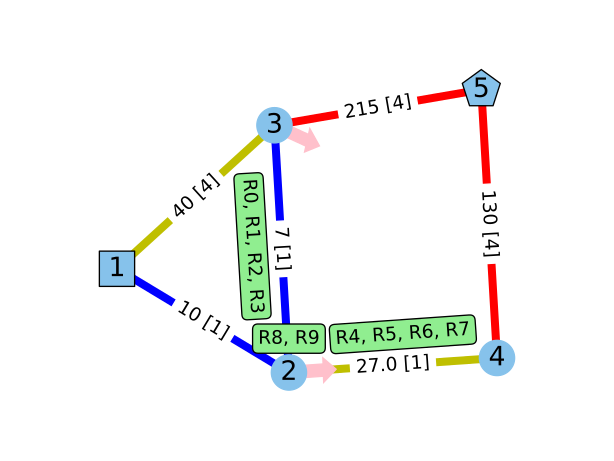}
		\caption{Time = 2, Total Cost = 65}
		\label{fig:toy_problem_solution_2}
	\end{subfigure}
	
	\begin{subfigure}[t]{0.45\columnwidth}
		\centering
		\includegraphics[trim={1.7cm 1.9cm 0.9cm 0.3cm},clip, width=\textwidth]{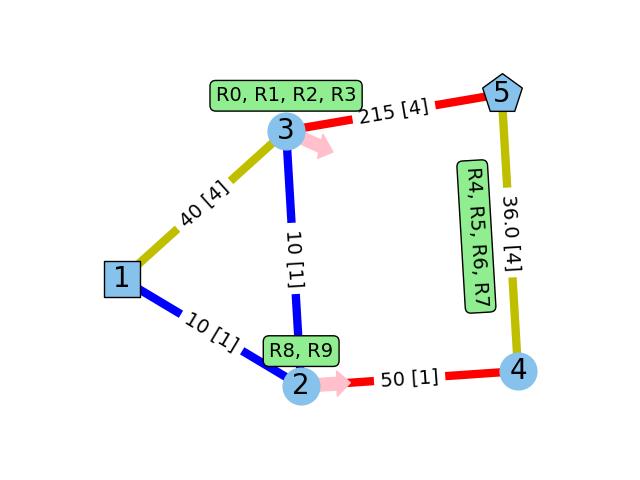}
		\caption{Time = 3, Total Cost = 131}
		\label{fig:toy_problem_solution_3}
	\end{subfigure}%1.4cm 1.4cm 1cm 0.5
	%	\hspace{.05\textwidth}%
	\begin{subfigure}[t]{0.45\columnwidth}
		\centering
		\includegraphics[trim={1.7cm 1.9cm 0.9cm 0.3cm},clip, width=\textwidth]{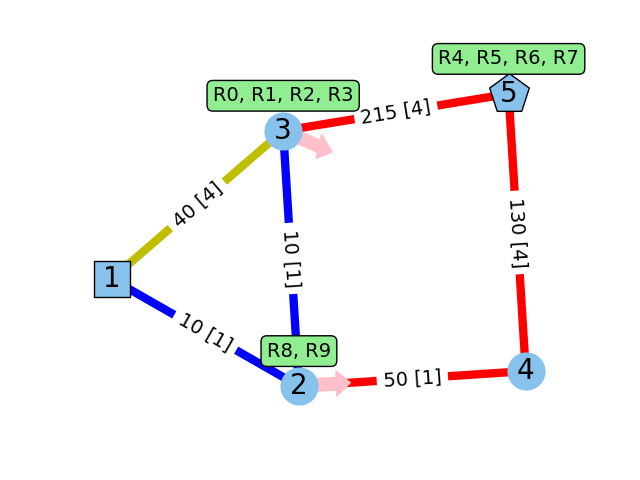}
		\caption{Time = 4, Total Cost = 131}
		\label{fig:toy_problem_solution_4}
	\end{subfigure}%
	\caption{\acs{MIP} problem solution to an illustrative MCRP, sketched in 
    % Fig.~\ref{fig:toy_problem_solution_0}. 
    (a).
    Ten robots start at node 1 with a goal of at least one robot reaching node 5 to observe the adversary units. The light green regions represent areas of cover. The color of the edges indicate the threat level for transitioning between nodes from low to high (LTH): blue, yellow, red. Edge (3,5) has the highest weight due to its visibility by the adversary. 
	In each subplot, the edge labels indicate the edge cost under the current conditions and show in brackets the desired number of robots for traversal of the edge, $a_e$, due to vulnerability. Overwatch opportunities, which are used in time steps 2 and 3, are shown with a pink arrow from the overwatch node pointing to the edge that can be monitored. At each time step, the position of each robot is shown.}
	\label{fig:toy_problem_solution}
    \vspace*{-4mm}
\end{figure}

Fig.~\ref{fig:toy_problem_solution} depicts a simple version of the MCRP in which 10 robots start at node 1 and at least one robot must reach node 5 while minimizing visibility. 
We generate a solution to this problem using our graph planner, as shown in Fig.~\ref{fig:toy_problem_solution}. 
In Fig.~\ref{fig:toy_problem_solution_0}, the edge costs are shown as the maximum cost for one robot to traverse without reductions from overwatch, teaming, or vulnerability. The number of robots desired for each edge is shown in brackets ([]). The color of the edge indicates threat level blue (lowest), yellow, or red (highest). 
To encourage teaming, each additional robot on an edge reduces the cost by 1. Both directions of edges (1,3), (3,5), and (4,5) are considered vulnerable and at least four robots are desired; the cost reduction for each robot up to four is 10. Overwatch opportunities are indicated by the pink arrows. Overwatch can be provided from node 2 for both directions of edge (2,4) and node 3 for both directions of edge (4,5), which can reduce the cost by up to 20 or 60, respectively, when two robots are providing overwatch. 
Each additional robot providing overwatch after the first two would reduce the cost by 2. The dynamic costs incurred on each edge, which are based on the state of the robot team, are updated in the graphs in Fig.~\ref{fig:toy_problem_solution}. 
In this example, we scale the time cost by 10 to encourage reaching the goal in minimal time. The total accumulated cost is tracked in the captions. 
In this solution, we see the robots break into three teams to work together for a subset of the robots to safely traverse to the target node. 
By solving our problem to optimality, we are guaranteed to have minimized the total cost. However, this depends greatly on the weights/cost assigned to the problem and the prioritization of minimizing traversing cost versus minimizing time.

\subsubsection{Bounding Overwatch Example}

\begin{figure}[t]
	% \vspace*{-3mm}
	\centering
	\includegraphics[trim={0cm 0.3cm 0cm 0.5cm},clip, width=0.8\columnwidth]{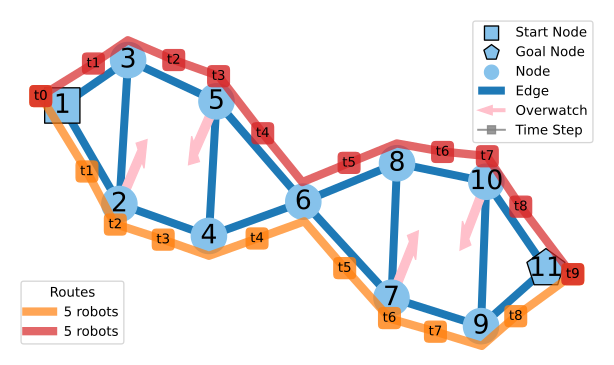}
    \vspace*{-2mm}
	\caption{Example bounding overwatch solution as all robots move from node 1 to 11. Robot coalition routes are displayed with the position of each coalition at each time step denoted by square markers along the routes. Time steps are labeled sequentially from $t0$ to $t9$. The overwatch opportunities, indicated by the pink arrows, are met at time steps 2, 3, 6, and 7. The two coalitions alternate providing overwatch as they move through the environment. 
	}
	\label{fig:problem4}
	\vspace*{-4mm}
\end{figure}

As a demonstration on a graph that particularly lends itself to the ``bounding overwatch'' paradigm, where robots alternate providing overwatch, Fig.~\ref{fig:problem4} shows a solution to our \ac{MILP} problem for the graph shown in blue. The robots start at node 1 and need to reach node 11. 
The solution yields two robot teams alternating traversing and providing overwatch. 
In this figure and subsequent figures, we depict our solution to the problem through routes for each coalition of robots. 
Along these routes, we denote each time step with square markers and text indicating the sequential step in the plan that the coalition traverses the corresponding edge or stops at the corresponding node. These markers enable correlating the relative positions of the coalitions' to identify when coalitions are moving together or overwatch is occurring (i.e., one coalition is at an overwatch node while the other one traverses the corresponding edge at the same time step). 

\begin{figure}[t]
	% \vspace*{1.5mm}
	\centering
	\includegraphics[trim={1.9cm 0.4cm 1.9cm 0cm},clip, width=0.7\columnwidth]{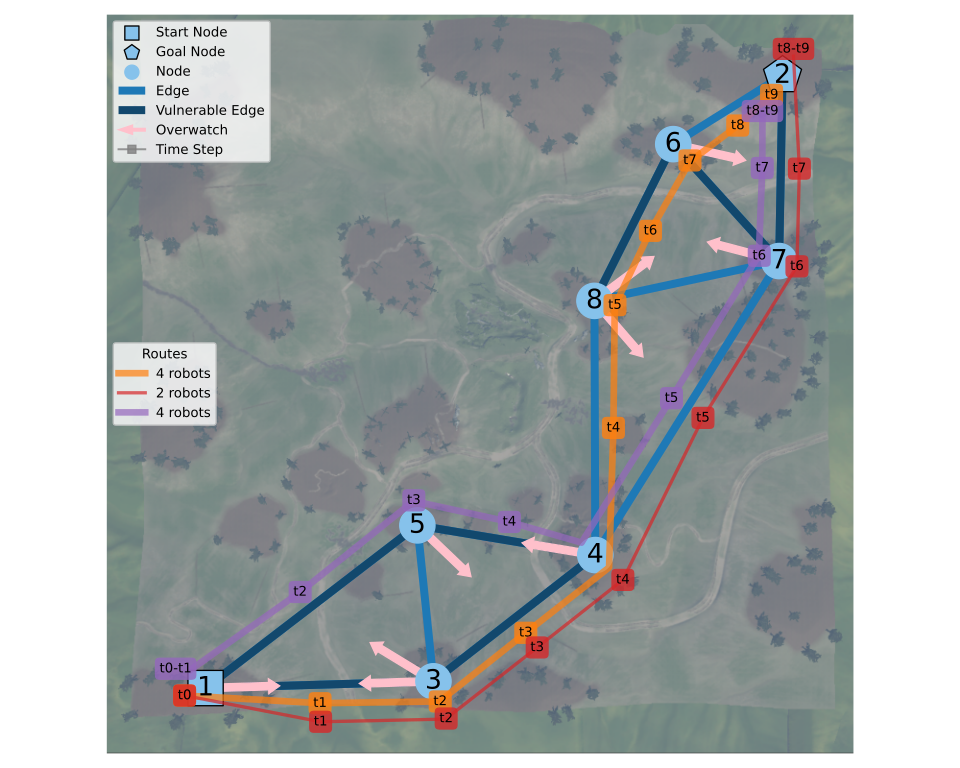}
    \vspace*{-2mm}
	\caption{Aerial map of a meadow environment showing a sample scenario. Nodes are in regions of cover. Vulnerable edges due to crossing roads are shown in dark blue. The largest teaming cost reductions on the vulnerable edges come from at least four robots traversing together. In the solution routes shown, robot teams split up and form new teams as they move through the terrain providing overwatch. 
	}
	\label{fig:built_graph}
	\vspace*{-3mm}
\end{figure}

\subsubsection{Meadow Map 1} 
\begin{figure}[t]
	% \vspace*{2mm}
	\centering
	\includegraphics[trim={0cm 1.2cm 0cm 1.2cm},clip, width=0.9\columnwidth]{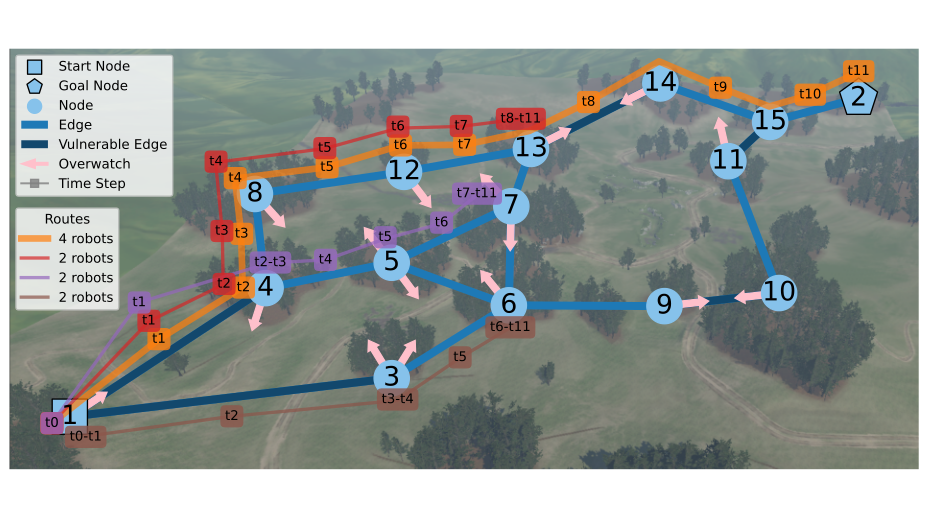}
    \vspace*{-2mm}
	\caption{Reconnaissance scenario spanning a meadow environment. 
    The nodes are placed in forested regions of cover. Ten robots start at node 1 with the goal of at least one robot reaching node 2, across the meadow. 
    The robot team routes through the graph, solved for with our proposed method, show extensive utilization of overwatch opportunities. 
	}
	\label{fig:built_side_problem}
	\vspace*{-4mm}
\end{figure}

In Figures~\ref{fig:built_graph} and \ref{fig:built_side_problem}, 
we demonstrate traversing between areas of cover in a high-risk environment, using a simulated meadow environment from \cite{NatureManufacture}. 

In Fig.~\ref{fig:built_graph}, we consider 10 robots starting at node 1 with a goal of all robots reaching node 2 within 10 time steps. In our generated solution, three distinct paths emerge. Through time step 3, six robots navigate together (orange and red teams) alternating providing overwatch with another team (purple) of four robots. At time step 4, two robots (red) remain at node 4 to provide overwatch for the purple team, while orange navigates toward the next overwatch position. This allows the purple and red teams to merge and proceed alternating providing overwatch with the orange team through the rest of the graph toward the goal node. 

\subsubsection{Meadow Map 2}

Fig.~\ref{fig:built_side_problem} is an example with further subdivision  
of the robots in the solution. Again, we consider 10 robots starting at node 1. 
In this scenario, the goal is for at least one robot to reach node 2 within 12 time steps.
In this example, four coalitions emerge: a primary coalition of four robots (orange) advancing toward the goal while providing overwatch as needed, and three smaller coalitions (red, purple, brown) supporting the primary coalition and each other.
Since all robots do not need to reach the goal node, this example highlights the trade-off between the cost of traversing and the benefit of overwatch. The supporting coalitions eventually halt their movement, as the cost of traversing does not outweigh the benefits of overwatch they could provide. 
This concept is consistent with expectations in an operational setting: the risk outweighs the reward. Ultimately, our approach's objective is to determine tactical maneuvers 
in complex environments. 

\subsubsection{Ablation Study}

\begin{figure}
	\centering
	\begin{subfigure}[t]{0.5\columnwidth}
		\centering
		\includegraphics[trim={0cm 0.5cm 0cm 0cm},clip,width=\textwidth]{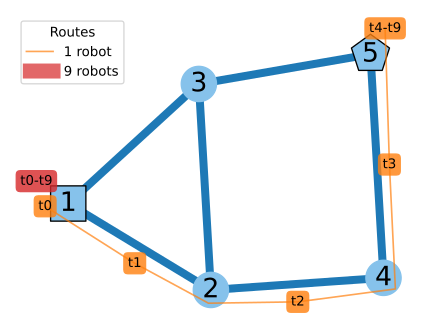}
		\captionsetup{font=scriptsize}
        \caption{Without overwatch, vulnerability, \\and teaming}
		\label{fig:toy1_without_overwatch_vul_teaming}
	\end{subfigure}%
	\begin{subfigure}[t]{0.5\columnwidth}
		\centering
		\includegraphics[trim={0cm 0.5cm 0cm 0cm},clip,width=\textwidth]{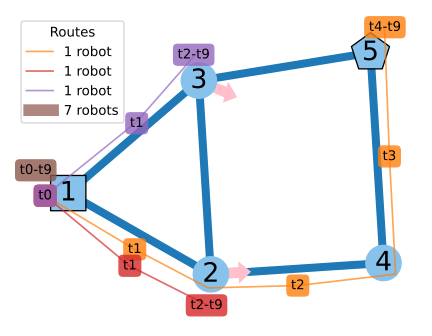}
		\captionsetup{font=scriptsize}
        \caption{With overwatch and without vulnerability and teaming}
		\label{fig:toy1_without_vul_teaming}
	\end{subfigure}%1.4cm 1.4cm 1cm 0.5
	
	\begin{subfigure}[t]{0.5\columnwidth}
		\centering
		\includegraphics[trim={0cm 0.5cm 0cm 0cm},clip,width=\textwidth]{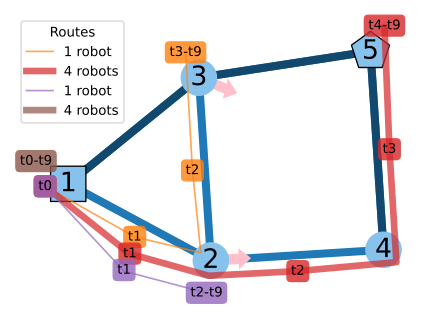}
		\captionsetup{font=scriptsize}
        \caption{With overwatch and vulnerability \\and without teaming}
		\label{fig:toy1_without_teaming}
	\end{subfigure}%
	\begin{subfigure}[t]{0.5\columnwidth}
		\centering
		\includegraphics[trim={0cm 0.5cm 0cm 0cm},clip,width=\textwidth]{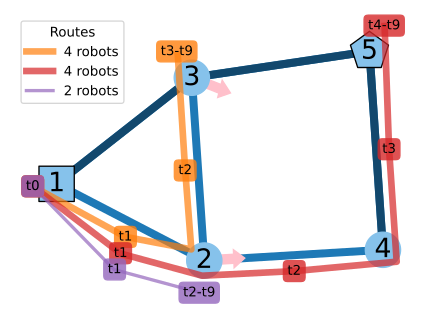}
		\captionsetup{font=scriptsize}
        \caption{Full solution (with overwatch, vulnerability, and teaming)}
		\label{fig:toy_problem_solution_paths}
	\end{subfigure}%
	\vspace*{1.5mm}
	\begin{subfigure}[t]{\columnwidth}
		\centering
		\includegraphics[width=\textwidth]{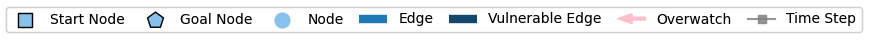}
	\end{subfigure}%
    \vspace*{-2mm}
	\caption{Ablation study illustrating the impact of key components in our algorithm: overwatch opportunities, edges with high vulnerability, and incentives for moving as a team. Robot team paths are shown in each subplot with components of our formulation incrementally added.}
	\label{fig:toy1_ablation}
    \vspace*{-4mm}
\end{figure}

\begin{figure*}[t]
    \centering
    % \vspace*{2mm}
    \begin{subfigure}[t]{0.21\textwidth}
		\centering
		\includegraphics[width=\textwidth]{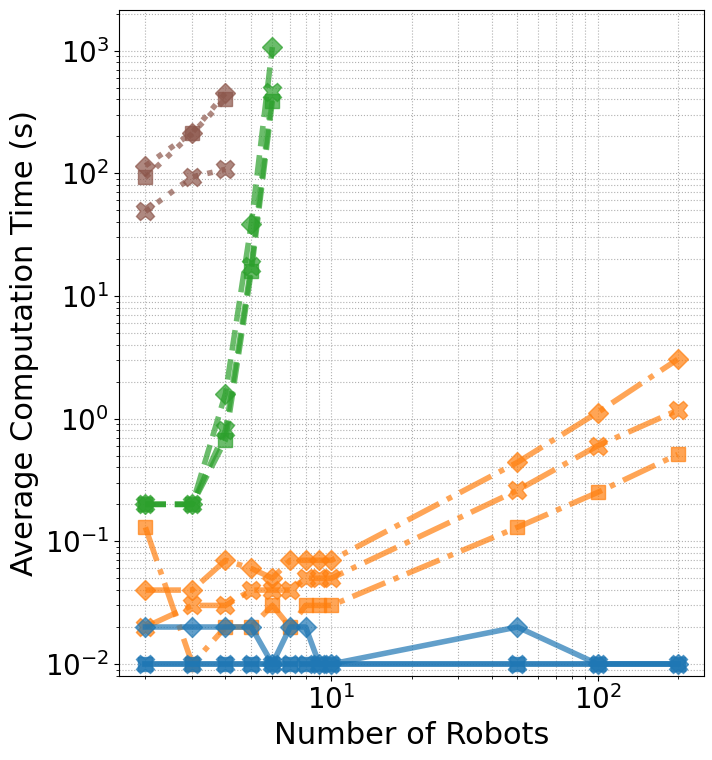}
		\captionsetup{font=scriptsize}
        \caption{5 Node Graphs}
        \label{fig:comparison_plots:5nodes}
	\end{subfigure}
	\begin{subfigure}[t]{0.21\textwidth}
		\centering
		\includegraphics[width=\textwidth]{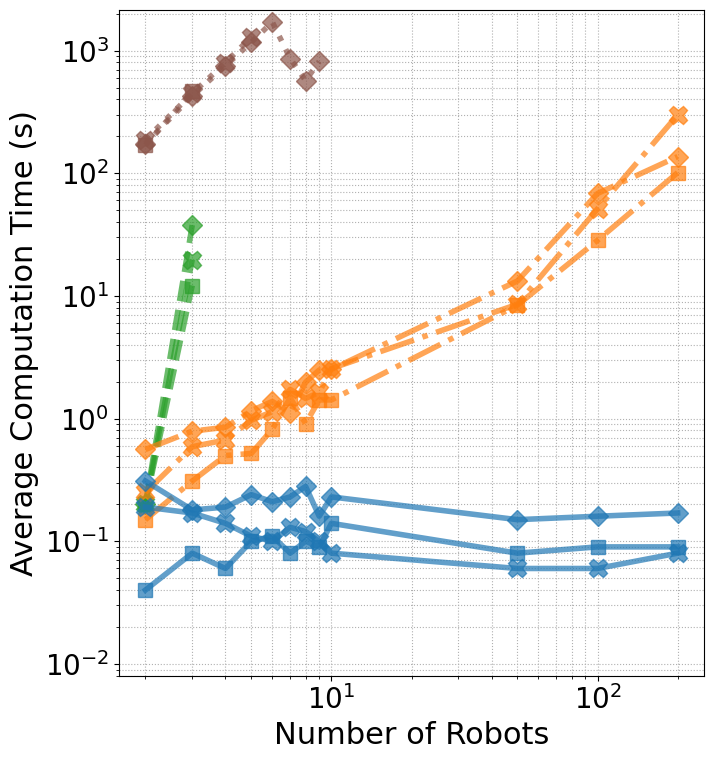}
		\captionsetup{font=scriptsize}
        \caption{15 Node Graphs}
        \label{fig:comparison_plots:15nodes}
	\end{subfigure}
	\begin{subfigure}[t]{0.21\textwidth}
		\centering
		\includegraphics[width=\textwidth]{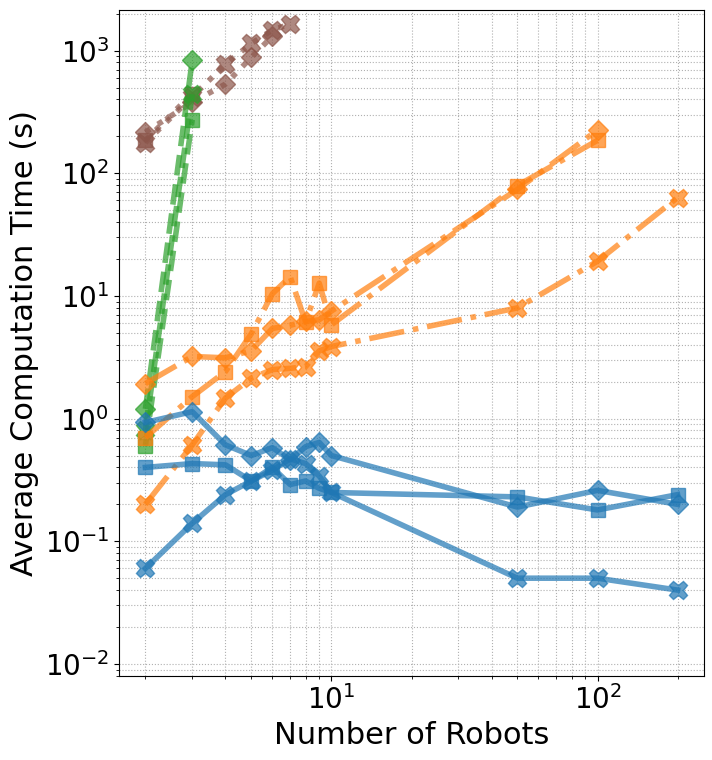}
		\captionsetup{font=scriptsize}
        \caption{25 Node Graphs}
        \label{fig:comparison_plots:25nodes}
	\end{subfigure}
    \begin{subfigure}[t]{0.21\textwidth}
		\centering
		\includegraphics[width=\textwidth]{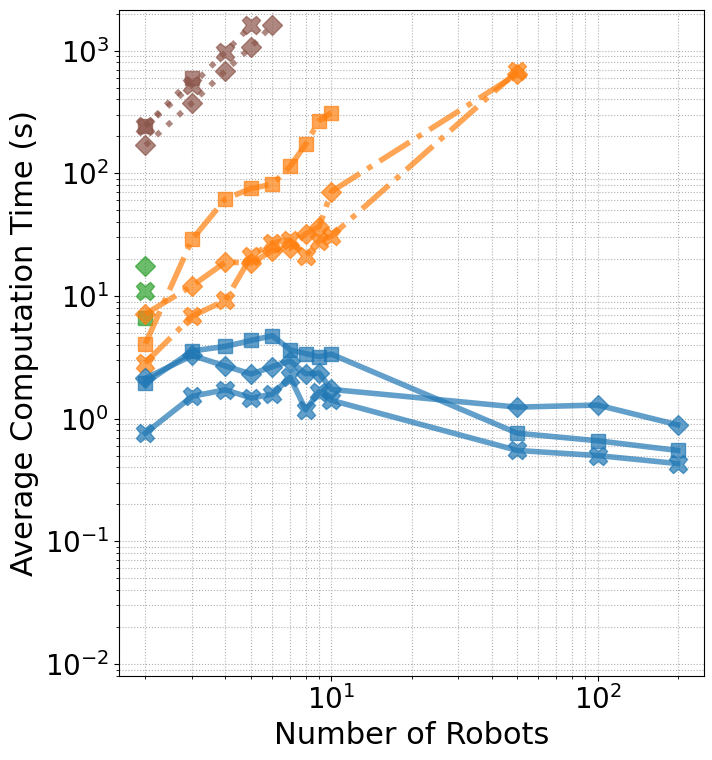}
		\captionsetup{font=scriptsize}
        \caption{50 Node Graphs}
        \label{fig:comparison_plots:50nodes}
	\end{subfigure}
    \begin{subfigure}[t]{0.8\textwidth}
		\centering
		\includegraphics[width=\textwidth]{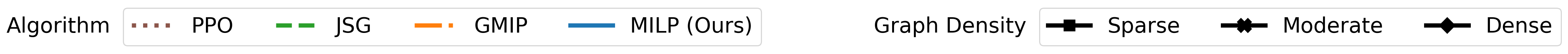}
	\end{subfigure}%
    \vspace*{-4mm}
	\caption{Comparison of average computation time across state-of-the-art algorithms for multi-robot planning with overwatch opportunities across various graph sizes, edge densities, and numbers of robots. Each algorithm is adapted to solve the same problem and our approach is labeled as ``MILP.''}
	\label{fig:comparison_plots}
    \vspace*{-4mm}
\end{figure*} % comparison_plots

\begin{figure}[thb]
    \centering
    \includegraphics[width=0.75\columnwidth]{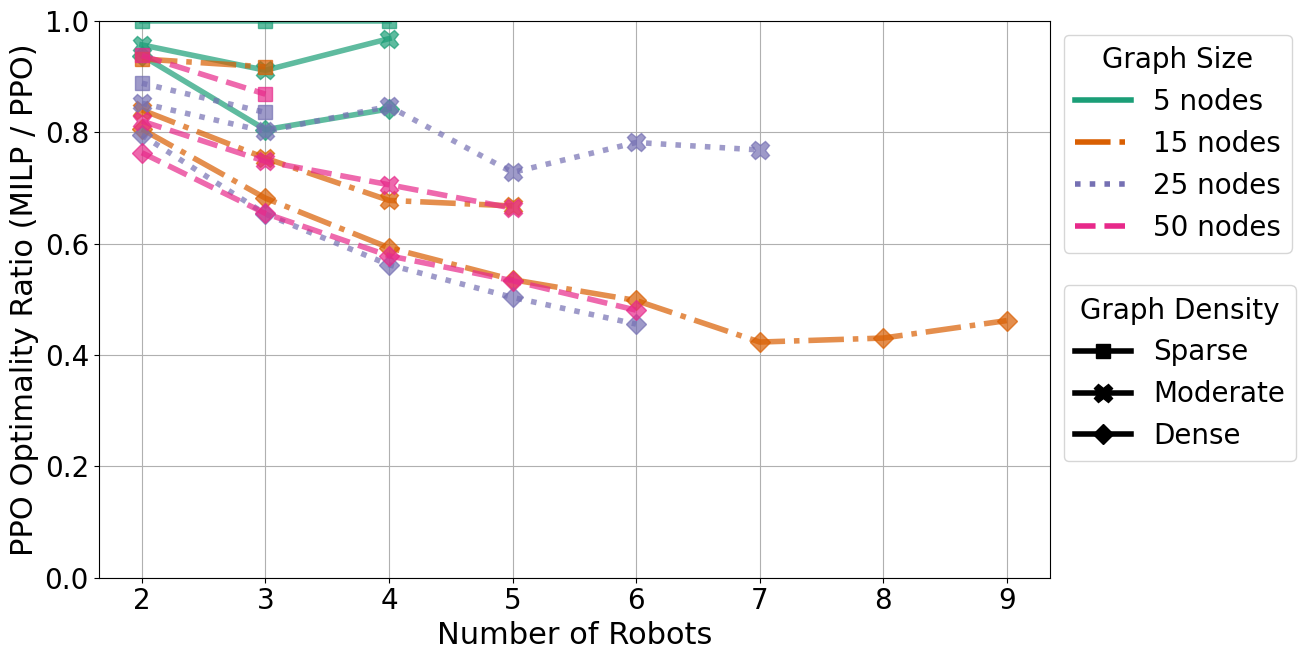}
    \vspace*{-2mm}
    \captionsetup{font=scriptsize}
    \caption{Optimality gap versus the number of robots for the learning-based \ac{PPO} algorithm. All other approaches were solved to optimality.}
    \label{fig:ppo_optimality_gap}
    \vspace*{-4mm}
\end{figure} % ppo_optimality_gap

We performed an ablation study to assess the impact of providing overwatch, formations on vulnerable edges, and moving as a team wherever possible.  
Fig.~\ref{fig:toy1_ablation} shows the solutions to our illustrative example, from Fig.~\ref{fig:toy_problem_solution}, when ablating these components. 

When we remove the overwatch, vulnerability, and teaming components, in Fig.~\ref{fig:toy1_without_overwatch_vul_teaming}, the problem becomes a shortest path problem. The weights in the graph are fixed and one robot takes the path with the least overall edge and time cost. 

In tactical maneuvers, overwatch allows minimizing detection risk. When we add our formulation of overwatch opportunities in Fig.~\ref{fig:toy1_without_vul_teaming}, the edge weights now vary based on the overwatch opportunities being utilized by the team and the optimal solution includes robots moving to the two overwatch positions while one robot traverses to the goal. 

As an incentive to travel in a formation on edges that are particularly dangerous, or vulnerable, we add our vulnerability component back in Fig.~\ref{fig:toy1_without_teaming}, making edges (1,3), (3,5), and (4,5) more costly to traverse alone. We see in the solution that the desired minimum of 4 robots traverse edge (4,5).

In a tactical scenario, more robots moving together would be advantageous (e.g., due to redundancy or additional sensing). 
When we add our teaming component back into the problem in Fig.~\ref{fig:toy_problem_solution_paths}, we see all robots moving for the first time, no longer leaving robots at the start node, since moving as a team provides cost reductions on each edge and greater cost reductions when providing overwatch. 
Ultimately, the overwatch, vulnerability, and teaming components yield practical plans for tactical maneuvers that minimize detection and risk.

\subsection{Graph Planning Computation Time}

\subsubsection{Comparison to \acf{SOTA}}
We compare our \ac{MILP} approach to the current \ac{SOTA} baselines for multi-robot planning with support: (i) JSG, which enumerates a joint state graph for all combinations of robots at locations and then performs a shortest path search \cite{limbu2024team}; (ii) \ac{PPO}, a reinforcement learning approach that trains on a single graph and targets the challenge of scaling to larger graphs and teams \cite{limbu2024scaling}; and (iii) GMIP, a generic \ac{MIP} formulation, based on standard paradigms \cite{Yu2016, surynek2022problem}, that enumerates binary variables for each robot's location at each time step (or $n_T n_L n_A$ variables). 
We adapted each algorithm to solve the same underlying problem. For this purpose, we compare to a simplified version of our algorithm (excluding the complexity of vulnerability and teaming), which we label ``\ac{MILP}," to show the benefits of our formulation. In particular, we aim to show the major impact of removing the dependence on the number of agents in our decision space (scaling by $n_T n_L$ variables).

We evaluate each algorithm for graph sizes from 5 to 50 nodes with edge density levels: ``Sparse,'' ``Moderate,'' and ``Dense,'' which correspond to 20\%, 50\%, and 80\% of a fully connected graph's edges. We set 40\% of those edges to have up to 2 overwatch opportunities. 
We then solve for multi-robot plans with team sizes in the set $\{2-10, 50, 100, 200\}$. Start and goal positions are selected based on the furthest apart nodes by edge cost and the MIP-based approaches use a time horizon twice the shortest-path length. 

\ac{PPO} is trained per graph, so we compare training and inference time to JSG's construction and shortest path search time, as in \cite{limbu2024scaling}. For GMIP and our MILP approach, we present the computation time to an optimal solution using the Gurobi optimizer \cite{GurobiOptimization2023}. All algorithms were run on a 13th Gen Intel® Core™ i9-13950HX × 32 CPU and \ac{PPO} additionally used a NVIDIA GeForce RTX 4090 Laptop GPU. We averaged computation times across three seed values for each graph and consider failure when, for any of the seed values, the solve time exceeds 30 minutes, requires more than 16GB of RAM, or (in the case of \ac{PPO}) does not successfully reach the goal. We terminate training for \ac{PPO} when the rolling 50-episode average reward changes by less than 0.2 for 500 episodes \cite{limbu2024scaling}.

Fig.~\ref{fig:comparison_plots} compares the average computation time of each algorithm for four graph sizes versus the number of robots. The computation time and number of robots are shown on a log-scale with the same Y-axis used in each plot to see relative magnitude. In these plots, less computation time is ideal.
Rapid computation time (on the order of magnitude of minutes for edge lengths of about $100-250m$) is essential to enable re-planning with new information.
Our MILP approach finds optimal solutions in under 5 seconds across all graph sizes, densities, and team sizes. This is several orders of magnitude faster than the baselines, which quickly become intractable for larger numbers of robots and graph sizes. Notably, in some cases, the solve time of our \ac{MILP} decreases for larger number of agents since the problem becomes easier when more overwatch opportunities can be utilized.

JSG, GMIP, and MILP all return optimal solutions. In Fig.~\ref{fig:ppo_optimality_gap}, we show the optimality gap for the \ac{PPO} approach relative to our MILP formulation. 
In this plot, a larger ratio is better (i.e., 1.0 represents optimality). 
\ac{PPO}'s optimality ratio decreases with team size and graph density, highlighting \ac{PPO}'s decreasing solution quality in these regimes. 

\begin{table}[t]
    % \vspace*{-3mm}
	\centering
	\caption{Example Graphs' Problem Size and Computation Time} 
	\label{tab:computation_time}
    \vspace*{-2mm}
	\begin{center}
		\renewcommand{\arraystretch}{1.3}
		\begin{tabular}{ c || c | c | c | c }
			% & \multicolumn{4}{c}{\textbf{Topological Graph}} \\
			& \textbf{Illustr.} & \textbf{Bounding} & \textbf{Map 1} & \textbf{Map 2} \\
			\hline \hline
			\textbf{Locations, $n_L$} & 17 & 43 & 32 & 51 \\
			\textbf{Overwatch, $n_\scriptO$} & 4 & 8 & 18 & 32 \\
			\textbf{Time Horizon, $n_T$} & 10 & 10 & 10 & 12 \\
			\hline
			\textbf{Total Variables} & 460 & 1160 & 990 & 1872 \\
			\hline
			\textbf{Mean Solve Time (s)} & 0.081 &	0.100 &	0.249 &	3.702 \\	
		\end{tabular}
	\end{center}
    \vspace*{-5mm}
\end{table}

\subsubsection{Example Graphs' Computation Time}

We additionally report the computation time for the simulated graphs discussed in this paper, where we consider a more complex problem with teaming and vulnerability considerations. 
Table~\ref{tab:computation_time} shows the parameters that affect the problem size (i.e., $n_L, n_\scriptO, n_T$), the resulting total optimization variables, and the solve time for our graph planner averaged across 100 trials. In all cases, the number of robots was 10. Varying the number of robots does not impact the total number of variables in our problem.
The graphs are labeled Illustrative, Bounding, Map 1, and Map 2, which correspond to Fig.~\ref{fig:toy_problem_solution}, Fig.~\ref{fig:problem4}, Fig.~\ref{fig:built_graph}, and Fig.~\ref{fig:built_side_problem}, respectively. 
We select time horizons for each problem based on the size of the graph and location of the goal to ensure that the problem is feasible. 
These results show computation times in seconds for complex operational scenarios.

\begin{figure}[tbh]
    \centering
    % \hspace*{6mm}
    % \begin{center}
    % \begin{minipage}{0.49\linewidth}
    % \hspace*{2.5mm}
	\begin{subfigure}[t]{0.3\columnwidth}
		\centering
		\includegraphics[height=0.11\textheight]{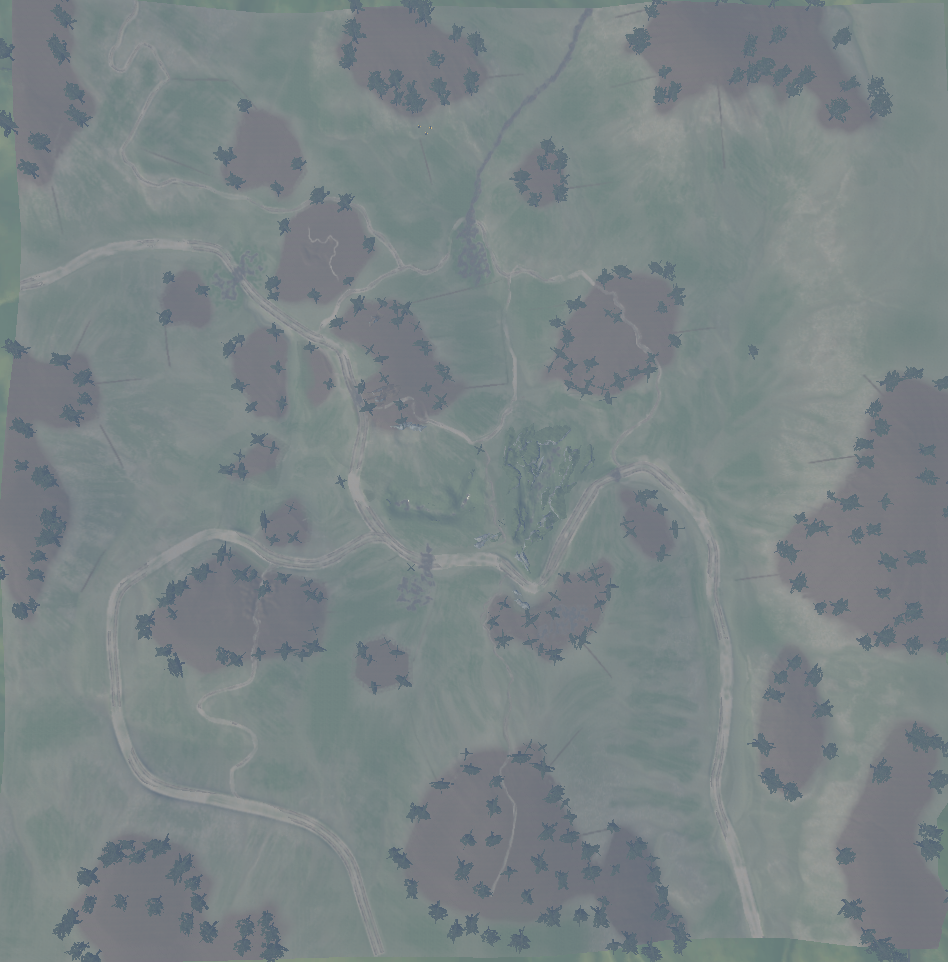}
		\captionsetup{font=scriptsize}
        \caption{Meadow environment}
		\label{fig:meadow}
	\end{subfigure}
    % \hspace*{0.01mm}
	\begin{subfigure}[t]{0.3\columnwidth}
		\centering
		\includegraphics[height=0.11\textheight]{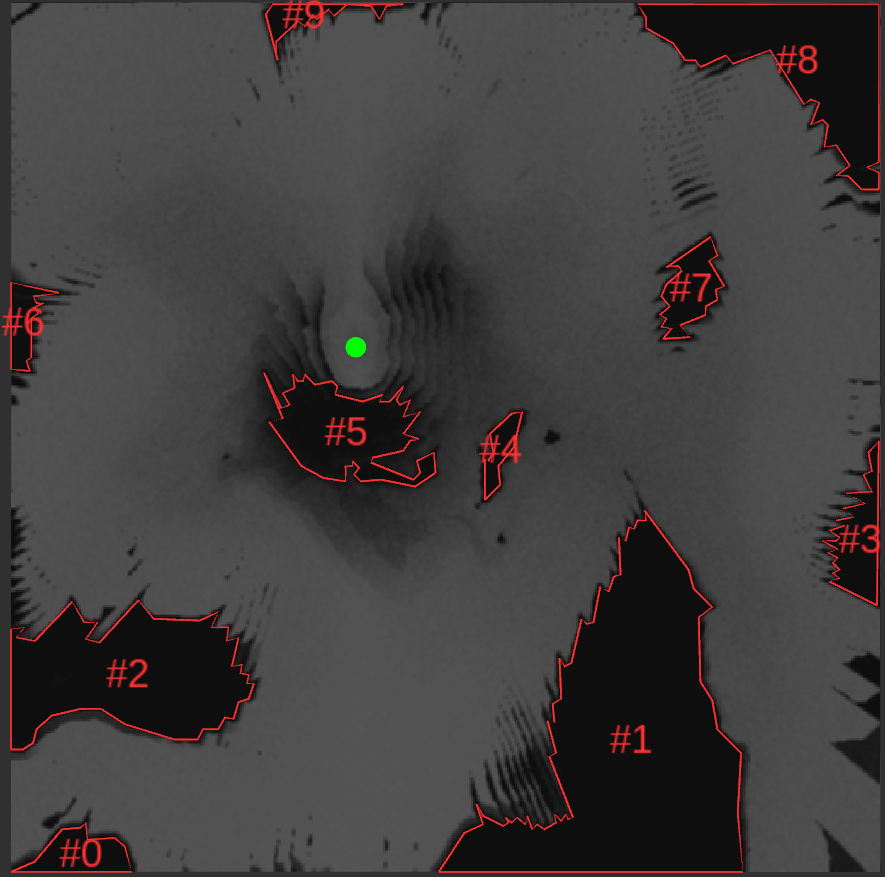}
		\captionsetup{font=scriptsize}
        \caption{Visibility map %: cover regions outlined in red and observer hilltop location indicated by green covariance ellipse.
        }
		\label{fig:vis_map_seq}
	\end{subfigure}
    % \hspace*{0.1mm}
    \begin{subfigure}[t]{0.3\columnwidth}
		\centering
		\includegraphics[height=0.11\textheight]{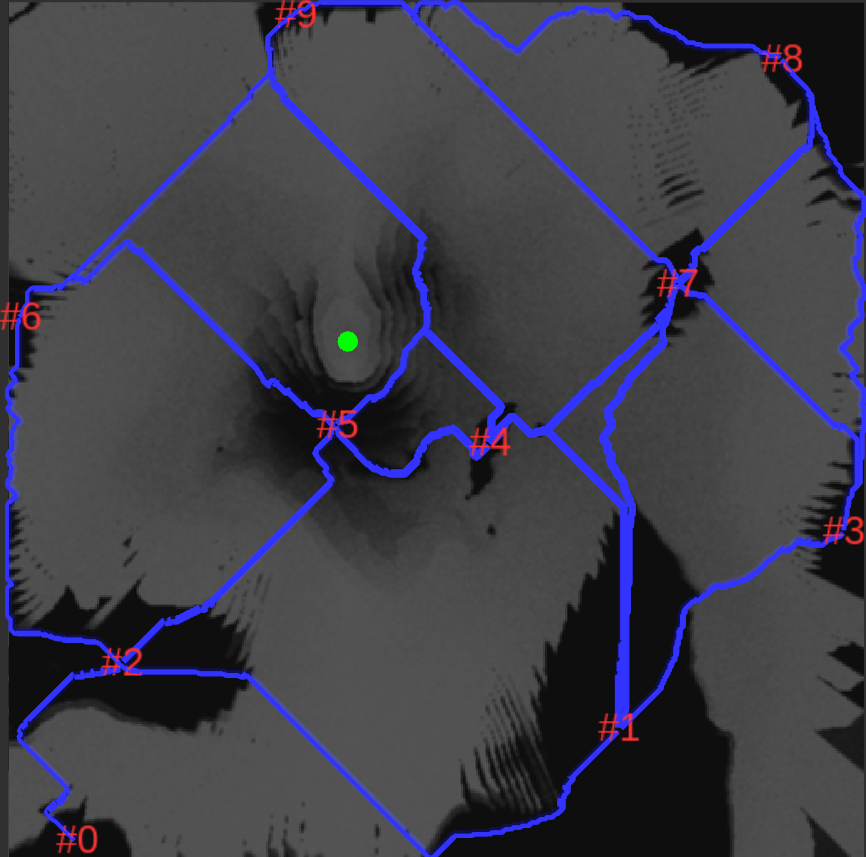}
		\captionsetup{font=scriptsize}
        \caption{Pruned A$^*$ paths %in blue between regions of cover.
        }
		\label{fig:astar_paths_seq}
	\end{subfigure}
    \vspace*{1mm}
    % \hspace*{0.5mm}
    
    \begin{subfigure}[t]{0.3\columnwidth}
		\centering
		\includegraphics[height=0.11\textheight]{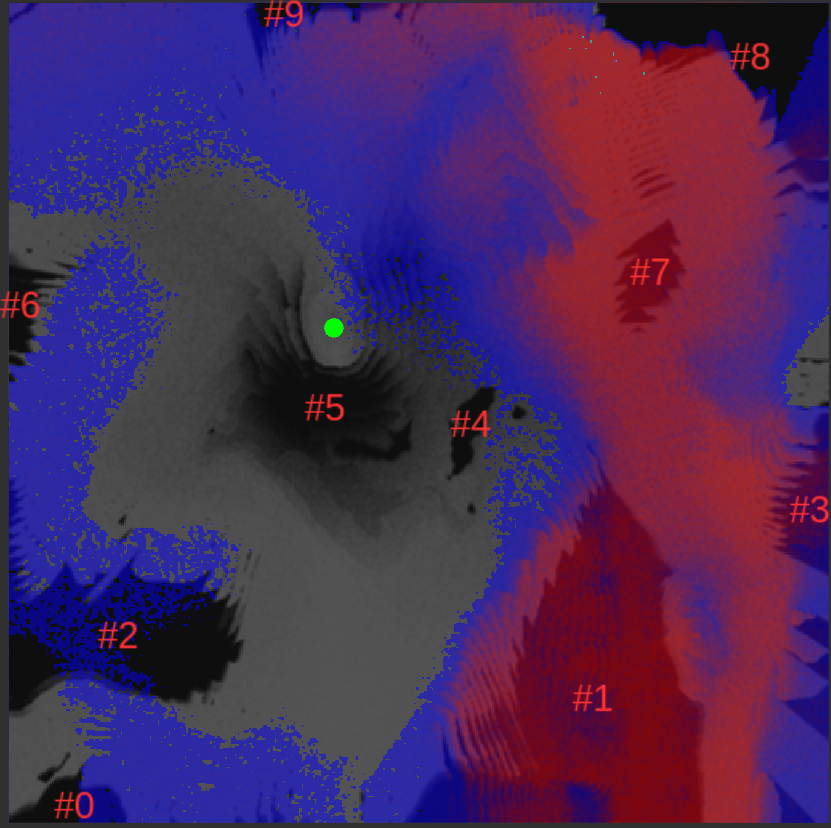}
		\captionsetup{font=scriptsize}
        \caption{Node 1 overwatch}
		\label{fig:overwatch_seq}
	\end{subfigure}
    % \hspace*{0.1mm}
	\begin{subfigure}[t]{0.3\columnwidth}
		\centering
		\includegraphics[height=0.11\textheight]{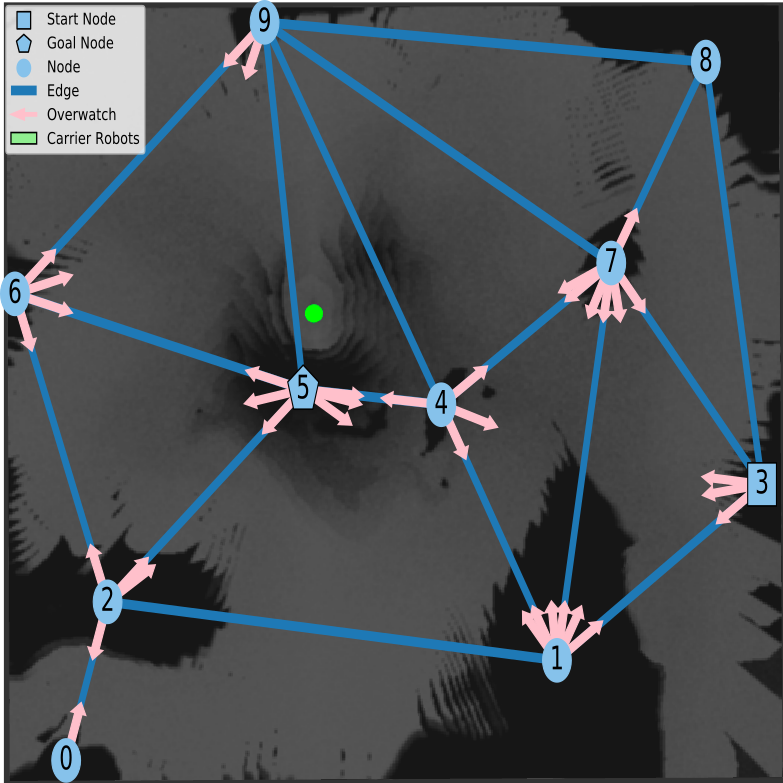}
		\captionsetup{font=scriptsize}
        \caption{DTG
        %Resulting topological graph with overwatch opportunities indicated from the overwatch node to edges that can be observed.
        }
		\label{fig:graph_seq}
	\end{subfigure}
    % \hspace*{0.1mm}
	\begin{subfigure}[t]{0.3\columnwidth}
		\centering
		\includegraphics[height=0.11\textheight]{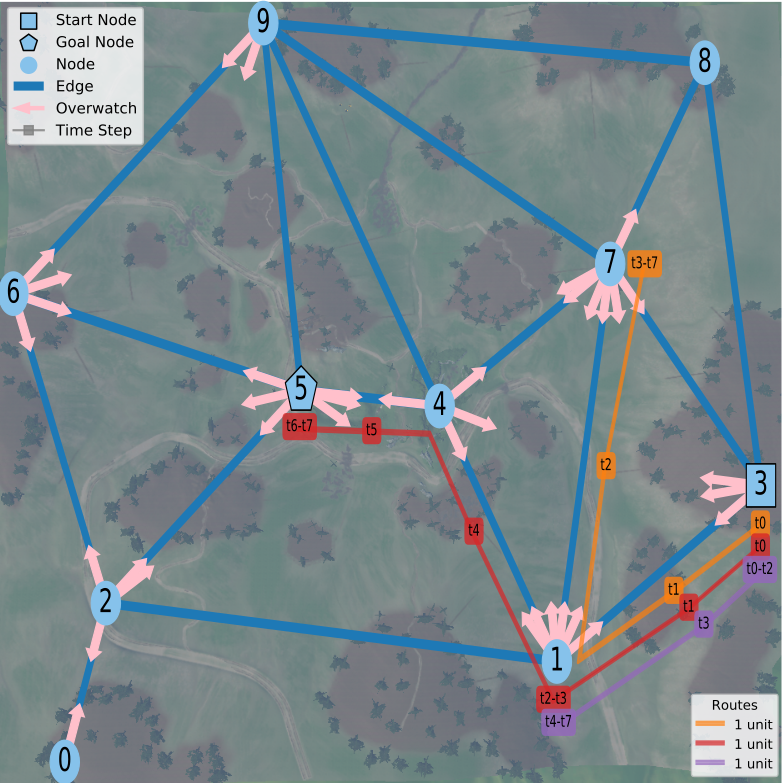}
		\captionsetup{font=scriptsize}
        \caption{Robot team solution}
		\label{fig:plans}
	\end{subfigure}
    % \end{minipage}
    % \end{center}
    % \hspace*{1mm}
    % \vspace*{-1mm}
    \caption{Environment segmentation steps in simulated meadows environment \cite{NatureManufacture}, as seen in (a). In (b), we generate a visibility map to segment the regions of cover (outlined in red). The observer's hilltop location is indicated by the green covariance ellipse. We then generate A$^*$ paths between these regions of cover and prune the extraneous edges, as seen in blue in (c). For each node we generate visibility maps for overwatch, (d) shows an example for node 1. Using this information, we form our topological map in (e). Overwatch opportunities are indicated from the overwatch nodes to edges that can be observed with pink arrows.
    We then use \ac{MIP} to solve for paths for each robot in the team through this graph, shown in (f). A robot team of 3 starts at node 3 with the objective of at least one robot reaching node 5 while minimizing visibility through overwatch and teaming.}
    \label{fig:sim_experiment}
    \vspace*{-2mm}
\end{figure}

\begin{figure}[tbh]
    % \vspace*{-3mm}
    \centering
	\begin{subfigure}[t]{0.41\columnwidth}
		\centering
		\includegraphics[width=\textwidth]{figures/cavazos_loc1/legend/paths_only_tif.png}
		\captionsetup{font=scriptsize}
        \caption{Forested Graph 1}
	\end{subfigure}
	\begin{subfigure}[t]{0.41\columnwidth}
		\centering
		\includegraphics[width=\textwidth]{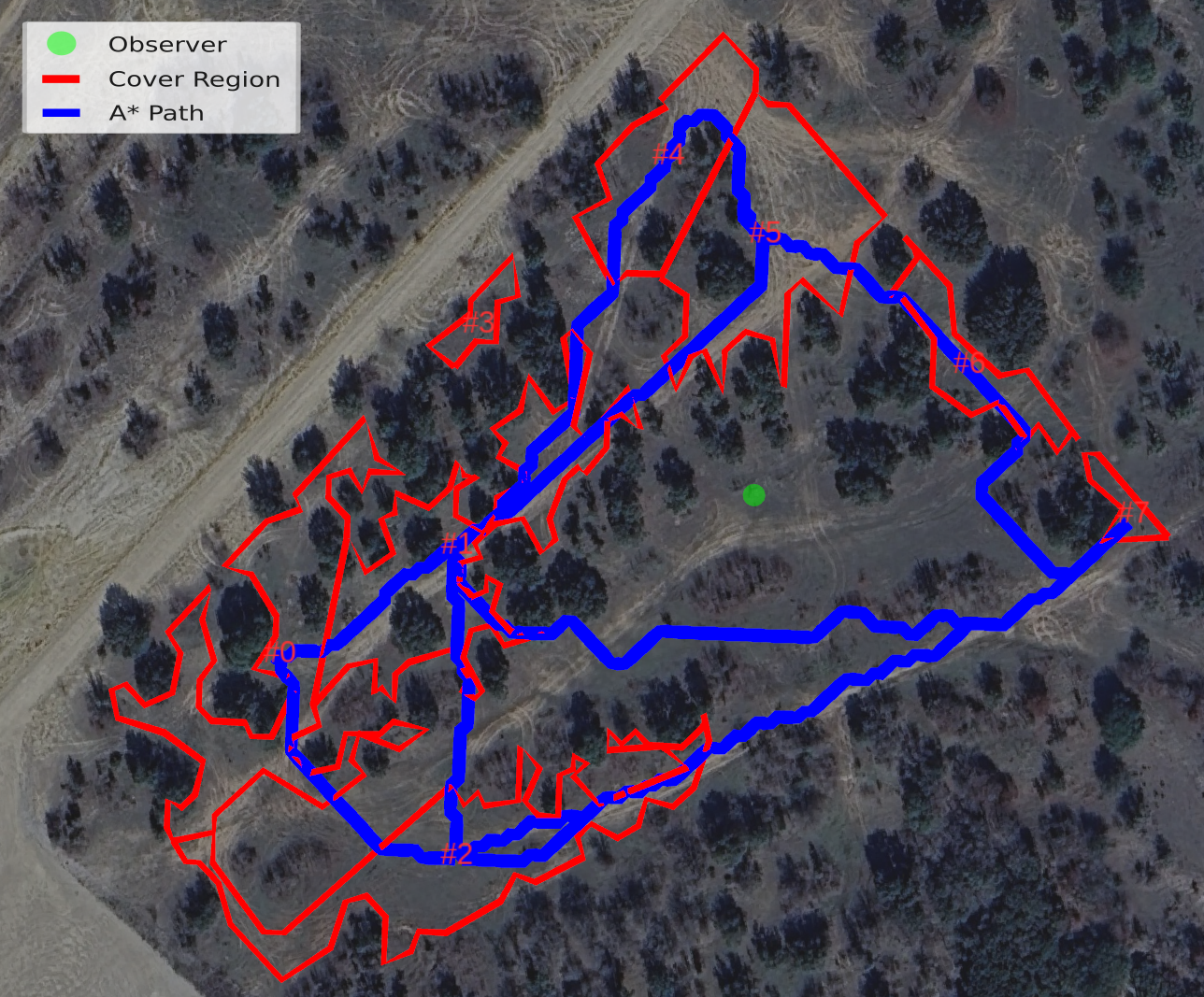}
		\captionsetup{font=scriptsize}
        \caption{Forested Graph 2}
	\end{subfigure}
	\caption{Graph generation for two different observer locations in an forested off-road environment. Resulting cover regions and A$^*$ paths between the regions are depicted for each.}
	\label{fig:cavazos_paths}
    \vspace*{-3mm}
\end{figure} % cavazos_paths

\subsection{\acs{STALC} Simulated Graph Generation and Planning}

Fig.~\ref{fig:sim_experiment} demonstrates the steps of our environment segmentation approach in a simulated meadow scenario for an observer on a hill. 
For a large simulated environment ($\approx$ 500m $\times$ 500m), we are able to generate a topological graph from a visibility map and we demonstrate coordinated maneuvers with a team of 3 ground vehicles. The paths of the vehicles are depicted in Fig.~\ref{fig:plans}.

\begin{figure*}[tbh]
    \centering
    % \vspace*{2mm}
    \begin{subfigure}[t]{0.58\columnwidth}
		\centering
		\includegraphics[width=\textwidth]{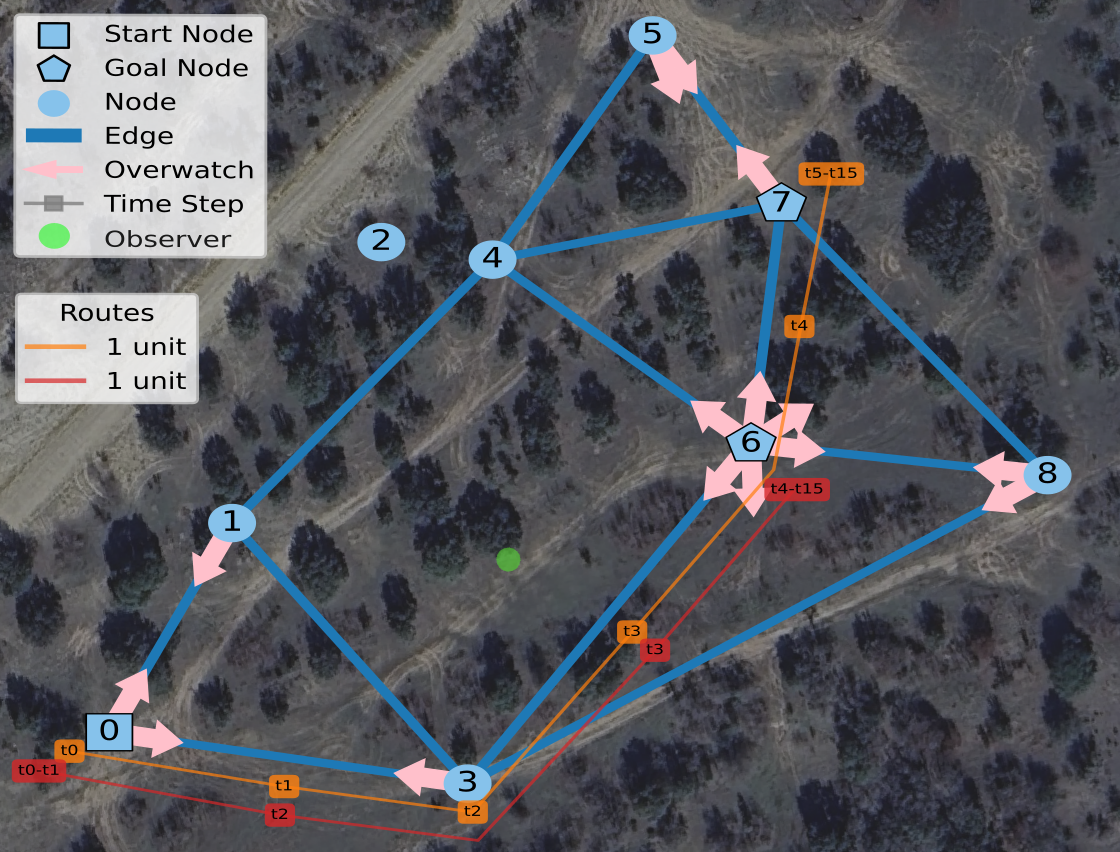}
		\captionsetup{font=scriptsize}
        \caption{Forested Graph 1: routes prioritize formation}
        \label{fig:cavazos_graphs:formation_loc1}
	\end{subfigure}
	\begin{subfigure}[t]{0.58\columnwidth}
		\centering
		\includegraphics[width=\textwidth]{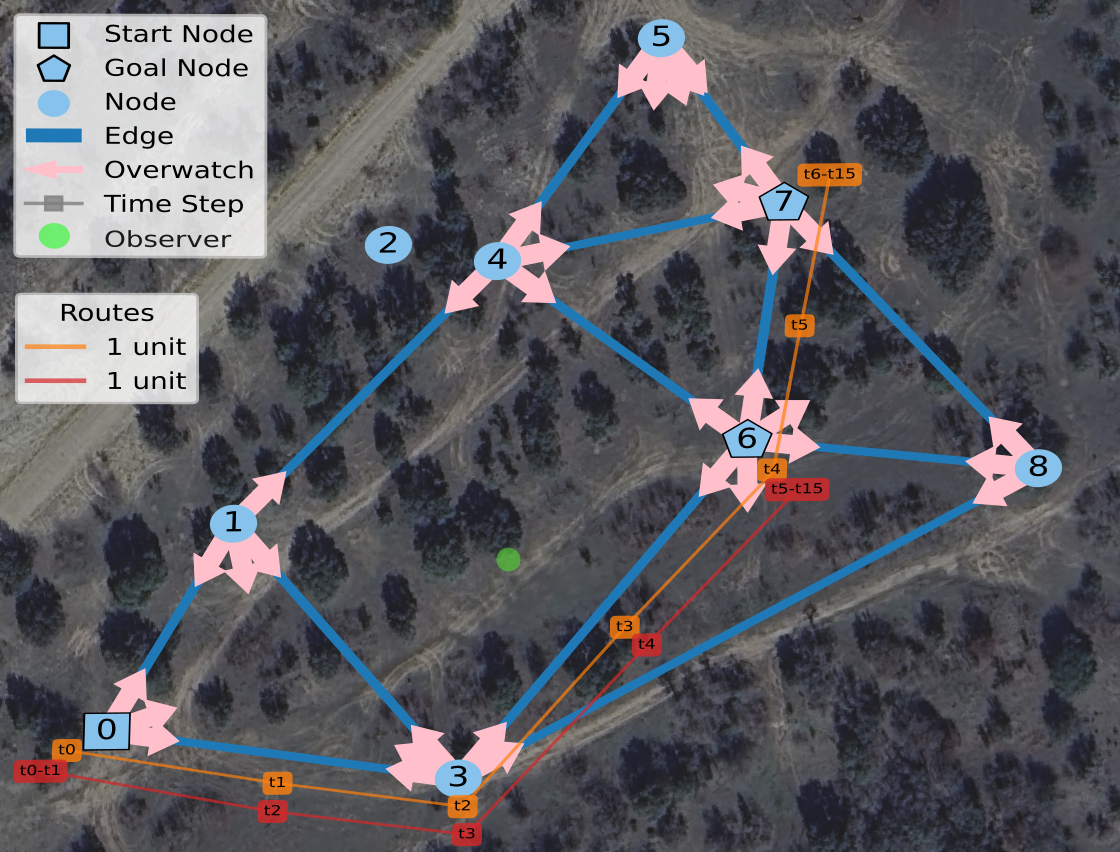}
		\captionsetup{font=scriptsize}
        \caption{Forested Graph 1: routes prioritize overwatch}
        \label{fig:cavazos_graphs:loc1}
	\end{subfigure}
	\begin{subfigure}[t]{0.58\columnwidth}
		\centering
		\includegraphics[width=\textwidth]{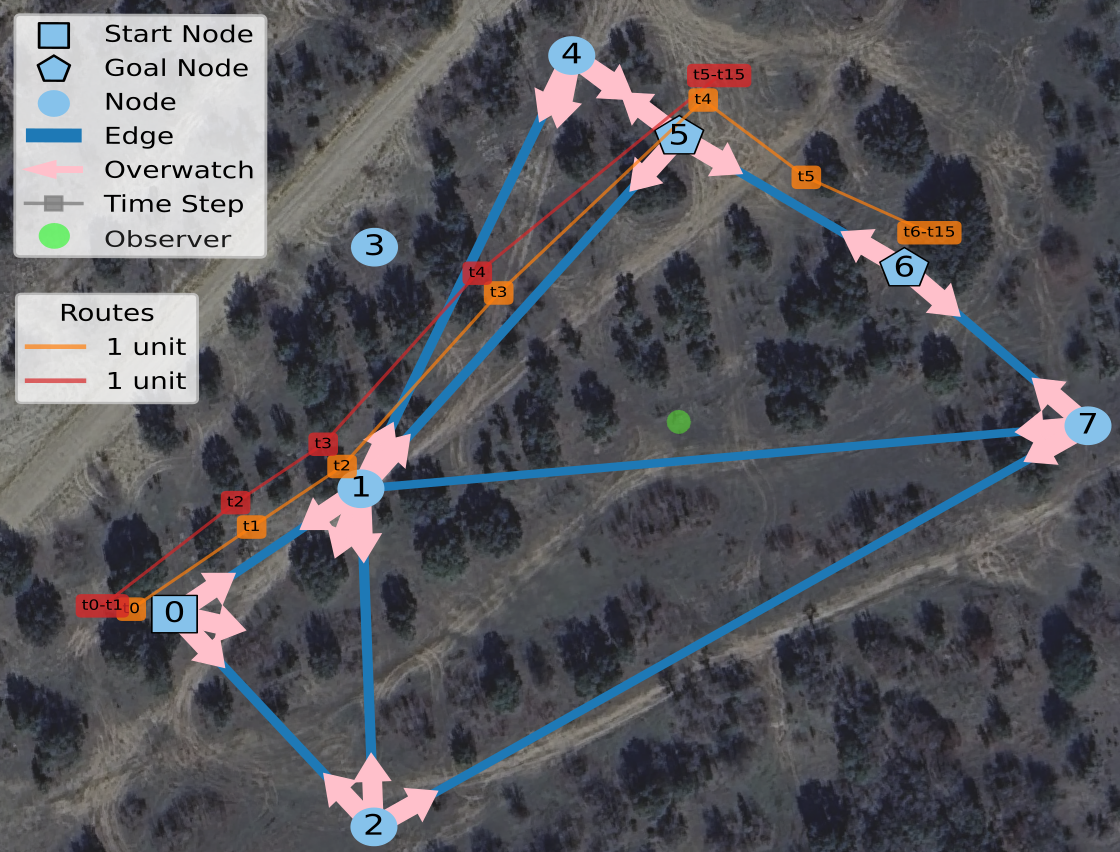}
		\captionsetup{font=scriptsize}
        \caption{Forested Graph 2}
        \label{fig:cavazos_graphs:loc2}
	\end{subfigure}
	\caption{Schematic forested graphs showing overwatch opportunities and team routes. Both robots start at node 0 and the goal is for one robot to node 6 and one robot to reach node 7. The priority of overwatch is varied in Forested Graph 1 to show the difference in the resulting paths.}
	\label{fig:cavazos_graphs}
    \vspace*{-5mm}
\end{figure*} % cavazos_graphs

\subsection{\acs{STALC} Hardware Experiments}

We evaluated our hierarchical planner \ac{STALC} in hardware experiments to validate our graph generation and visibility calculations in real-world environments. Additionally, we evaluated the feasibility of solving (\ref{eq:basic_opt_problem}) using a high-level and low-level planning paradigm. 
We used Clearpath Warthogs in these experiments as a surrogate platform, though our hierarchical planner is broadly applicable to other systems.  

Our hardware experiments demonstrate 
minimum-visibility planning with a team of autonomous ground vehicles with our hierarchical approach: coupling our high-level graph-based planner with our mid-level A$^*$ planner and low-level stochastic \ac{NMPC}. 
We evaluated our approach in two highly constrained environments: forested off-road terrain and an urban environment. In both cases, the vehicles needed to avoid both static and dynamic obstacles (coalition members), operate in formation, and provide overwatch to their team members. We use \emph{a priori} data of the environment to construct our visibility and traversability maps. We considered two different observer locations in each environment to demonstrate how the environment segmentation and ultimate route through the environment will vary based on the scenario. Additionally, we show results with different emphases on overwatch versus moving in formation to demonstrate the adaptability of our graph planner to different scenario objectives.

\begin{figure}[t]
    \centering
    \includegraphics[width=0.75\columnwidth]{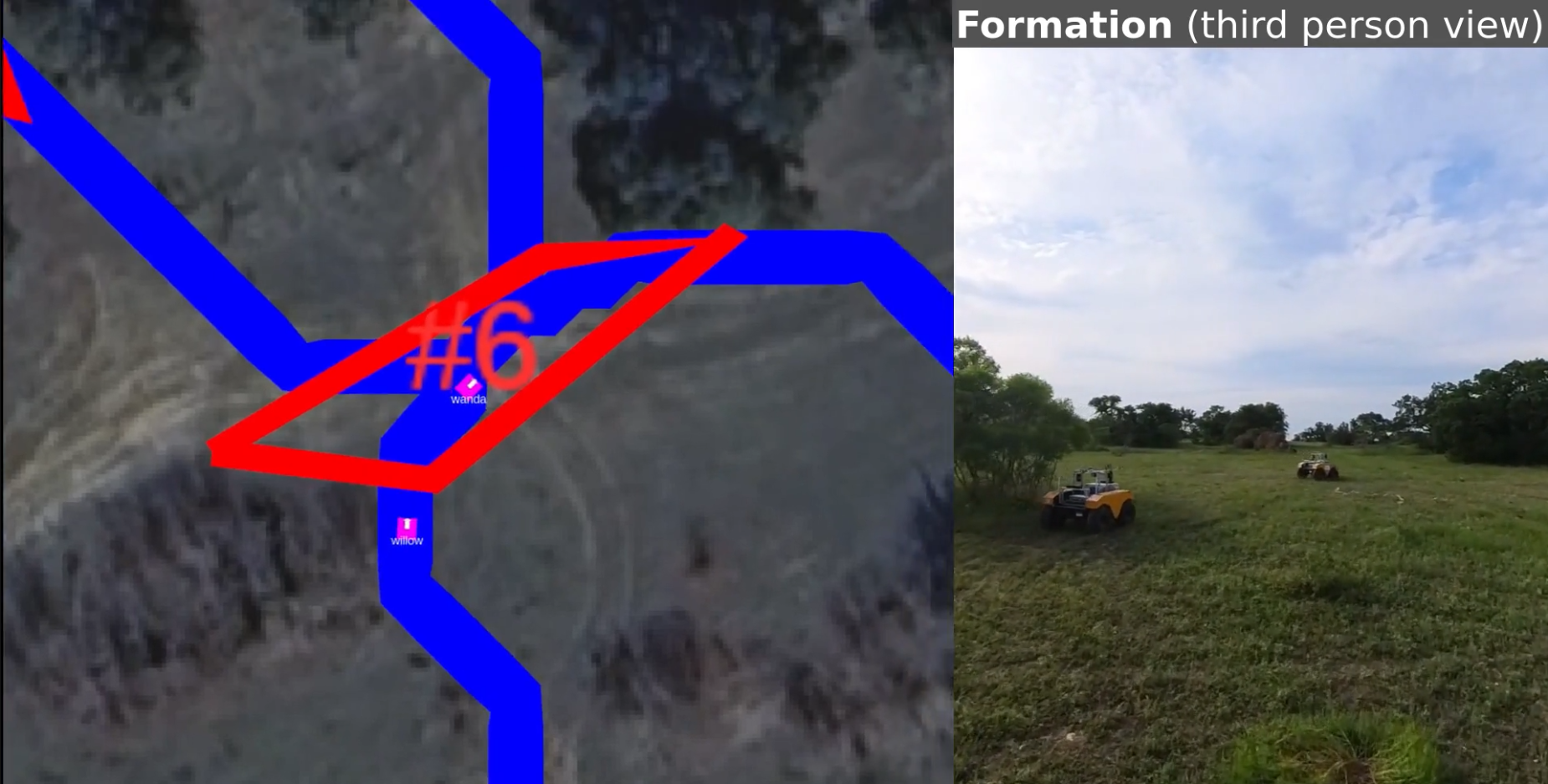}
    \vspace*{-2mm}
    \captionsetup{font=scriptsize}
    \caption{Robots moving in formation approaching a cover region.}
    \label{fig:formation_view}
    \vspace*{-2mm}
\end{figure} % formation_view

\begin{figure}[t]
    \centering
    \includegraphics[width=0.75\columnwidth]{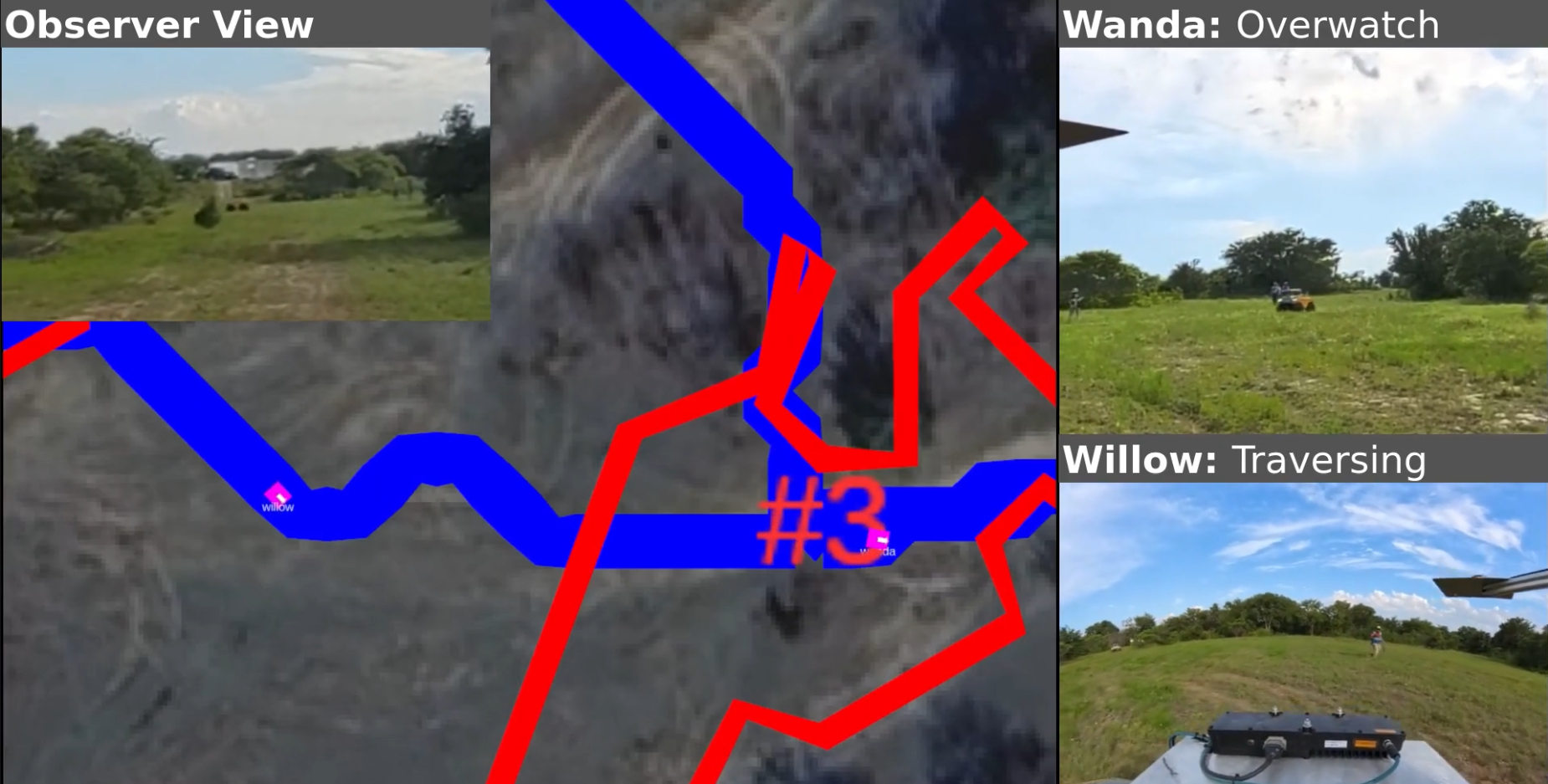}
    \vspace*{-2mm}
    \captionsetup{font=scriptsize}
    \caption{Overwatch view of robot Wanda while robot Willow traverses an edge. Observer view in the top left corner shows the traversing robot being detectable from the observer perspective.}
    \label{fig:overwatch_view}
    \vspace*{-4mm}
\end{figure} % overwatch_view

\begin{figure}[t]
    \centering
    \includegraphics[width=0.7\columnwidth]{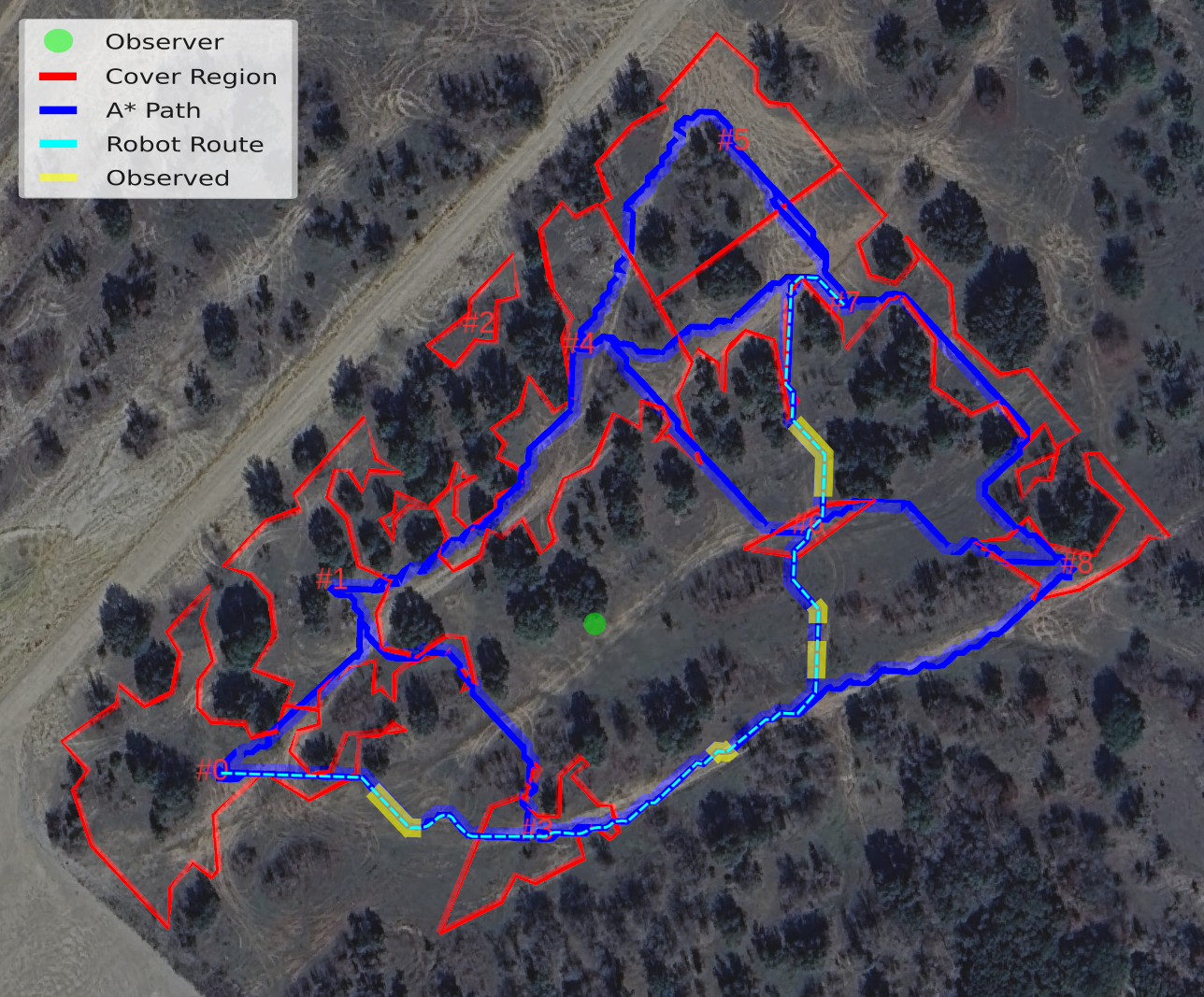}
    \vspace*{-2mm}
    \captionsetup{font=scriptsize}
    \caption{Depiction of the route traversed by the robots in Forested Graph 1, with path segments visible to the observer highlighted.}
    \label{fig:cavazos_observed}
    \vspace*{-4mm}
\end{figure} % cavazos_observed

\subsubsection{Forested Off-road Environment}

We consider a trapezoidal region of a forested off-road environment with varying types of brush, trees, and terrain. We segment the environment around two different observer locations, as depicted in Fig.~\ref{fig:cavazos_paths}. The change in the observer location results in significant variations in the cover regions and A$^*$ paths, ultimately resulting in different topologies for our \ac{DTG}.

\begin{figure*}[tbh]
    \centering
    % \vspace*{2mm}
    \begin{subfigure}[t]{0.6\columnwidth}
		\centering
		\includegraphics[width=\textwidth]{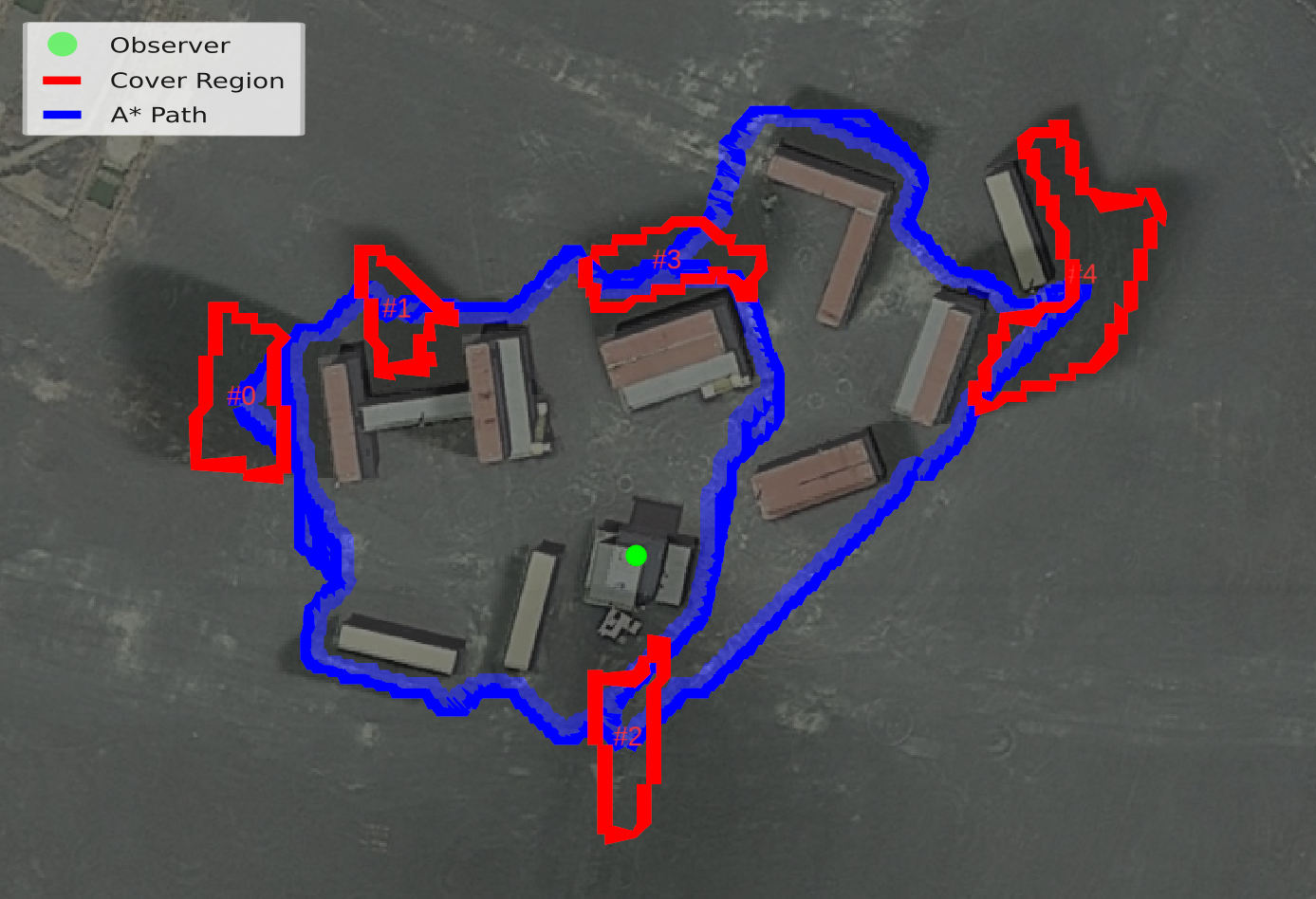}
		\captionsetup{font=scriptsize}
        \caption{Urban Graph 1}
        \label{fig:cover_paths:graph1}
	\end{subfigure}
	\begin{subfigure}[t]{0.6\columnwidth}
		\centering
		\includegraphics[width=\textwidth]{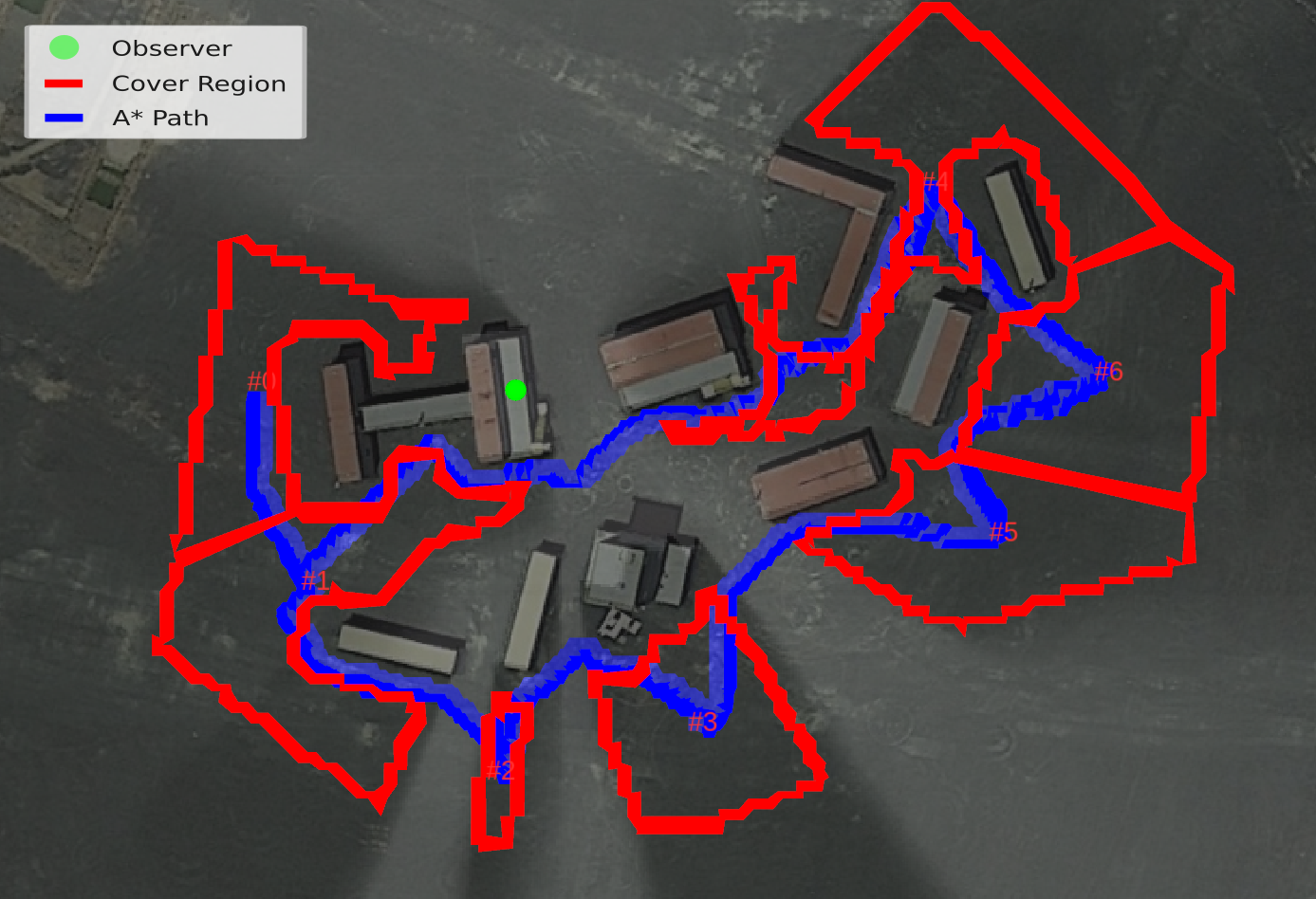}
		\captionsetup{font=scriptsize}
        \caption{Urban Graph 2}
        \label{fig:cover_paths:graph2}
	\end{subfigure}
	\begin{subfigure}[t]{0.6\columnwidth}
		\centering
		\includegraphics[width=\textwidth]{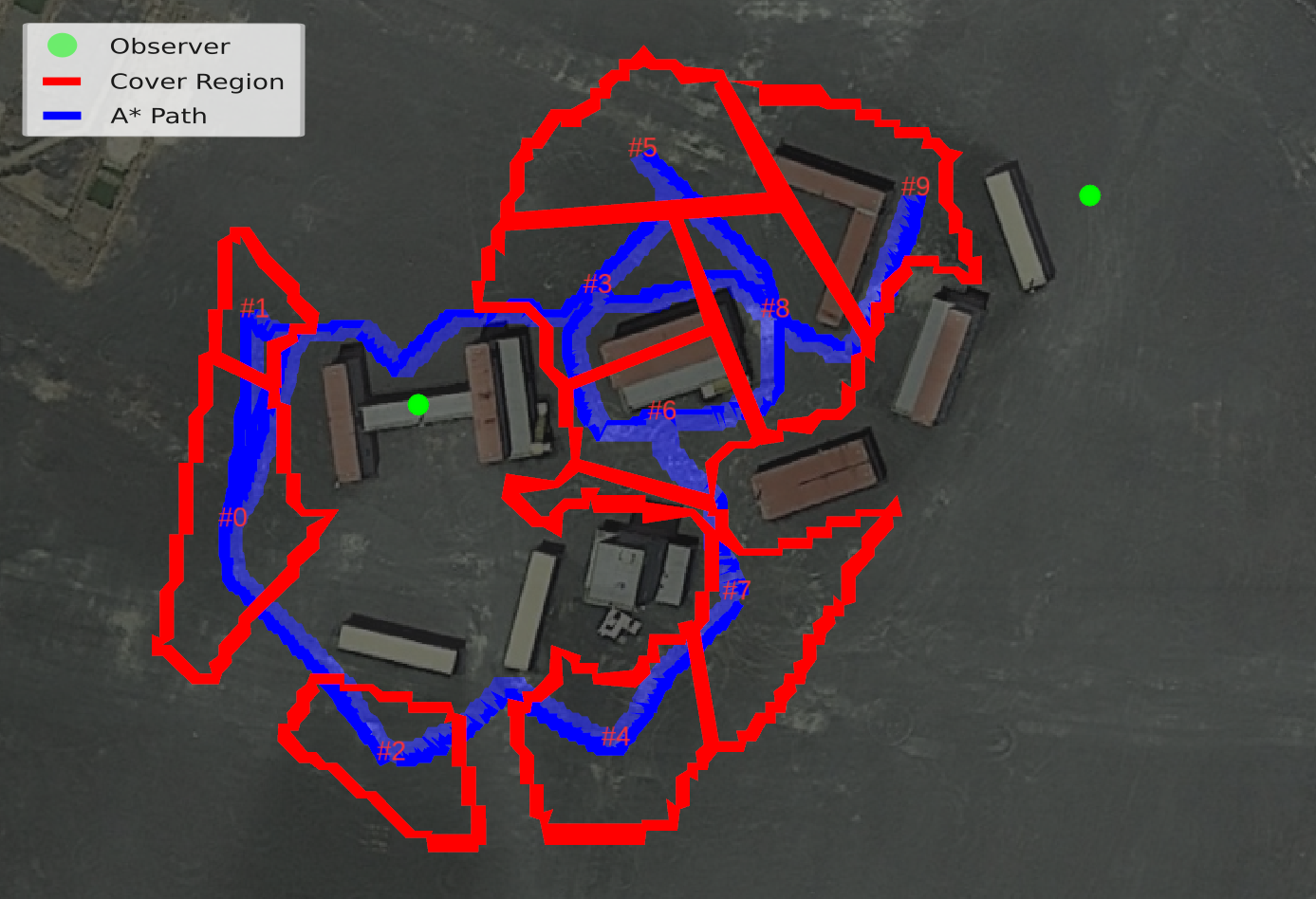}
		\captionsetup{font=scriptsize}
        \caption{Urban Graph 3}
        \label{fig:cover_paths:2adv}
	\end{subfigure}
	\caption{Environment segmentation for three different observer locations in an urban environment, including a case with multiple observers. The visibility map is overlaid to show shadows in the regions of cover.}
	\label{fig:cover_paths}
    \vspace*{-5mm}
\end{figure*} % cover_paths

\begin{figure}[tbh]
    \centering
    % \vspace*{2mm}
    \begin{subfigure}[t]{0.48\columnwidth}
		\centering
		\includegraphics[width=\textwidth]{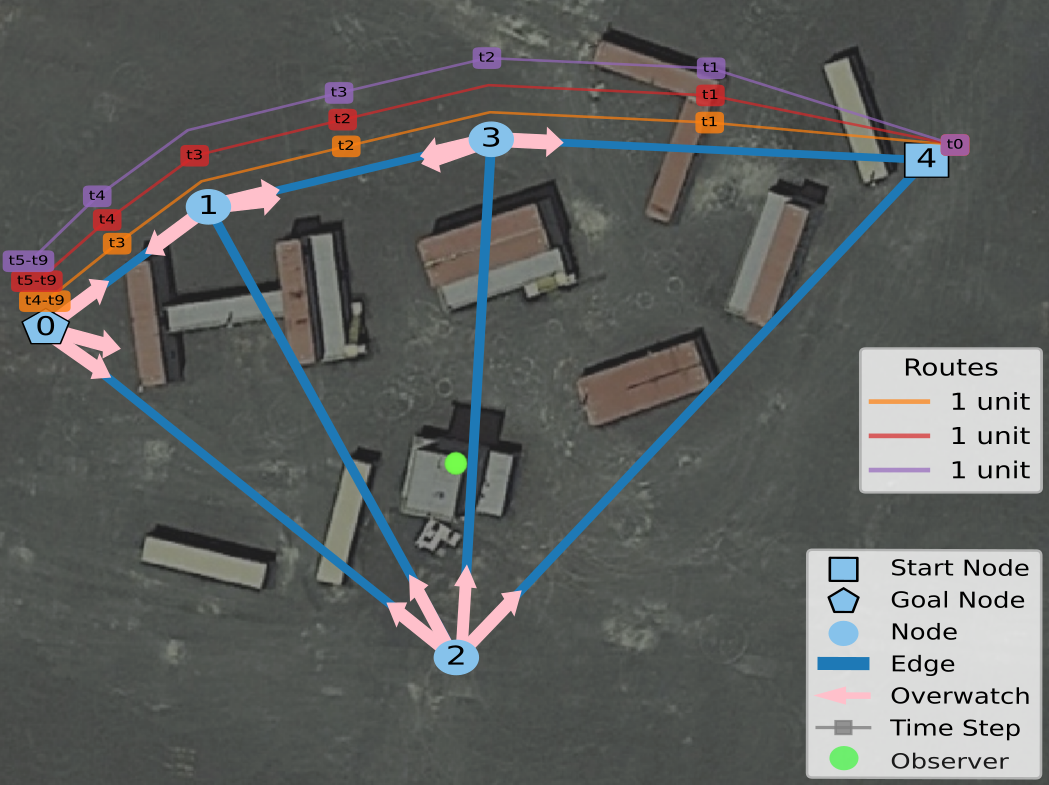}
		\captionsetup{font=scriptsize}
        \caption{Urban Graph 1: 1.0 overwatch scale}
        \label{fig:cover_graph1:w1}
	\end{subfigure}
	\begin{subfigure}[t]{0.48\columnwidth}
		\centering
		\includegraphics[width=\textwidth]{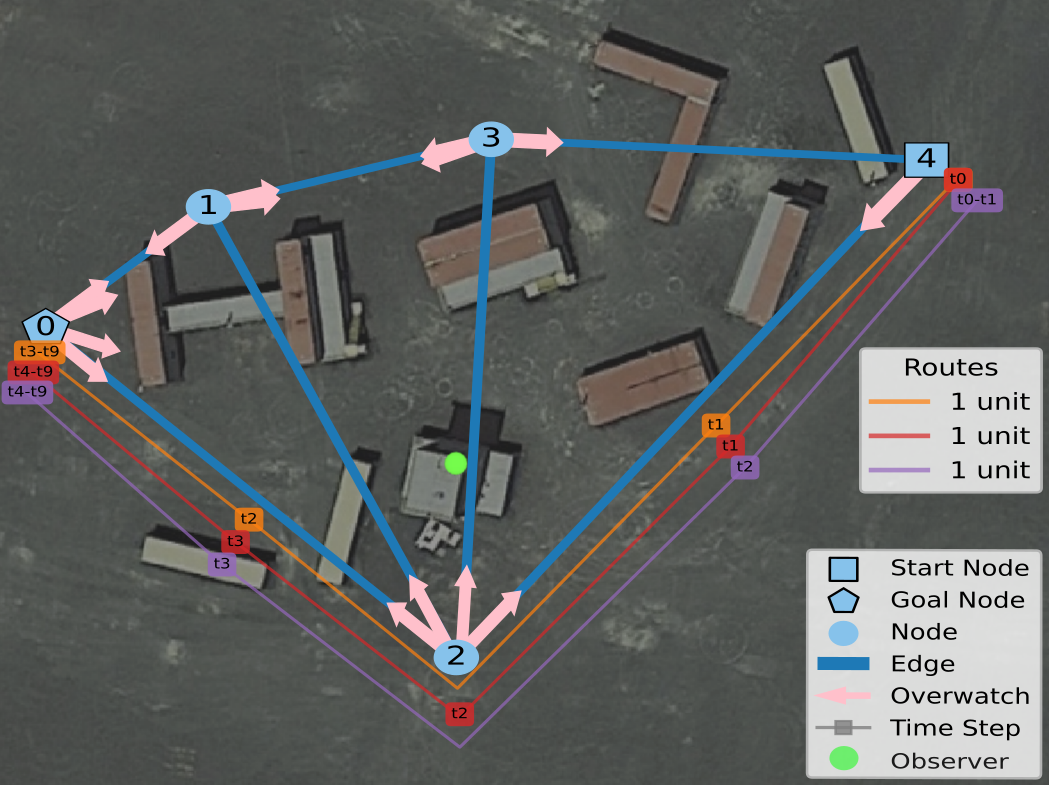}
		\captionsetup{font=scriptsize}
        \caption{Urban Graph 1: 1.3 overwatch scale}
        \label{fig:cover_graph1:w2}
	\end{subfigure}
	\caption{Planned routes on Urban Graph 1 showing the difference in routes when the overwatch scale increases. In this graph, the overwatch scale change resulted in different overwatch opportunities and a significant difference in the resulting team routes through the graph.}
	\label{fig:cover_graph1}
    \vspace*{-2mm}
\end{figure} % cover_graph1

\begin{figure}[tbh]
    \centering
    % \vspace*{2mm}
    \begin{subfigure}[t]{0.48\columnwidth}
		\centering
		\includegraphics[width=\textwidth]{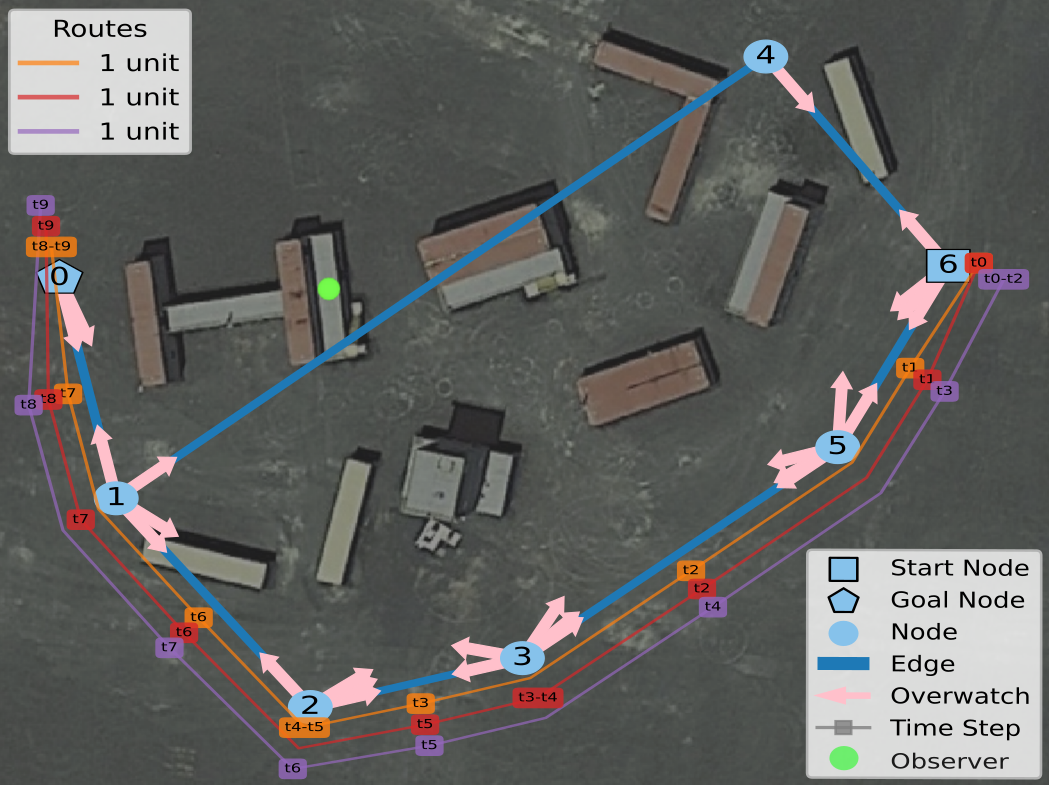}
		\captionsetup{font=scriptsize}
        \caption{Urban Graph 2: 1.0 overwatch scale}
	\end{subfigure}
	\begin{subfigure}[t]{0.48\columnwidth}
		\centering
		\includegraphics[width=\textwidth]{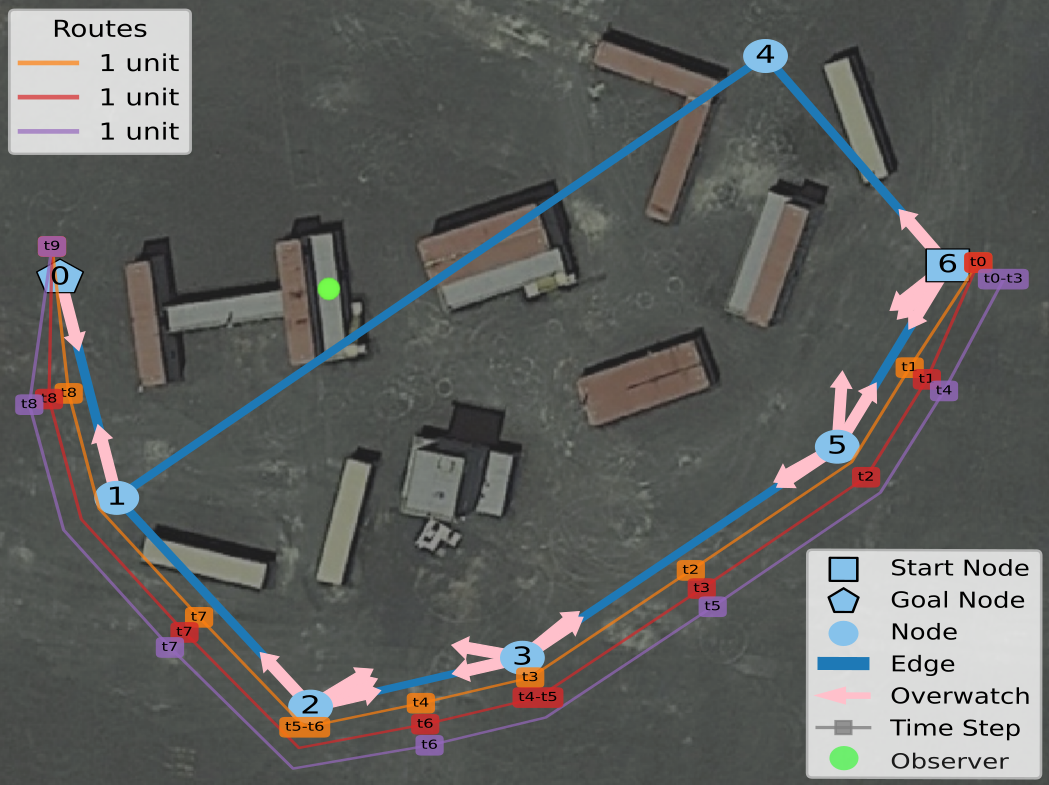}
		\captionsetup{font=scriptsize}
        \caption{Urban Graph 2: 0.4 overwatch scale}
	\end{subfigure}
	\caption{Planned routes on Urban Graph 2 showing the difference in routes when the overwatch scale decreases. In this graph, the overwatch scale change resulted in different overwatch opportunities and a subtle change to the resulting team paths through the graph.}
	\label{fig:cover_graph2}
    \vspace*{-3mm}
\end{figure} % cover_graph2

\begin{figure}[t]
    \centering
    \includegraphics[width=0.76\columnwidth]{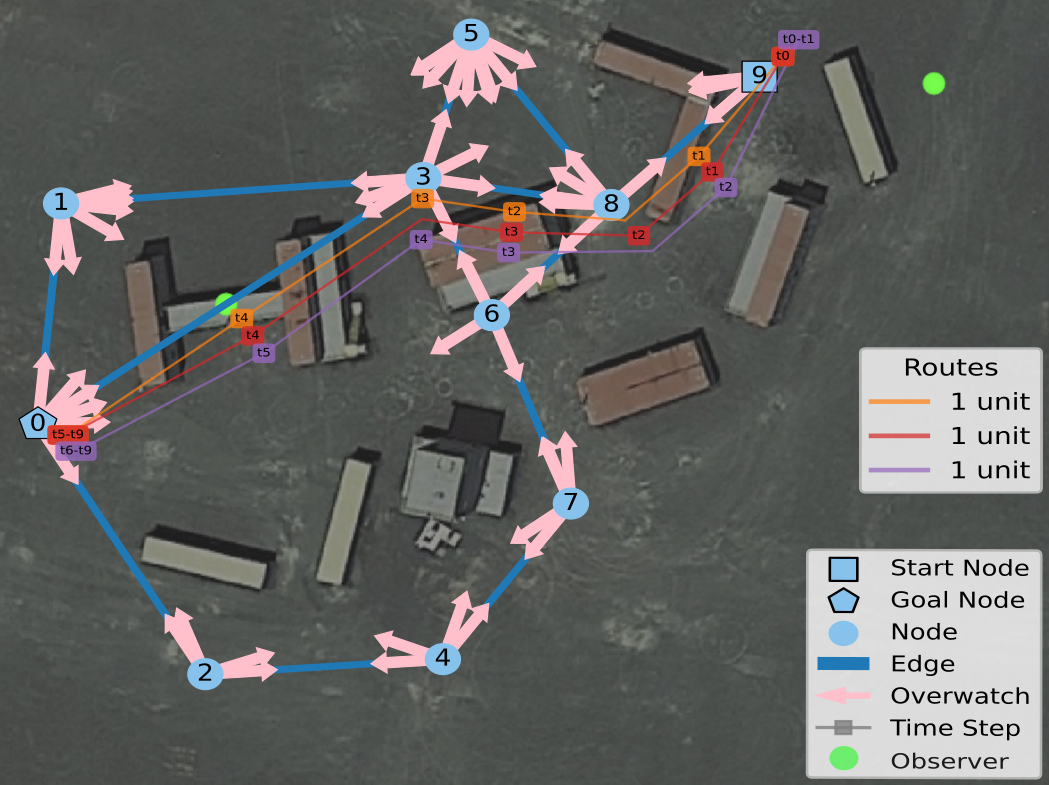}
    \vspace*{-2mm}
    \captionsetup{font=scriptsize}
    \caption{Planned routes on Urban Graph 3 with two observers.}
    \label{fig:cover_2adv}
    \vspace*{-2mm}
\end{figure} % cover_2adv

In Fig.~\ref{fig:cavazos_graphs}, we depict our schematic representation of the corresponding topological graphs with the overwatch opportunities indicated 
and optimal team routes, as solved for using our graph planning approach. Between Fig.~\ref{fig:cavazos_graphs:formation_loc1} and Fig.~\ref{fig:cavazos_graphs:loc1}, we varied the priority of overwatch, which results in more overwatch opportunities in Fig.~\ref{fig:cavazos_graphs:loc1}. We do this by setting the allowable ratio of overwatch reduction to edge cost (as described in Sec.~\ref{sec:env_seg_constr}) from 50\% to 10\%. When overwatch opportunities are available, they offer a significant reduction to the edge costs and thus are desirable to be used. This change results in the robots moving in formation on edge (3,6) when overwatch is less valuable versus performing bounding overwatch across that edge when overwatch is more valuable. Ultimately, in an operational scenario these types of factors and thresholds would be set based on the priorities of the scenario and the perceived risk associated with performing overwatch (i.e., the relative value of a small amount of overwatch compared to moving as a team). These two different solutions on the same graph demonstrate the adaptivity of our algorithm to varying objectives. 

Fig.~\ref{fig:cavazos_graphs:loc1} and Fig.~\ref{fig:cavazos_graphs:loc2} utilize the same overwatch parameters on the graphs generated from the different observer locations in Fig.~\ref{fig:cavazos_paths}. The observer location significantly impacts the resulting \ac{DTG}, and thus the routes through the environment. 

Fig.~\ref{fig:formation_view} depicts the robots moving in formation on edge (3,6), following the paths in Fig.~\ref{fig:cavazos_graphs:formation_loc1}. The vehicles maintain a set following distance along the A$^*$ paths while avoiding environmental obstacles.
Fig.~\ref{fig:overwatch_view} shows the robot Wanda performing overwatch at node 3 while the robot Willow traverses along edge (0,3). Each robot is equipped with a 360 degree camera. Additionally, the top left corner shows the observer view while Wanda is visible from the observer's perspective.

The route traversed by the robots in Forested Graph 1, in Figures \ref{fig:cavazos_graphs:formation_loc1} and \ref{fig:cavazos_graphs:loc1}, is outlined in Fig.~\ref{fig:cavazos_observed}. In Fig.~\ref{fig:cavazos_observed}, we highlight the regions of this route where the robots were observed by a camera located at the observer position. This demonstrates that our cover regions were a conservative estimate of areas that would not be visible from the observer location. 

\subsubsection{Protection Metric}

To evaluate the experimental performance of the multi-robot team, we propose the following metric to characterize the team's level of protection:
\begin{align}
\mathcal{M}_{\textrm{protect}} = \frac{1}{n_A} \sum_{i=1}^{n_A} \frac{1}{D^i} \sum_{e \in E} (d^i_{o,e} + d^i_{f,e} + d^i_{c,e}).
\end{align}
In this metric, for each robot $i$ and edge $e$, we consider the distances the robots traverse using three forms of protection: (i) the distance a robot was monitored through overwatch $d^i_{o,e}$; (ii) the distance traversed in formation $d^i_{f,e}$; and (iii) the distance traversed in cover $d^i_{c,e}$.
The total distance traveled by a particular robot is $D^i$. 
When a vehicle utilizes one form of protection (traversing in cover, formation, or being overwatched) throughout the entire path, this metric is $\mathcal{M}_{\textrm{protect}}=1$. Additional protection will increase this score to a maximum of $3$. 

We consider three scenarios on Forested Graph 1 to compare the levels of protection: (1) two robots moving in formation on edge (3,6) following the plan in Fig.~\ref{fig:cavazos_graphs:formation_loc1}, (2) two robots moving in bounding overwatch from Fig.~\ref{fig:cavazos_graphs:loc1}, and (3) a single robot traversing the route shown in Fig.~\ref{fig:cavazos_observed}. 
For each of these scenarios, we computed the protection metric using camera data. A vehicle is considered to be in cover when it is not visible from the observer's location.
We present the protection metric values in Table~\ref{tab:protection_metric}. As expected, the single robot scenario has the lowest score since it is not able to be further protected by a teammate. In this graph, the robots moving in formation on edge (3,6) received a higher protection metric than the bounding overwatch scenario since the robots in formation maintain visual contact with each other throughout the path. In comparison, the duration of overwatch on edge (3,6) is shorter due to terrain features blocking the robot's field of view.  
If one form of protection were determined to be superior to the others for a particular scenario, the protection metric could be weighted accordingly. 

\begin{table}[tbh]
    \vspace*{-1mm}
	\centering
	\caption{Protection Metric for Forested Graph 1}
	\label{tab:protection_metric}
    \vspace*{-2mm}
	\begin{center}
		\renewcommand{\arraystretch}{1.3}
		\begin{tabular}{ c | c }
			\textbf{Scenario} & \textbf{Protection Metric} \\
			\hline \hline
			With Formation & 1.78 \\
			Without Formation, Bounding Overwatch & 1.65 \\
			Single Robot & 0.85 
		\end{tabular}
	\end{center}
    \vspace*{-3mm}
\end{table}

\subsubsection{Urban Environment Experiments}

We considered three different segmentations of an urban environment based on different observer locations, as seen in Fig.~\ref{fig:cover_paths}. We consider two scenarios with one observer and one scenario with two observers. In Fig.~\ref{fig:cover_paths}, the visibility map for each scenario is overlaid to show shadows in the regions of cover. 

\begin{figure}[tbh]
    \centering
    % \vspace*{2mm}
    \begin{subfigure}[t]{0.34\columnwidth}
		\centering
		\raisebox{0pt}[\height][\depth]{%
            \includegraphics[height=0.1\textheight]{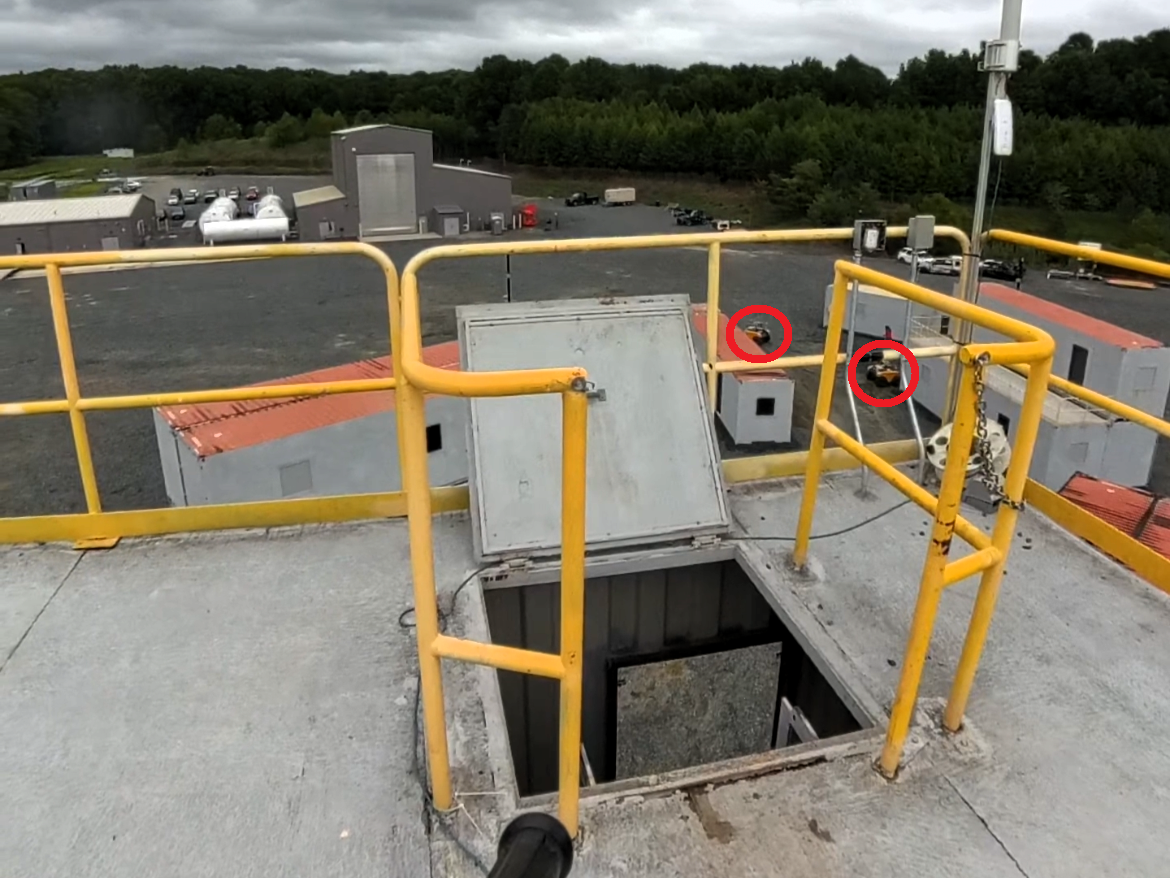}}
		\captionsetup{font=scriptsize}
        \caption{Urban Graph 1: robots in formation on edge (4,3).}
	\end{subfigure}
    \hspace*{1mm}
	\begin{subfigure}[t]{0.33\columnwidth}
        \centering
        \raisebox{0pt}[\height][\depth]{%
            \includegraphics[height=0.1\textheight]{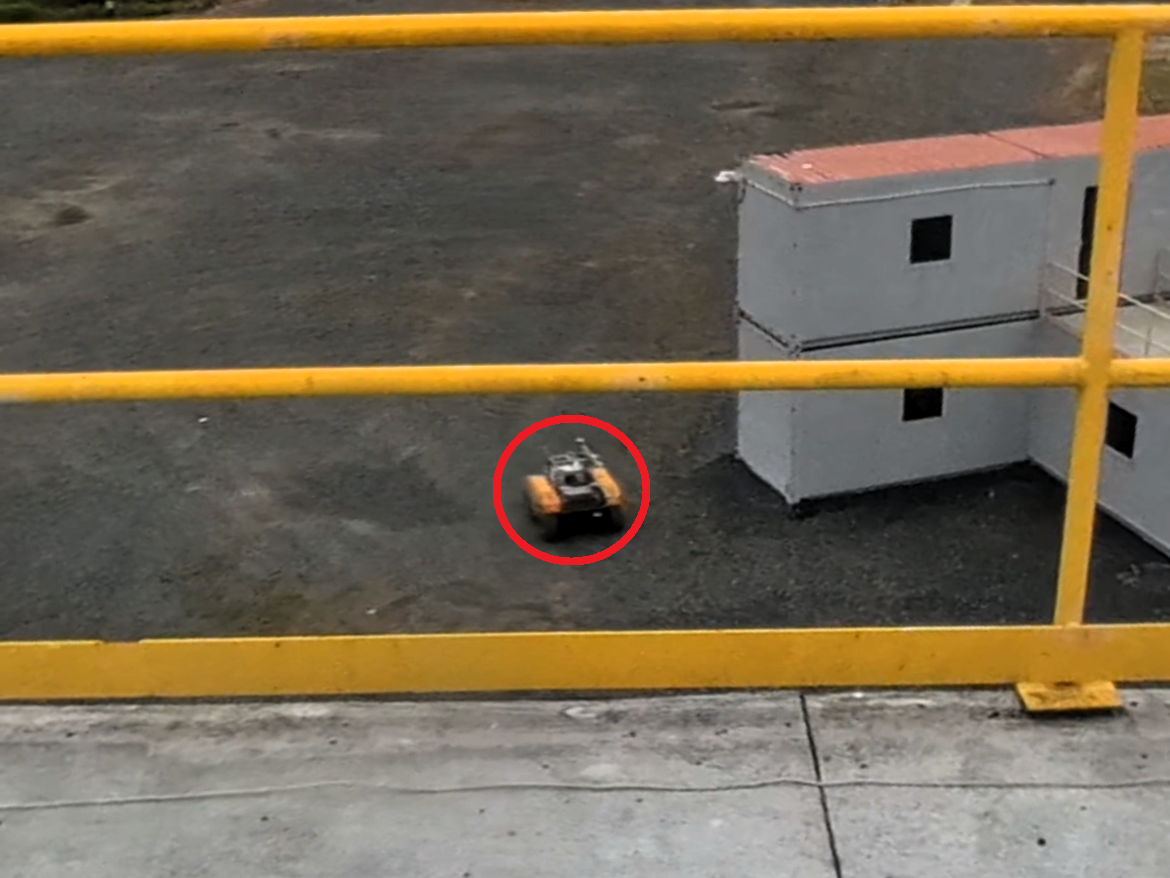}}
		\captionsetup{font=scriptsize}
        \caption{Urban Graph 1: robot on edge (2,0).}
	\end{subfigure}
    \hspace*{1.3mm}
	\begin{subfigure}[t]{0.22\columnwidth}
        \centering
        \raisebox{0pt}[\height][\depth]{%
            \includegraphics[height=0.1\textheight]{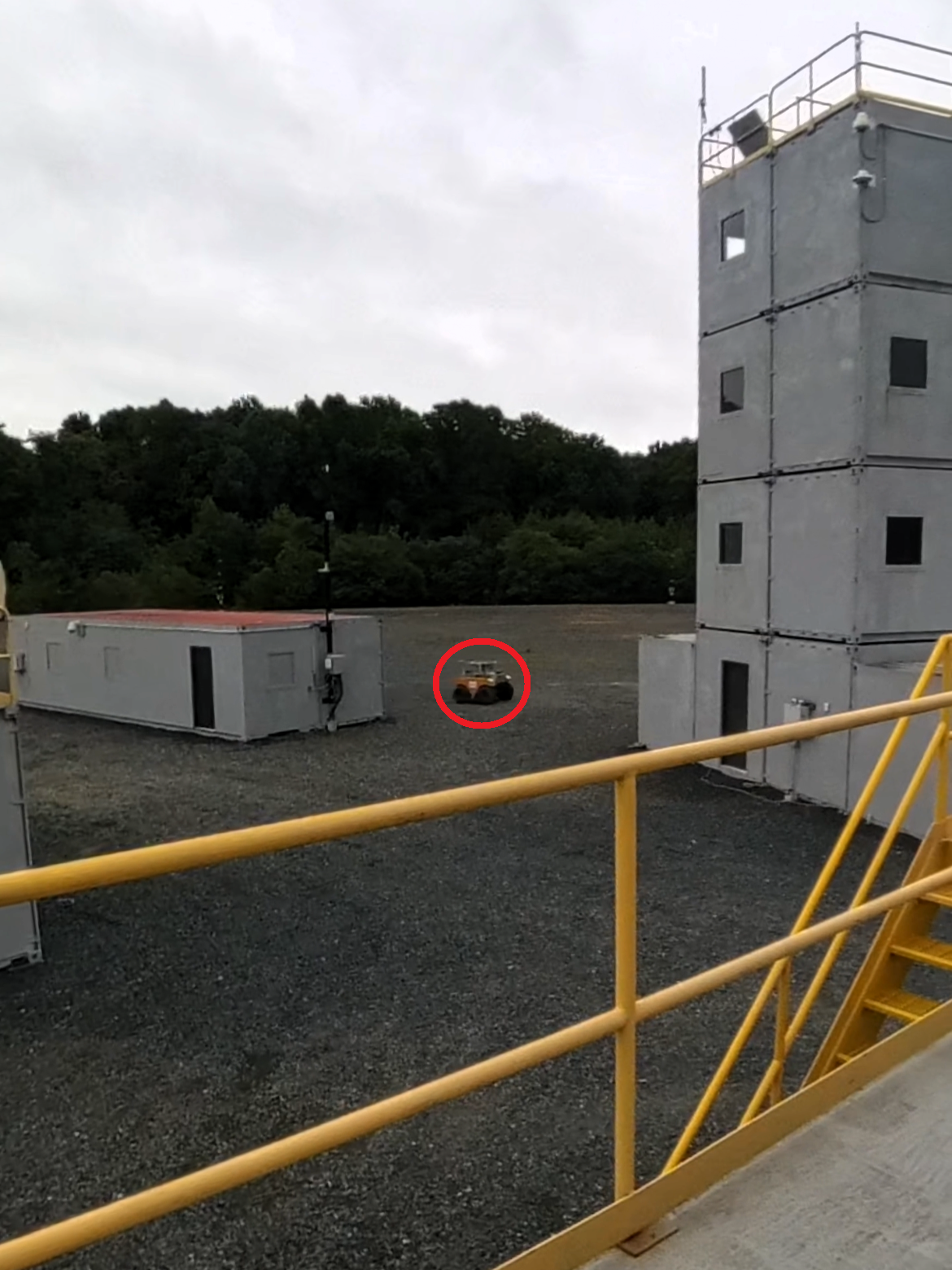}}
		\captionsetup{font=scriptsize}
        \caption{Urban Graph 2: robot on edge (5,3).}
	\end{subfigure}
    % \vspace*{-3mm}
	\caption{Observer views for Urban Graphs 1 and 2 with traversing robots circled in red. In (c), the Urban Graph 1 observer tower is shown on the right.}
	\label{fig:observer_views}
    \vspace*{-3mm}
\end{figure}

We generated graph plans for the segmentations of the environment shown in Fig.~\ref{fig:cover_paths:graph1} and Fig.~\ref{fig:cover_paths:graph2} using two different scale factors for overwatch. As described in Sec.~\ref{sec:env_seg_constr}, a particular scenario may consider overwatch to be more valuable and the overwatch scale factor allows that high-level guidance to vary the \ac{DTG}. This graph variation demonstrates the control the user has over the high-level behaviors of the robots when generating these coordinated maneuvers. The resulting graph plans are shown in Fig.~\ref{fig:cover_graph1} and Fig.~\ref{fig:cover_graph2}. In Fig.~\ref{fig:cover_2adv} we do not scale the overwatch (i.e., scale of 1) to instead demonstrate planning with two observer locations. 
Fig.~\ref{fig:observer_views} shows segments of the observers' views when the robots were able to be detected in Urban Graphs 1 and 2. 

% Increased scale of overwatch from 1.0 to 1.3 - graph 1
% Reduced scale of overwatch from 1.0 to 0.4 - graph 2
% 2 adv is the base params

Fig.~\ref{fig:cover_graph1_formation} shows a third person view of the robots arranged at node 4 before moving into formation to traverse edge (4,3) in the first step of the plan in Fig.~\ref{fig:cover_graph1:w1}. Fig.~\ref{fig:cover_graph1_formation} denotes the leader and follower goal points in cover region 5 and shows the \ac{MPPI} trajectories of the two robots that begin moving first. The third robot waits for the other two to get ahead of it before moving into its position in the formation.

 % observer_views

%%%%%%%%%%%%%%%%%%%%%%%%%%%%%%%%%%%%%%%%%%%%%%%%%%%%%%%%%%%%%%%%%%%%%%%%%%%%%%%%
\section{Conclusion}

Our hierarchical planning approach \ac{STALC} efficiently generates coordinated tactics using \emph{a priori} data while compensating for changing environments with the lower-levels of the autonomy stack. We demonstrated expressing complicated scenarios compactly with a \acf{DTG} and efficient \ac{MILP} formulation, enabling rapid end-to-end planning in real-world scenarios. We evaluated our approach in simulation and hardware experiments showing the full system's autonomous operation, including swapping between overwatch and formation conditions in highly constrained environments with static and dynamic obstacles. Future research will include receding-horizon graph-based planning to update the visibility map online in changing conditions, and mid-range planning that utilizes real-time sensor data.

%\addtolength{\textheight}{-1.8cm} % This command serves to balance the column lengths

%%%%%%%%%%%%%%%%%%%%%%%%%%%%%%%%%%%%%%%%%%%%%%%%%%%%%%%%%%%%%%%%%%%%%%%%%%%%%%%%

%%%%%%%%%%%%%%%%%%%%%%%%%%%%%%%%%%%%%%%%%%%%%%%%%%%%%%%%%%%%%%%%%%%%%%%%%%%%%%%%

%%%%%%%%%%%%%%%%%%%%%%%%%%%%%%%%%%%%%%%%%%%%%%%%%%%%%%%%%%%%%%%%%%%%%%%%%%%%%%%%
%\section*{APPENDIX}

%%%%%%%%%%%%%%%%%%%%%%%%%%%%%%%%%%%%%%%%%%%%%%%%%%%%%%%%%%%%%%%%%%%%%%%%%%%%%%%%
\section*{Acknowledgment}

Research was sponsored by the Army Research Laboratory and was accomplished under Cooperative Agreement Number W911NF-22-2-0241 and W911NF-25-2-0033. The views and conclusions contained in this document are those of the authors and should not be interpreted as representing the official policies, either expressed or implied, of the Army Research Laboratory or the U.S. Government. The U.S. Government is authorized to reproduce and distribute reprints for Government purposes notwithstanding any copyright notation herein.

%%%%%%%%%%%%%%%%%%%%%%%%%%%%%%%%%%%%%%%%%%%%%%%%%%%%%%%%%%%%%%%%%%%%%%%%%%%%%%%%

\bibliographystyle{IEEEtran}
\bibliography{references.bib}

\vspace{-2\baselineskip}
\begin{IEEEbiography}[{\includegraphics[width=1in,height=1in,clip,keepaspectratio]{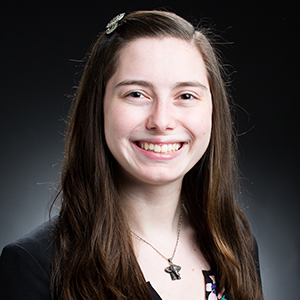}}]{Cora A. Duggan}
received her B.S. in Mechanical Engineering and Applied Mathematics and Statistics in 2017, M.S.E. in Robotics in 2019, and Ph.D. in Mechanical Engineering in 2025 from Johns Hopkins University, Baltimore, MD, USA. 
She is currently a Robotics Researcher at the Johns Hopkins Applied Physics Laboratory, Laurel, MD, USA. Her research interests include multi-robot coordination, autonomous scene understanding, robot agility, and motion planning.\end{IEEEbiography}

\vspace{-2\baselineskip}
\begin{IEEEbiography}[{\includegraphics[width=1in,height=1in,clip,keepaspectratio]{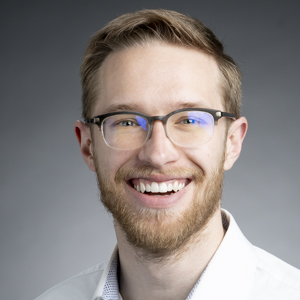}}]{Adam Goertz}
received the B.S. degree in Computer Engineering from the University of Arkansas, Fayetteville, AR, USA, in 2020 and the M.S. degree in Robotics from the University of Michigan, Ann Arbor, MI, USA in 2021. He is currently a Research Engineer at the Johns Hopkins Applied Physics Laboratory, Laurel, MD, USA, where he is engaged in research on multi-robot planning \& control; agentic control, testing, and evaluation; and AI for science.\end{IEEEbiography}

\vspace{-2\baselineskip}
\begin{IEEEbiography}[{\includegraphics[width=1in,height=1in,clip,keepaspectratio]{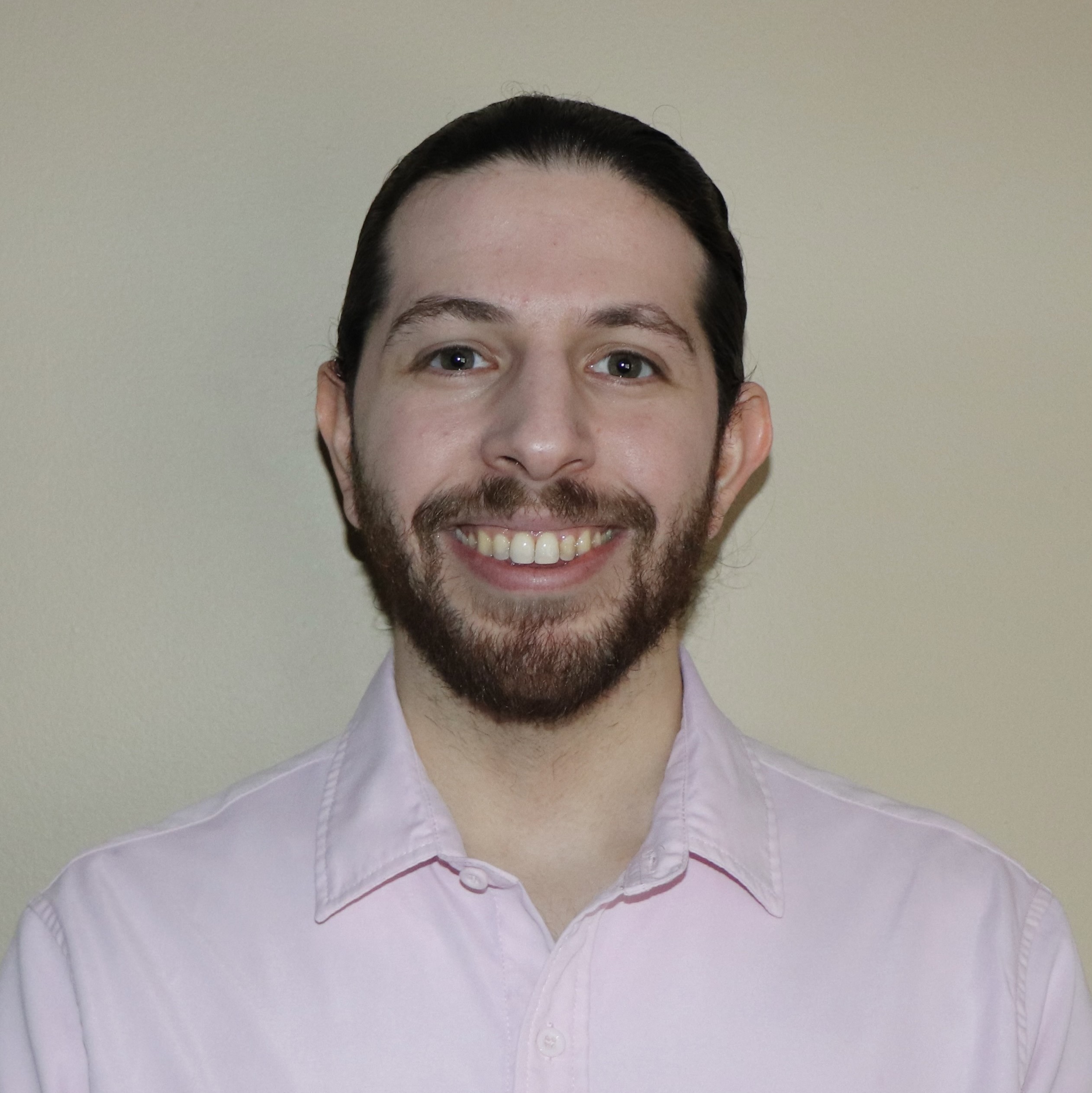}}]{Adam Polevoy}
received a bachelor's degree in Biomedical Engineering and a master's degree in Robotics from Johns Hopkins University, Baltimore, MD, USA in 2019 and 2020. He is currently a robotics engineer at the Johns Hopkins Applied Physics Laboratory and a mechanical engineering Ph.D. student at Johns Hopkins University. His research interests include robot navigation, stochastic optimal control, and reinforcement learning.\end{IEEEbiography}

\vspace{-2\baselineskip}
\begin{IEEEbiography}[{\includegraphics[width=1in,height=1in,clip,keepaspectratio]{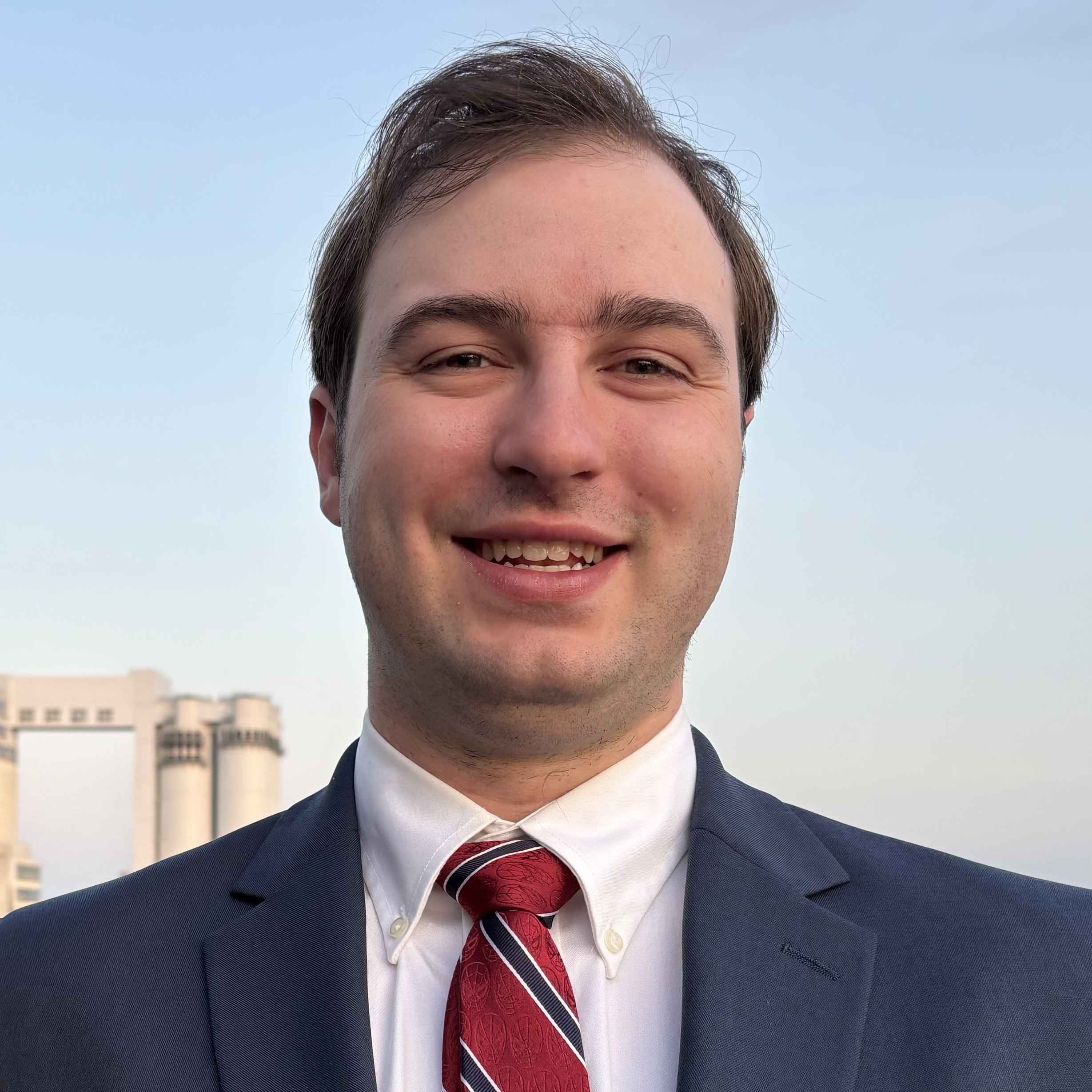}}]{Mark Gonzales}
received his B.S. in Mechanical Engineering and M.S.E. in Robotics from Johns Hopkins University, Baltimore, MD, USA. He is currently pursuing a Ph.D. with the Agile and Intelligent Robotics Laboratory under the supervision of Dr. Joseph Moore. His research focuses on multi-agent formation planning for systems with nonholonomic dynamics operating in constrained environments.\end{IEEEbiography}

\vspace{-2\baselineskip}
\begin{IEEEbiography}[{\includegraphics[width=1in,height=1in,clip,keepaspectratio]{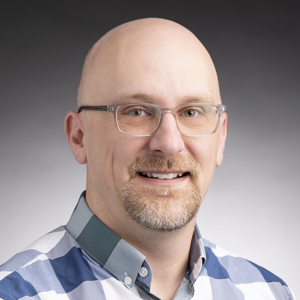}}]{Kevin C. Wolfe}
received the B.S. degree in mechanical engineering from The College of New Jersey, Ewing, NJ, USA, in 2007, and the M.S.E. and Ph.D. degrees in mechanical engineering from Johns Hopkins University, Baltimore, MD, USA, in 2010 and 2012, respectively. He is currently a Senior Roboticist and Chief Scientist for the Tailored Autonomous Systems Group at the Johns Hopkins Applied Physics Laboratory, Laurel, MD, USA. His research interests include robotic systems for challenging environments and medical applications, multi-robot coordination and planning, robotic perception, and miniature robotic platforms.
\end{IEEEbiography}

\vspace{-2\baselineskip}
\begin{IEEEbiography}[{\includegraphics[width=1in,height=1in,clip,keepaspectratio]{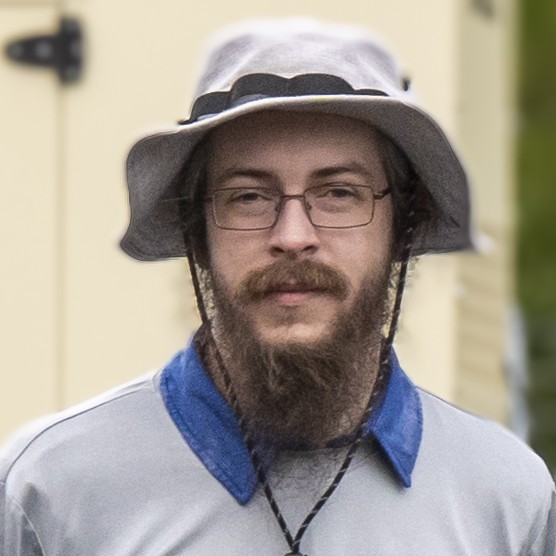}}]{Bradley Woosley}
received his B.S. in Computer Engineering in 2013 from the University of Nebraska-Lincoln, Lincoln, NE, USA, his M.S. in Computer Science in 2015 and his Ph.D. in Information Technology in 2020 from the University of Nebraska at Omaha, Omaha, NE, USA. He is a research scientist at the DEVCOM Army Research Laboratory specializing in autonomy for mutli-robot systems.
\end{IEEEbiography}

\vspace{-2\baselineskip}
\begin{IEEEbiography}[{\includegraphics[width=1in,height=1in,clip,keepaspectratio]{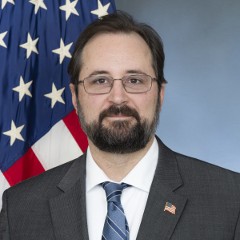}}]{John G. Rogers III}
is a senior research scientist and program manager at the DEVCOM Army Research Laboratory (ARL) specializing in machine learning, artificial intelligence, and autonomy for multi-robot systems. Dr. Rogers has completed bachelor’s and master’s degrees at Carnegie Mellon in electrical and computer engineering, a master’s degree in computer science at Stanford, and a Ph.D. degree in robotics at Georgia Tech. At ARL, Dr. Rogers is leading efforts in multi-robot distributed state estimation and coordinated tactical autonomy.\end{IEEEbiography}

\vspace{-2\baselineskip}
\begin{IEEEbiography}[{\includegraphics[width=1in,height=1in,clip,keepaspectratio]{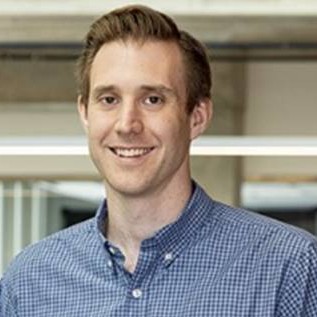}}]{Joseph Moore} is an Assistant Professor of Mechanical Engineering at Johns Hopkins University (JHU) and the director of the Agile and Intelligent Robotics (AIRO) Laboratory. He received his Ph.D. in Mechanical Engineering from the Massachusetts Institute of Technology in 2014. Following his Ph.D., he joined the JHU Applied Physics Laboratory (APL), first as a Postdoctoral Fellow, and then as a member of the technical staff. From 2021-2024, he served as the Chief Scientist for the APL Robotics Group. Dr. Moore’s research explores methods at the intersection of computational control, computational physics, and machine learning to create robust, highly-agile robotic systems.
\end{IEEEbiography}
\vspace{-2\baselineskip}

\end{document}